\documentclass[letterpaper, 10 pt, conference]{ieeeconf}  

\IEEEoverridecommandlockouts                              

\usepackage{multicol}
\usepackage[bookmarks=true]{hyperref}
\usepackage{graphicx}
\usepackage{multicol}
\usepackage{subcaption}
\usepackage{caption}
\usepackage{gensymb}
\usepackage{tabularx}
\usepackage{amsmath}
\usepackage{amssymb}
\usepackage[ruled,vlined]{algorithm2e}
\usepackage{algorithmic}
\newcommand{\argmin}{\operatornamewithlimits{argmin}}
\usepackage{float}
\usepackage{wrapfig}

\usepackage{cite}

\usepackage{xargs}
\usepackage[pdftex,dvipsnames]{xcolor}  

\begin{document}

\title{\LARGE \bf
Radar SLAM: A Robust SLAM System for All Weather Conditions}

\author{Ziyang Hong, Yvan Petillot, Andrew Wallace and Sen Wang
\thanks{The authors are with Edinburgh Centre for Robotics, Heriot-Watt University, Edinburgh, EH14 4AS, UK.
        {\tt\small \{zh9, y.r.petillot, a.m.wallace, s.wang\}@hw.ac.uk}}%
}

\maketitle
\thispagestyle{empty}
\pagestyle{empty}

\begin{abstract}

A Simultaneous Localization and Mapping (SLAM) system must be robust to support long-term mobile vehicle and robot applications. However, camera and LiDAR based SLAM systems can be fragile when facing challenging illumination or weather conditions which degrade their imagery and point cloud data. Radar, whose operating electromagnetic spectrum is less affected by environmental changes, is promising although its distinct sensing geometry and noise characteristics bring open challenges when being exploited for SLAM.
This paper studies the use of a Frequency Modulated Continuous Wave radar for SLAM in large-scale outdoor environments. We propose a full radar SLAM system, including a novel radar motion tracking algorithm that leverages radar geometry for reliable feature tracking. It also optimally compensates motion distortion and estimates pose by joint optimization. Its loop closure component is designed to be simple yet efficient for radar imagery by capturing and exploiting structural information of the surrounding environment.
Extensive experiments on three public radar datasets, ranging from city streets and residential areas to countryside and highways, show competitive accuracy and reliability performance of the proposed radar SLAM system compared to the state-of-the-art LiDAR, vision and radar methods. The results show that our system is technically viable in achieving reliable SLAM in extreme weather conditions, e.g. heavy snow and dense fog, demonstrating the promising potential of using radar for all-weather localization and mapping.

\end{abstract}


\begin{figure*}[t]
    \includegraphics[width=\linewidth]{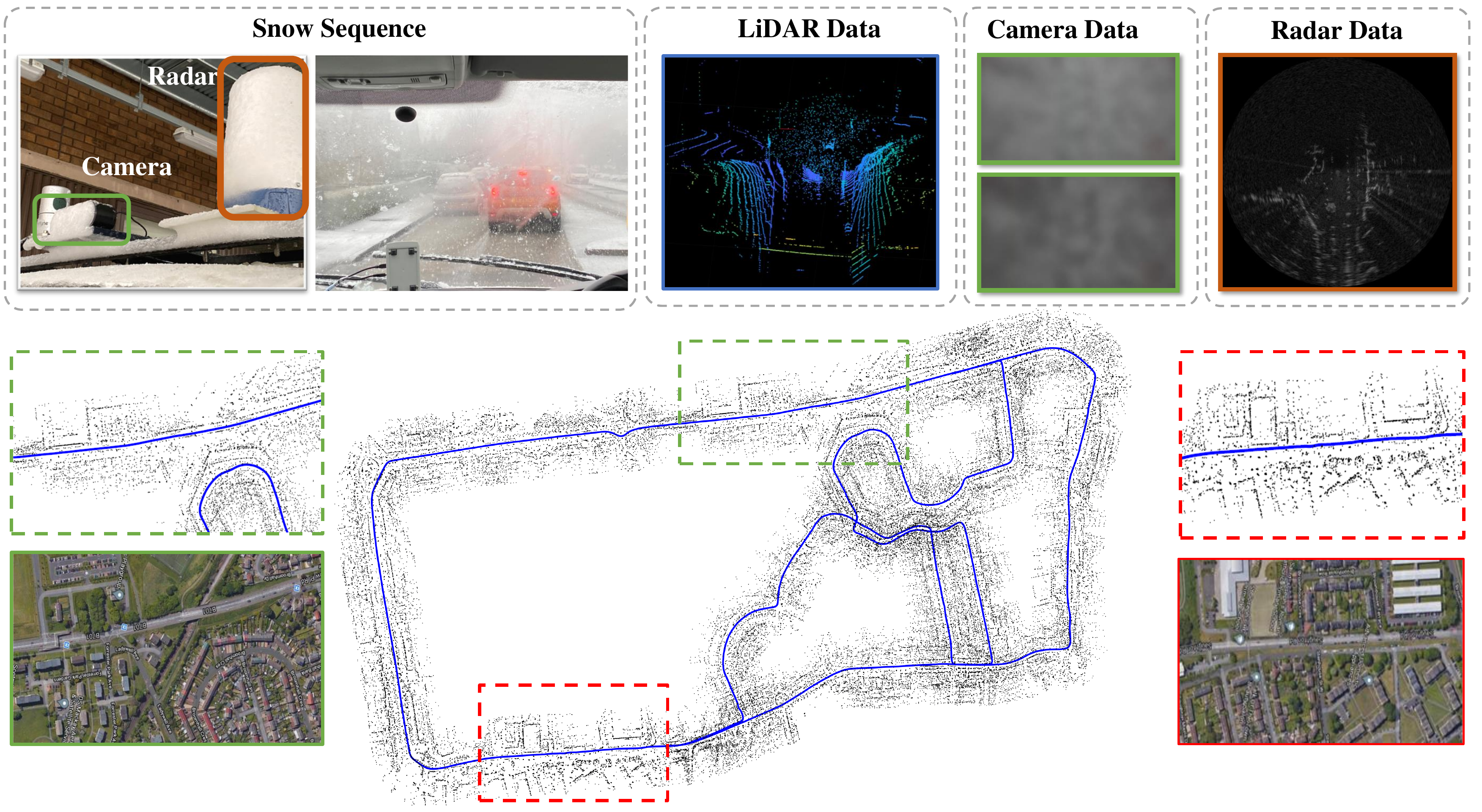}
    \caption{Map and trajectory estimated by our proposed radar SLAM system on a self-collected Snow Sequence. We can observe random noisy LiDAR points around the vehicle due to reflection from snowflakes. The camera is completely covered by frozen snow. The magnified areas are compared with satellite images showing reconstructed buildings and roads. Our proposed radar SLAM method can successfully handle this challenging sequence with heavy snowfall.}
\end{figure*}

\section{Introduction}

Simultaneous Localization and Mapping (SLAM) has attracted substantial interest over recent decades, and extraordinary progress has been made in the last 10 years in both the robotics and computer vision communities. In particular, camera and LiDAR based SLAM algorithms have been extensively investigated \cite{engel2014lsd,mur2015orb,loamZhang,shan2018lego} and progressively applied to various real-world applications. Their robustness and accuracy are also improved further by fusing with other sensing modalities, especially Inertial Measurement Unit (IMU) based motion as a prior \cite{qin2018vins,campos2020orb,liosam2020shan}.

Most existing camera and LiDAR sensors fundamentally operate within or near visible electromagnetic spectra, which means that they are more susceptible to illumination changes, floating particles and water drops in environments. It is well-known that vision suffers from low illumination, causing image degradation with dramatically increased motion blur, pixel noise and texture losses. The qualities of LiDAR point clouds and camera images can also degenerate significantly, for instance, when facing a realistic density of fog particles, raindrops and snowflakes in misty, rainy and snowy weather. Given the fact that a motion prior is mainly effective in addressing short-period and temporary sensor degradation, even visual-inertial or LiDAR-inertial SLAM systems are anticipated to fail in these challenging weather conditions. Therefore, how to construct a robust localization and mapping system operating in adverse weathers is still an open problem.




Radar is another type of active sensor, whose electromagnetic spectrum usually lies in the much lower frequency (GHz) band than camera and LiDAR (from THz to PHz). Therefore, it can operate more reliably in the majority of weather and light conditions. It also offers extra values, e.g. further sensing range, relative velocity estimates from the Doppler effect and absolute range measurement. Recently, radar has been gradually considered to be indispensable for safe autonomy and has been increasingly adopted in the automotive industry for obstacle detection and Advanced Driver-Assistance Systems (ADAS). Meanwhile, recent advances in Frequency-Modulated Continuous-Wave (FMCW) radar systems make radar sensing more appealing since it is able to provide a relatively dense representation of the environment, instead of only returning sparse detections.

However, radar has a distinct sensing geometry and its data is formed very differently from vision and LiDAR. Therefore, there are new challenges for radar based SLAM compared to vision and LiDAR based SLAM. For example, its noise and clutter characteristics are rather complex as a mixture of many sources, e.g., electromagnetic radiation in the atmosphere and multi-path reflection, and its noise level tends to be much higher. This means that existing feature extraction and matching algorithms may not be well suited for radar images. The usual lack of elevation information is also distinct from camera and LiDAR sensors. Therefore, the potential of using recently developed FMCW radar sensors to achieve robust SLAM is yet to be explored.

In this paper, we propose such a novel SLAM system based on an FMCW radar. It can operate robustly in various outdoor scenarios, e.g. busy city streets and highways, and weather conditions, e.g. heavy snowfall and dense fog. Our main contributions are:
\begin{itemize}
    \item A robust data association and outlier rejection mechanism for radar based feature tracking by leveraging radar geometry.
    \item A novel motion compensation model formulated to reduce motion distortion induced by a low scanning rate. The motion compensation is jointly optimized with pose estimation in an optimization framework.
    \item A fast and effective loop closure detection scheme designed for a FMCW radar with dense returns.
    \item Extensive experiments on three available public radar datasets, demonstrating and validating the feasibility of a SLAM system operating in extreme weather conditions.
    \item Unique robustness and minimal parameter tuning, i.e., the proposed radar SLAM system is the only competing method which can work properly on all data sequences, in particular using \textit{an identical set of parameters} without much parameter tuning.
\end{itemize}

The rest of the paper is structured as follows. In Section \ref{sec:related_work}, we discuss related work. In Section \ref{sec:radar_sensing}, we elaborate on the geometry of radar sensing and the challenges of using radar for SLAM. An overview of the proposed system is given in Section \ref{sec:system_overview}. The proposed motion compensation tracking model is presented in Section \ref{sec:tracking_model}, followed by the loop closure detection and pose graph optimization in Section \ref{sec:loop_pose_graph}. Experiments, results and system parameters are presented in Section \ref{sec:result}. Finally, the conclusions and future work are discussed in Section \ref{sec:conclusion}.


\section{Related Work} \label{sec:related_work}
In this section, we discuss related work on localization and mapping in extreme weather conditions using optical sensor modalities, i.e. camera and LiDAR. We also review the past and current state-of-the-art radar based localization and mapping methods.

\subsection{Vision and LiDAR based Localization and Mapping in Adverse Weathers}
Typical adverse weather conditions include rain, fog and snow which usually cause degradation on image quality or produce undesired effects, e.g. due to rain streaks or ice. Therefore, significant efforts have been made to alleviate this impact by pre-processing image sequences to remove the effects of rain \cite{garg2004detection}, \cite{ren2017video}, for example using a model based on matrix decomposition to remove the effects of both snow and rain in the latter case.  In contrast, \cite{li2016rain} removes the effects of rain streaks from a single image by learning the static and dynamic background using a Gaussian Mixture Model.  A de-noising generator that can remove noise and artefacts induced by the presence of adherent rain droplets and streaks is trained in \cite{porav2019can} using data from a stereo rig. A rain mask generated by temporal content alignment of multiple images is also used for keypoint detection \cite{huang2019reliable, yamada2019vision}. In spite of these pre-processing strategies, existing visual SLAM and visual odometry (VO) methods tend to be susceptible to these image degradation and there are hardly any visual SLAM/VO methods that are designed specifically to work robustly under such condition.

The quality of LiDAR scans can also be degraded when facing rain droplets, snowflakes and fog particles in extreme weather. A filtering based approach is proposed in \cite{charron2018noising} to de-noise 3D point cloud scans corrupted by snow before using them for localization and mapping. To mitigate the noisy effects of LiDAR reflection from random rain droplets, \cite{zhang2018robust} proposes ground-reflectivity and vertical features to build a prior tile map, which is used for localization in a rainy weather. In contrast to process 3D LiDAR scans, \cite{aldibaja2016improving} suggests the use of 2D LiDAR images reconstructed and smoothed by Principal Component Analysis (PCA). An edge-profile matching algorithm is then used to match the run-time LiDAR images with a mapped set of LiDAR images for localization. However, these methods are not reliable when the rain, snow or fog is moderate or heavy. The results of LIO-SAM \cite{liosam2020shan}, a LiDAR based odometry and mapping algorithm fused with IMU data, in light snow show that a LiDAR based approach can work to some degree in snow. However, as the snow increases, the reconstructed 3D point cloud map is corrupted to a high degree with random points from the reflection of snowflakes, which reduces the map's quality and its re-usability for localization.

In summary, camera and LiDAR sensors are naturally sensitive to rain, fog and snow. Therefore, attempts to use these sensors to perform localization and mapping tasks in adverse weather are limited.

\begin{figure*}[t]
    \centering
    \begin{subfigure}{0.46\linewidth}
    \includegraphics[width=\linewidth]{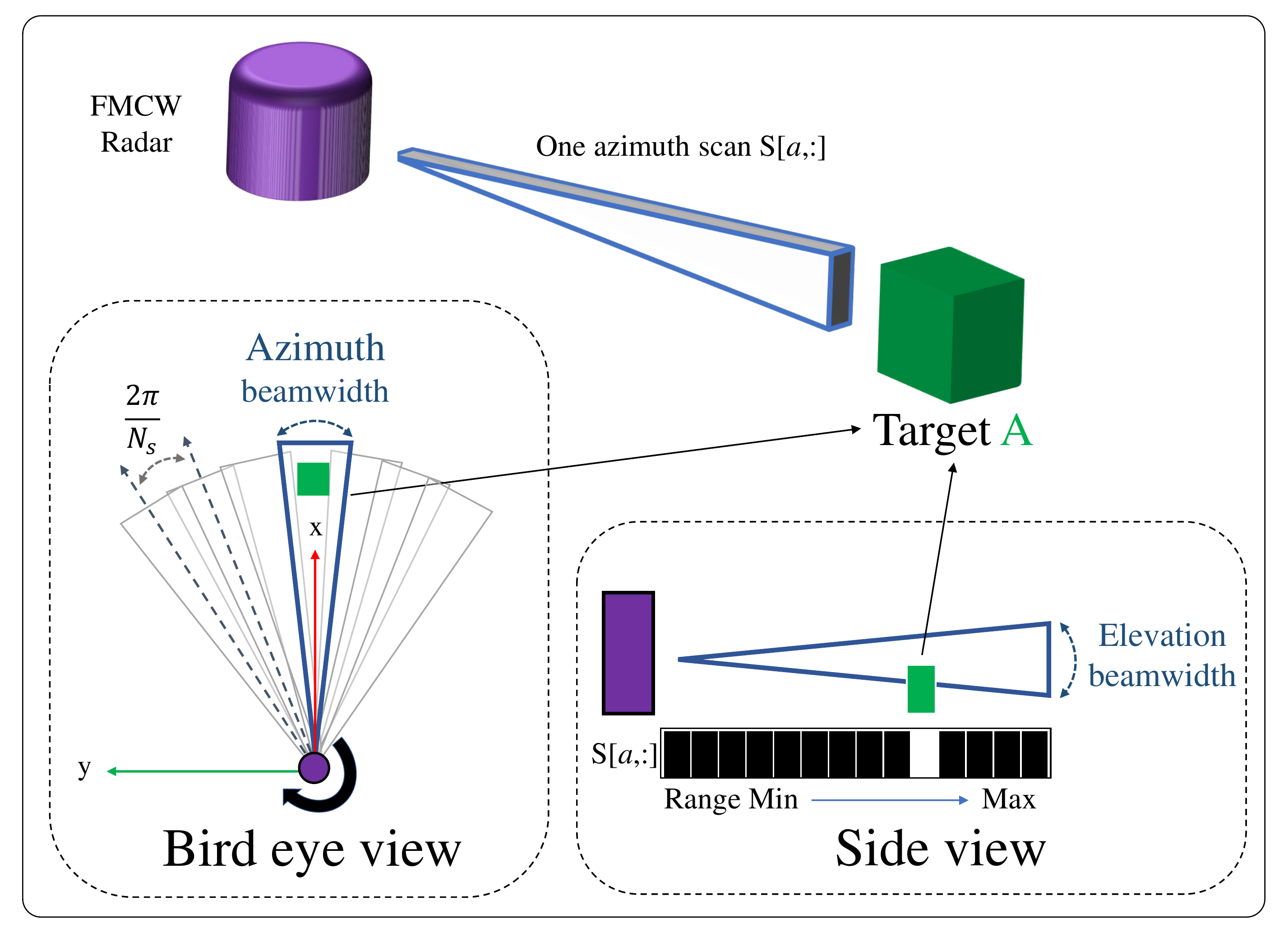}
    \caption{}
    \label{fig:radar_geometry}
    \end{subfigure}
    \begin{subfigure}{0.52\linewidth}
    \includegraphics[width=\linewidth]{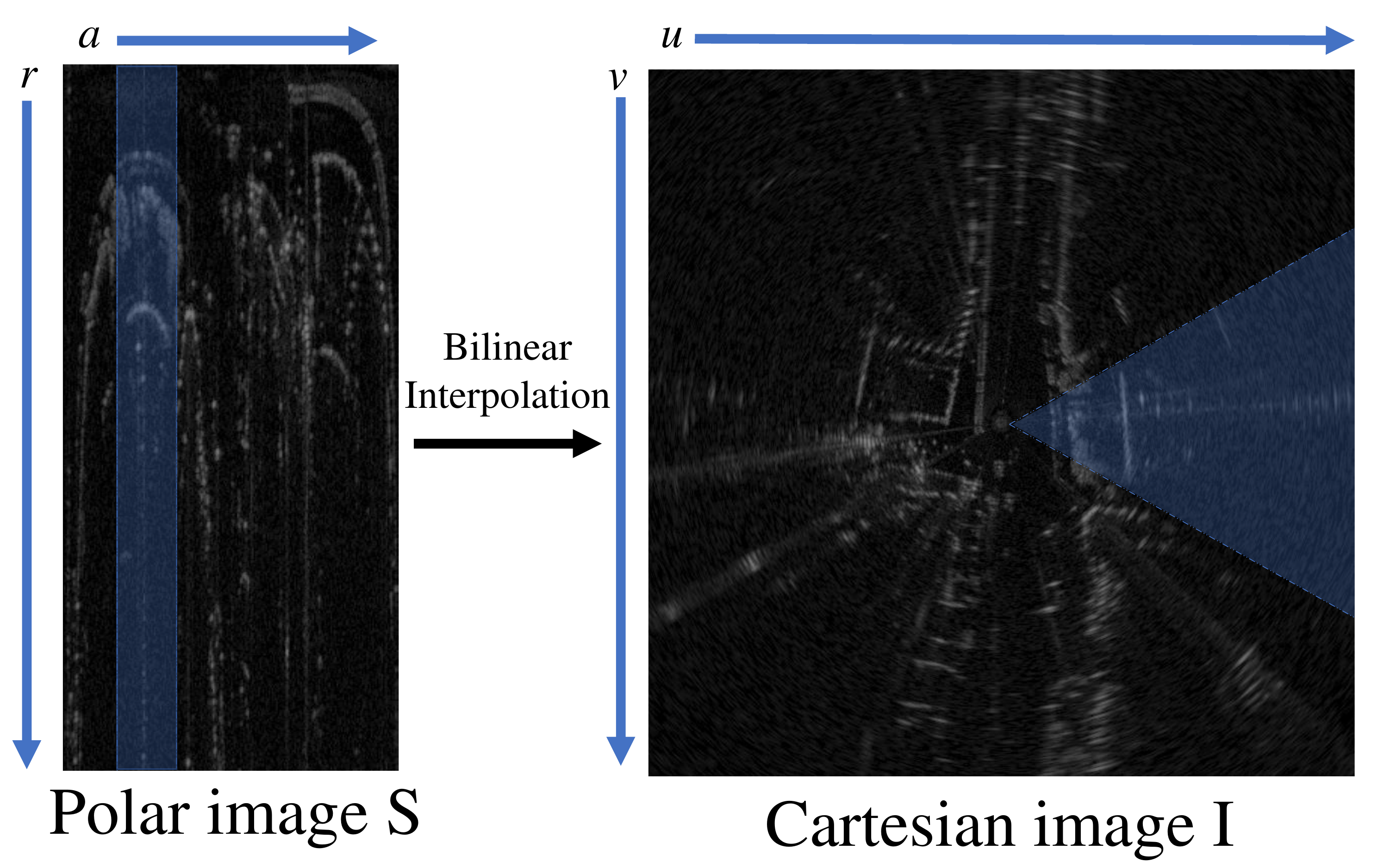}
    \caption{}
    \label{fig:polar_to_cartesian}
    \end{subfigure}
    \caption{Radar sensing and radar image formation. (a): A radar sends a beam with certain azimuth and elevation beamwidths, and the receiver waits for echoes from the target objects. Elevation information, like object height, is usually not retained and collapsed to one point of $\mathbf{S}[a,:]$. (b): Bilinear interpolation from a polar scan to a Cartesian image.}
\end{figure*}

\subsection{Radar based Localization and Mapping}
Using millimeter wave (MMW) radar as a guidance sensor for autonomous vehicle navigation can be traced back two or three decades. An Extended Kalman Filter (EKF) based beacon localization system is proposed by \cite{clark1998autonomous} where the wheel encoder information is fused with range and bearing obtained by radar. One of the first substantial solutions for MMW radar based SLAM is proposed in \cite{dissanayake2001solution}, detecting features and landmarks from radar to provide range and bearing information. \cite{jose2005augmented} further extends the landmark description and formalizes an augmented state vector containing rich absorption and localization information about targets. A prediction model is formed for the augmented SLAM state. Instead of using the whole radar measurement stream to perform scan matching, \cite{chandran2006motion} suggests treating the measurement sequence as a continuous signal and proposes a metric to access the quality of map and estimate the motion by maximizing the map quality. A consistent map is built using a FMCW radar, an odometer and a gyroscope in \cite{rouveure2009high}. Specifically, vehicle motion is corrected using odometer and gyrometer while the map is updated by registering radar scans. Instead of extracting and registering feature points, \cite{checchin2010radar} uses Fourier-Mellin Transform (FMT) to estimate the relative transformation between two radar images. In \cite{vivet2013mobile}, two approaches are evaluated for localization and mapping in a semi-natural environment using only a radar. The first one is the aforementioned FMT computing relative transformation from whole images, while the second one uses a velocity prior to correct a distorted scan \cite{vivet2012radar}. However, both methods are evaluated without any loop closure detection. A landmark based pose graph radar SLAM system proves that it can work in dynamic environments \cite{schuster2016landmark}. Radar is also utilized in \cite{marck2013indoor,mercuri2013practical,park2019radar,almalioglu2020milli,gadd2020look} for GPS-denied indoor mobile robot localization and mapping.

Recently, FMCW radar sensors have been increasingly adopted for vehicles and autonomous robots. \cite{cen2018precise} extract meaningful landmarks for robust radar scan matching, demonstrating the potential of using radar to provide odometry information for mobile vehicles in dynamic city environments. This work is extended with a graph based matching algorithm for data association \cite{cen2019radar}. Radar odometry might fail in challenging environments, such as a road with hedgerows on both sides. Therefore, \cite{aldera2019could} train a classifier to detect failures in the radar odometry and fuse it with an IMU to improve its robustness. Recently, a direct radar odometry method is proposed to estimate relative pose using FMT, with local graph optimization to further boost the performance (\cite{ yspark-2020-icra}. In \cite{burnett2021we}, they study the necessity of motion compensation and Doppler effects on the recent emerging spinning radar for urban navigation.

Deep Learning based radar odometry and localization approaches have been explored in \cite{barnes2019masking,aldera2019fast,UnderTheRadarICRA2020,gadd2020look,tang2020rsl,tang2020self,suaftescu2020kidnapped}. Specifically, in \cite{aldera2019fast} the coherence of multiple measurements is learnt to decide which information should be kept in the reading. In \cite{barnes2019masking}, a mask is trained to filter out the noise from radar data and Fast Fourier Transform (FFT) cross correlation is applied on the masked images to compute the relative transformation. The experimental results show impressive accuracy of odometry using radar. A self-supervised framework is also proposed for robust keypoint detection on Cartesian radar images which are further used for both motion estimation and loop closure detection \cite{UnderTheRadarICRA2020}.

Full radar based SLAM systems are able to reduce drift and generate a more consistent map once a loop is closed. A real-time pose graph SLAM system is proposed in \cite{holder2019real}, which extracts keypoints and computes the GLARE descriptor \cite{himstedt2014large} to identify loop closure. However, the system depends on other sensory information, e.g. rear wheel speed, yaw rates and steering wheel angles.

\subsubsection{Adverse Weather}
Although radar is considered more robust in adverse weather, the aforementioned methods do not directly demonstrate its operation in these conditions. \cite{yoneda2018vehicle} proposes a radar and GNSS/IMU fused localization system by matching query radar images with mapped ones, and tests radar based localization in three different snow conditions: without snow, partially covered by snow and fully covered by snow. It shows that the localization error grows as the volume of snow increases. However they did not evaluate their system during snow but only afterwards. To explore the full potential of FMCW radar in all weathers, our previous work \cite{hong2020radarslam} proposes a feature matching based radar SLAM system and performs experiments in adverse weather conditions without the aid of other sensors. It demonstrates that radar based SLAM is capable of operating even in heavy snow when LiDAR and camera both fail. In another interesting recent work, ground  penetrating radar is used for localization in inclement weather \cite{ort2020autonomous}
This takes a completely different perspective to address the problem. The ground penetrating radar (GPR) is utilized for extracting stable features beneath the ground. During the localization stage, the vehicle needs an IMU, a wheel encoder and GPR information to localize.


In this work, we extend our preliminary results presented in \cite{hong2020radarslam} with a novel motion tracking algorithm optimally compensating motion distortion and an improved loop closure detection. Based on extensive experiments, we also demonstrate a robust and accurate SLAM system operating in extreme weather conditions using radar perception.


\section{Radar Sensing} \label{sec:radar_sensing}
In this section, we describe the working principle of a FMCW radar and its sensing geometry. We also elaborate the challenges of employing a FMCW radar for localization and mapping.

\subsection{Notation}
Throughout this paper, a reference frame $j$ is denoted as $\mathcal{F}_j$ and a homogeneous coordinate of a 2D point $\mathcal{P}_j$ in frame $\mathcal{F}_j$ is defined as $\mathbf{p}_j = [x_j,y_j,1]^\top $. A homogeneous transformation $\mathbf{T}_{i,j} \in \mathbf{SE}(2)$ which transforms a point from the coordinate frame $\mathcal{F}_j$ to $\mathcal{F}_i$ is denoted by a transformation matrix:
\begin{equation}
    \mathbf{T}_{i, j} = \left[\begin{array}{cc}
    \mathbf{R}_{i,j}  & \mathbf{t}_{i,j} \\
    \mathbf{0}          &   1
    \end{array}\right]
\end{equation}
where $\mathbf{R}_{i,j} \in \mathbf{SO}(2)$ is the rotation matrix and $\mathbf{t}_{i,j} \in \mathbb{R}^2$ is the translation vector. Perturbation $\omega \in \mathbb{R}^3$ around the pose $\mathbf{T}_{i,j}$ uses a minimal representation and its Lie algebra representation is expressed as $\omega^{\wedge} \in \mathfrak{se}(2)$. We use the left multiplication convention to define its increment on $\mathbf{T}_{i,j}$ with an operator $\oplus$, i.e.,
\begin{equation} \label{eq:lie_compounding}
\omega \oplus \mathbf{T}_{i,j} = \exp(\omega^{\wedge} \cdot \mathbf{T}_{i,j}
\end{equation}

A polar radar image and its bilinear interpolated Cartesian counterpart are denoted as $\mathbf{S}$ and $\mathbf{I}$, respectively. A point in the Cartesian image $\mathbf{I}$ is represented by its pixel coordinates $\mathbf{P} = [u,v]^\top$.

\subsection{Geometry of a Rotating FMCW Radar}
There are two types of continuous-wave radar: unmodulated and frequency-modulated radars. Unmodulated continuous-wave radar can only measure the relative velocity of targeted objects using the Doppler effect, while a FMCW radar is also able to  measure distances by detecting time shifts and/or frequency shifts between the transmitted and received signals. Some recently developed FMCW radars make use of multiple consecutive observations to calculate targets' speeds so that Doppler processing is strictly required. This improves the processing performance and accuracy of target range measurements.


Assume a radar sensor rotates 360 degrees clockwise in a full cycle with a total of $N_s$ azimuth angles as shown in Fig. \ref{fig:radar_geometry}, i.e., the step size of the azimuth angle is $2\pi/N_s$. For each azimuth angle, the radar emits a beam and collapses the return signal to the point where a target is sensed along a range without considering elevation. Therefore, a radar image is able to provide absolute metric information of distance, different from a camera image which lacks depth by nature. As shown in Fig. \ref{fig:polar_to_cartesian}, given a point $(a,r)$ in a polar image $\mathbf{S}$ where $a$ and $r$ denote its azimuth and range, its homogeneous coordinates $\mathbf{p}$ can be computed by
\begin{equation}
    \mathbf{p} = \begin{bmatrix}
    \mu_p\cdot r\cdot\cos\theta
    \\ \mu_p \cdot r\cdot \sin \theta
    \\ 1 \end{bmatrix}
\end{equation}
where $\theta = -a \cdot 2\pi/ N_s$ is the ranging angle in Cartesian coordinates, and $\mu_p$ (m/pixel) is the scaling factor between the image space and the world metric space. This point on the polar image can also be related to a point on the Cartesian image $\mathbf{I}$ with a pixel coordinate $\mathbf{P} $ by
\begin{equation}
\begin{array}{l}
    u = \frac{w}{2} - \frac{\mu_p}{\mu_c}\cdot r\cdot\sin\theta  \\
    v = \frac{h}{2} - \frac{\mu_p}{\mu_c}\cdot r\cdot\cos\theta
\end{array}
\end{equation}
where $w$ and $h$ are the width and height of the Cartesian image, and $\mu_c$ (m/pixel) is the scale factor between the pixel space and the world metric space used in the Cartesian image. Therefore, the raw polar scan $\mathbf{S}$ can be transformed into a Cartesian space, represented by a grey-scale Cartesian image $\mathbf{I}$ through bilinear interpolation, as shown in Fig. \ref{fig:polar_to_cartesian}.


\subsection{Challenges of Radar Sensing for SLAM} \label{sec:radar_challenges}
Despite the increasingly widespread adoption of radar systems for perception in autonomous robots and in Advanced Driver-Assistance Systems (ADAS), there are still significant challenges for an effective radar SLAM system.

\subsubsection{Coupled Noise Sources.} \label{ssc:noise_source}
As a radio active sensor, radar suffers from multiple sources of noise and clutter, e.g. speckle noise, receiver saturation and multi-path reflection, as shown in Fig. \ref{fig:radar_noise}. Speckle noise is the product of interaction between different radar waves which introduces light and dark random noisy pixels on the image. Meanwhile, multi-path reflection may create ``ghost'' objects, presenting repetitive similar patterns on the image. The interaction of these multiple sources adds another dimension of complexity and difficulty when applying traditional vision based SLAM techniques to radar sensing.

\subsubsection{Discontinuities of Detection.} \label{ssc:persepectiveChanges}
 Radar operates at a longer wavelength than LiDAR, offering the advantage of perceiving beyond the closest object on a line of sight. However, this could become problematic for some key tasks in pose estimation, e.g. frame-to-frame feature matching and tracking, since objects or clutter detected (not detected) in the current radar frame might suddenly disappear (appear) in next frame. As shown in Fig. \ref{fig:discontinuity}, this can happen even during a small positional change. This discontinuity of detection can introduce ambiguities and challenges for SLAM, reducing robustness and accuracy of motion tracking and loop closure.

\subsubsection{Motion Distortion.}

\begin{figure}
    \centering
    \begin{subfigure}{\linewidth}
    \centering
    \includegraphics[width=0.7\linewidth]{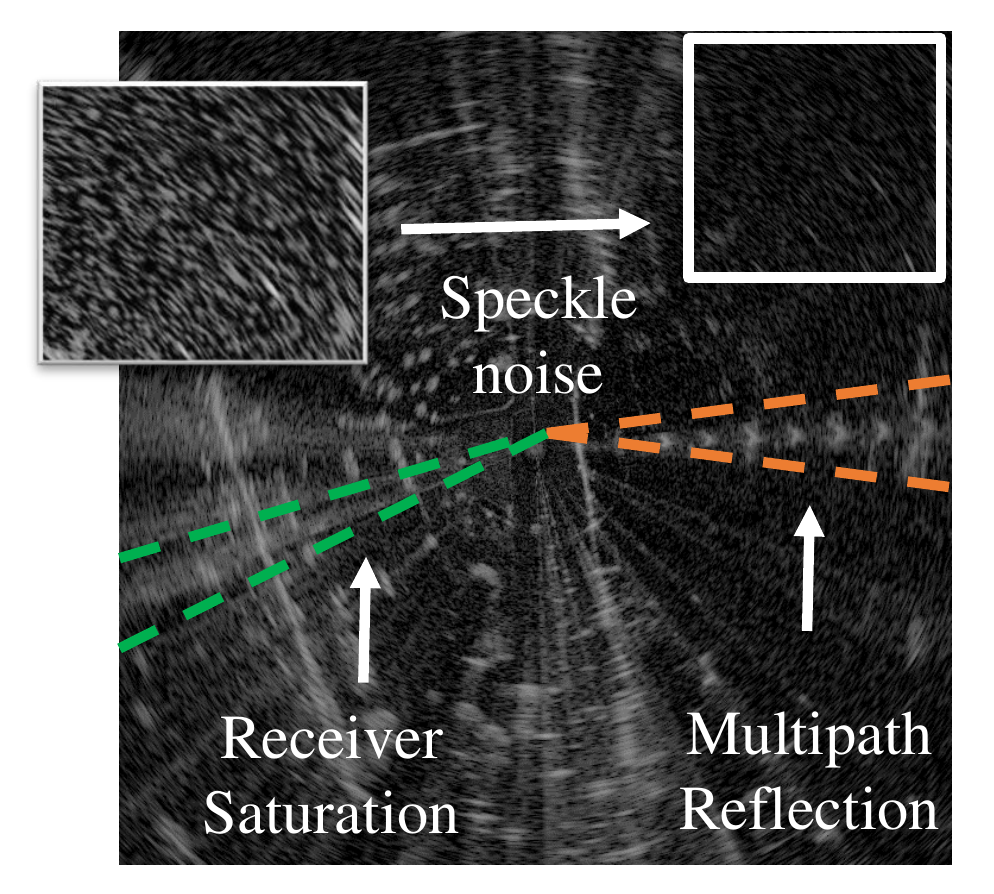}
    \caption{Major types of radar image degradation include speckle noise, receiver saturation and multipath reflection.}
    \label{fig:radar_noise}
    \end{subfigure}
    \begin{subfigure}{\linewidth}
    \includegraphics[width=\linewidth]{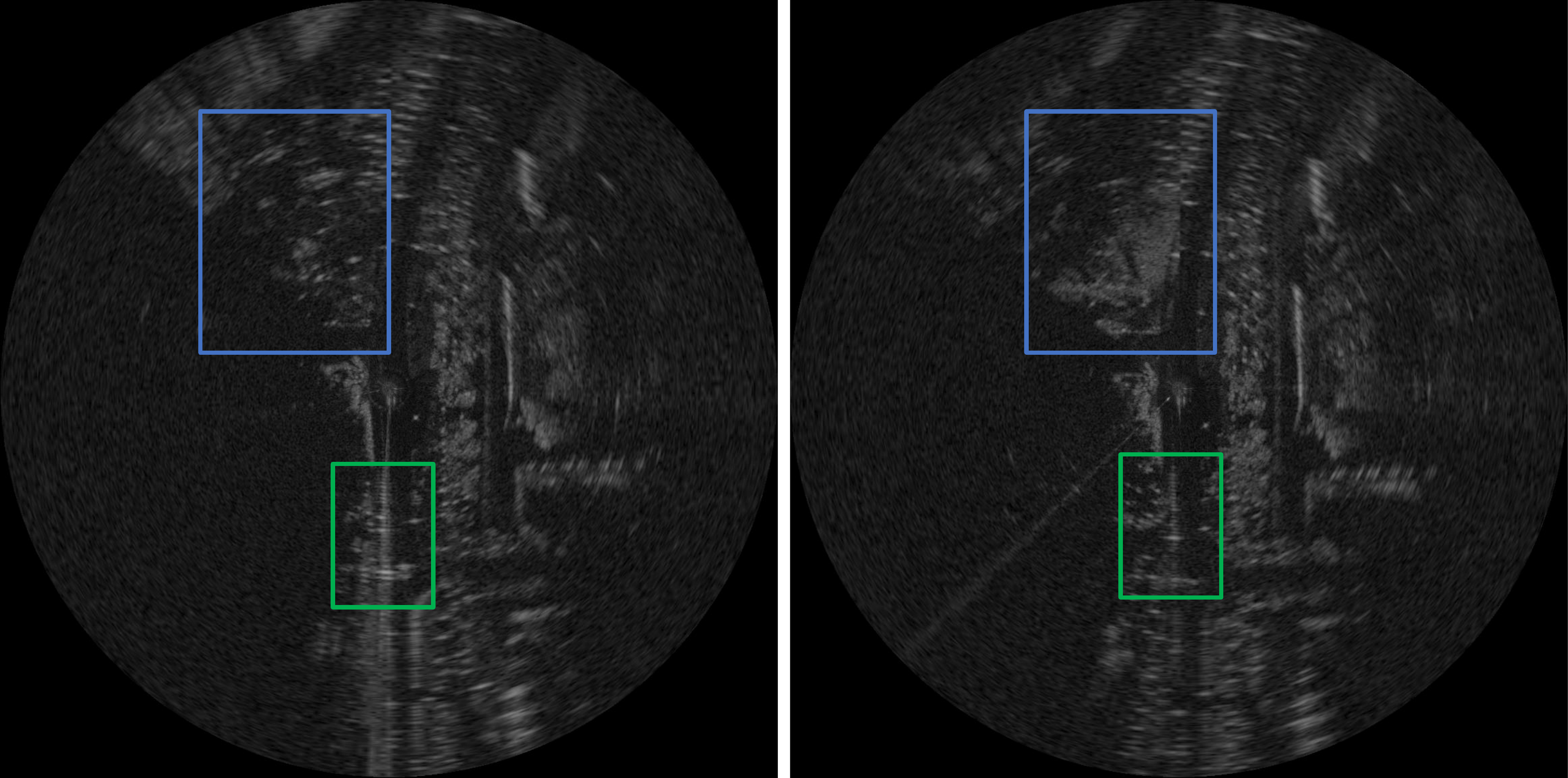}
    \caption{Discontinuity between two consecutive radar frames: two inconsistent detections are highlighted.}
    \label{fig:discontinuity}
    \end{subfigure}
    \begin{subfigure}{\linewidth}
    \includegraphics[width=\linewidth]{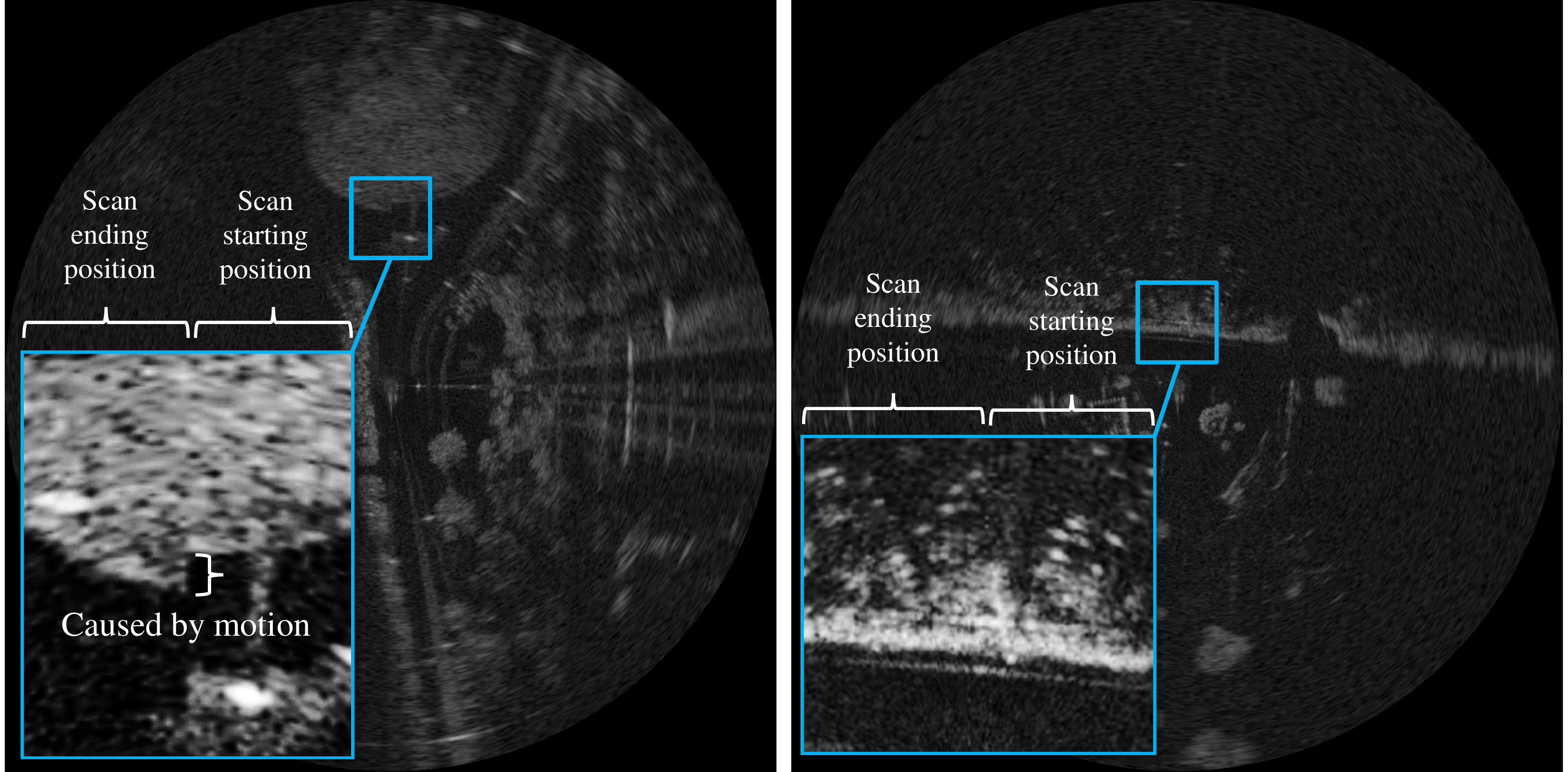}
    \caption{Motion distortion shown on a Cartesian radar image. Left image: the right half of the radar where the azimuth scan starts does not match the left half which belongs to the end of the azimuth scan due to a high speed. Right image: No distortion between the start and end scans while the radar is static.}
    \label{fig:motion_distortion}
    \end{subfigure}
    \caption{Three major types of challenges for radar SLAM.}
\end{figure}

In contrast to camera and LiDAR, current mechanical scanning radar operates at a relatively low frame rate ($4$Hz for our radar sensor). Within a full 360-degree radar scan, a high-speed vehicle can travel several meters and degrees, causing serious motion distortion and discontinuities on radar images, in particular between scans at 0 and 360 degrees. An example in Fig. \ref{fig:motion_distortion} shows this issue on the Cartesian image on the left, i.e. skewed radar detections due to motion distortion. By contrast, there are no skewed detections when it is static. Therefore, directly using these distorted Cartesian images for geometry estimation and mapping can introduce errors. 

In the next sections, we propose an optimization based motion tracking algorithm and a graph SLAM system to handle these challenges.

\begin{figure*}[t]
    \centering
    \includegraphics[width=\linewidth]{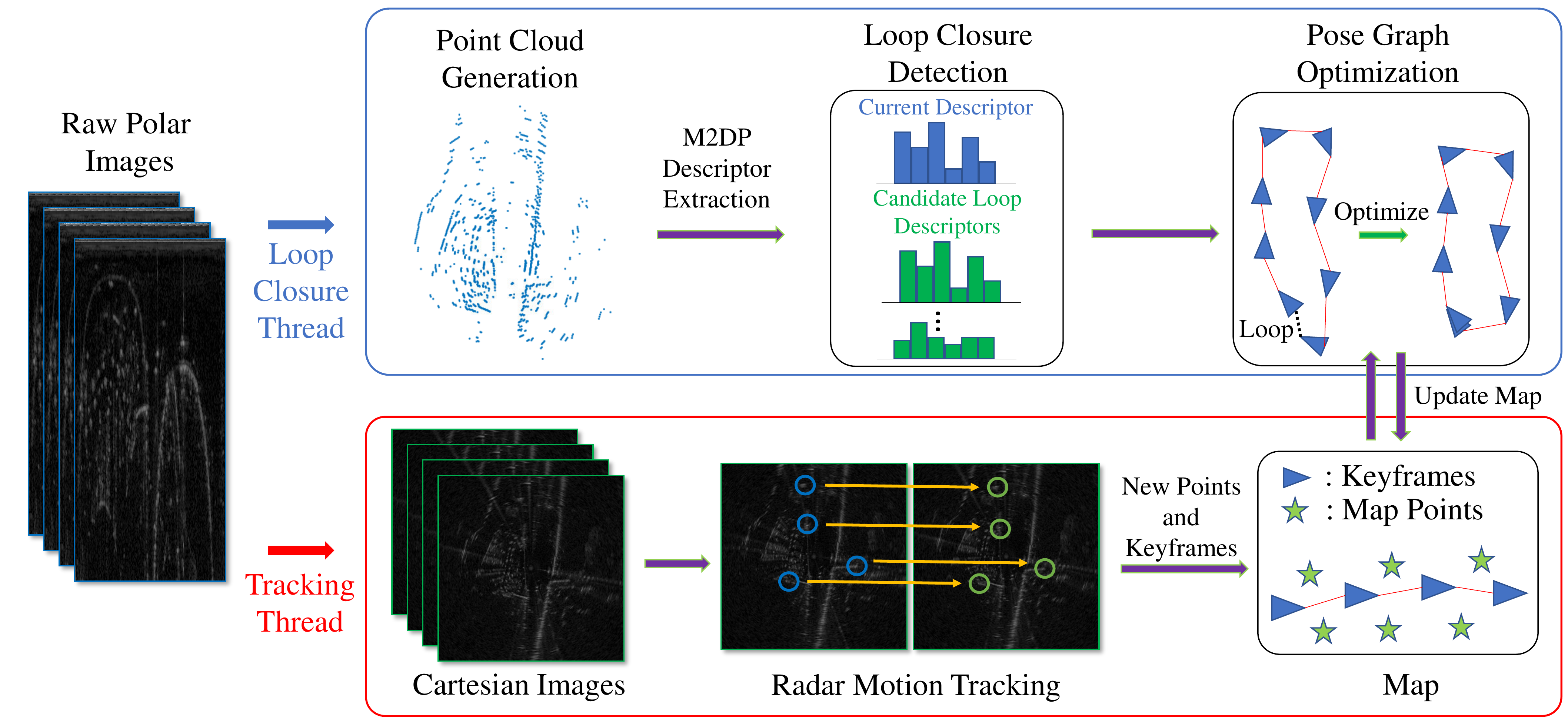}
    \caption{System diagram.}
    \label{fig:system_diagram}
\end{figure*}

\section{System Overview} \label{sec:system_overview}



The proposed system includes radar motion tracking, loop closure detection and pose graph optimization. The system, shown in Fig. \ref{fig:system_diagram},
is divided into two threads.
The main thread is the tracking thread which takes the Cartesian images as input, tracks the radar motion and creates new points and keyframes for mapping. The other parallel thread takes the polar images as input and is responsible for generation of the dense point cloud and computation of descriptors for loop closure detection. Finally, once a loop is detected, it performs pose graph optimization to correct the drift induced by tracking before updating the map.

\section{Radar Motion Tracking} \label{sec:tracking_model}



This section describes the proposed radar motion tracking algorithm, which includes feature detection and tracking, graph based outlier rejection and radar pose tracking with optimal motion distortion compensation.

\subsection{Feature Detection and Tracking} \label{sec:tracking}

For each radar Cartesian image $\mathbf{I}_j$, we first detect keypoints purely using a blob detector based on a Hessian matrix. Keypoints with Hessian responses larger than a threshold are selected as candidate points. The candidate points are then selected based on the adaptive non-maximal suppression (ANMS) algorithm \cite{bailo2018efficient}, which selects points that are homogeneously spatially distributed. Instead of using a descriptor to match keypoints as in \cite{hong2020radarslam}, we track them between frames $\mathbf{I}_{j-1}$ and $\mathbf{I}_j$ using the KLT tracker \cite{lucas1981iterative}.
\subsection{Graph based Outlier Rejection}

It is inevitable that some keypoints are detected and tracked on dynamic objects, e.g., cars, cyclist and pedestrians, and on radar noise, e.g., multi-path reflection. 
We leverage the absolute metrics that radar images directly provide to form geometric constraints used for detecting and removing these outliers.

\begin{figure}[h]
    \centering
    \includegraphics[width=0.9\linewidth]{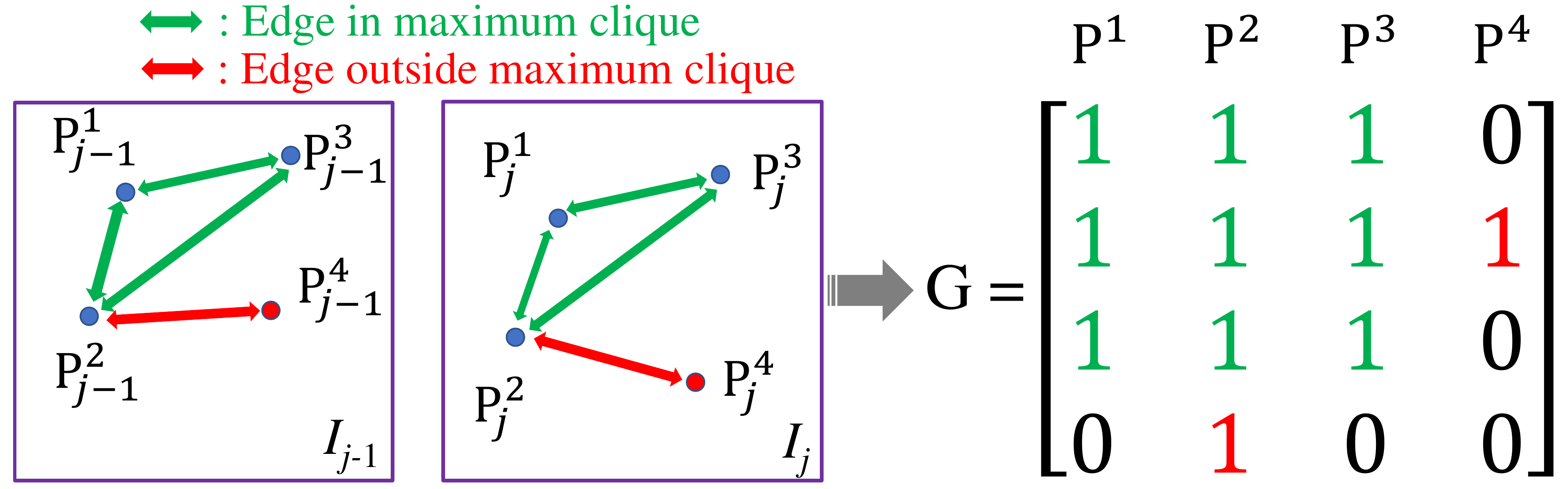}
    \caption{Pairwise constraint: the pairwise constraint is checked by comparing the edge length difference between the points. The maximum clique is found through the consistency matrix G. Points that are not within the maximum clique are considered as outliers, e.g., $\mathbf{P}^4$.}
    \label{fig:pariwise_constriant}
\end{figure}

We apply a graph based outlier rejection algorithm described in \cite{howard2008real}. We impose a pairwise geometric consistency constraint on the tracked keypoint pair based on the fact that they should follow a similar motion tendency. The assumption is that most of the tracked points are from static scene data. Therefore, for any two pairs of keypoint matches between the current $\mathbf{I}_j$ and the last $\mathbf{I}_{j-1}$ radar frames, they should satisfy the following pairwise constraint:
\begin{equation}
  \left|\lVert\mathbf{P}_{j-1}^m - \mathbf{P}_{j-1}^n\rVert_2 - \lVert\mathbf{P}_j^m - \mathbf{P}_j^n\lVert_2\right| < \delta_c
  \label{eq:pairwise_constraint}
\end{equation}
where $\left|\cdot\right|$ is the absolute operation, $\lVert\cdot\lVert_2$ is the Euclidean distance, $\delta_c$ is a small distance threshold, and $\mathbf{P}_{j-1}^m$, $\mathbf{P}_{j-1}^n$, $\mathbf{P}_{j}^m$ and $\mathbf{P}_{j}^n$ are the pixel coordinates of two pairs of tracked points between $\mathbf{I}_{j-1}$ and $\mathbf{I}_j$. 
 Hence, $\mathbf{P}_{j-1}^m$ and $\mathbf{P}_{j}^m$ denote a pair of associated points while $\mathbf{P}_{j-1}^n$ and $\mathbf{P}_{j}^n$ is another pair, see Fig. \ref{fig:pariwise_constriant} for an intuitive example. A consistency matrix $\mathbf{G}$ is then used to represent all the associations that satisfy this pairwise consistency. If a pair of associations satisfies this constraint, the corresponding entry in $\mathbf{G}$ is set as 1 shown in Fig. \ref{fig:pariwise_constriant}. Finding the maximum inlier set of all matches that are mutually consistent is equivalent to deriving the maximum clique of a graph represented by $\mathbf{G}$, which can be solved efficiently using \cite{konc2007improved}. Once the maximum inlier set is obtained, it is used to compute the relative transformation $\mathbf{T}_{j-1,j}$, which transforms a point from local frame $j$ to local frame $j-1$ using Singular Value Decomposition (SVD) \cite{challis1995procedure}. Given $\mathbf{T}_{j-1,j}$ and the fixed radar frame rate, an initial guess of current velocity $\mathbf{v}_j = [v_x,v_y,v_{\theta}]^\top\in \mathbb{R}^3 $ can be computed for the motion compensation tracking model.

\subsection{Motion Distortion Modelling}

After the tracked points are associated, they can be used to estimate the motion. However, since the radar scanning rate is slow, they tend to suffer from serious motion distortion as discussed in Section \ref{sec:radar_challenges}. This can dramatically degrade the accuracy of motion estimation, which is different from most of the vision and LiDAR based methods. Therefore, we explicitly model and compensate for motion distortion in radar pose tracking using an optimization approach.

\begin{figure}[t]
    \centering
    \includegraphics[width=\linewidth]{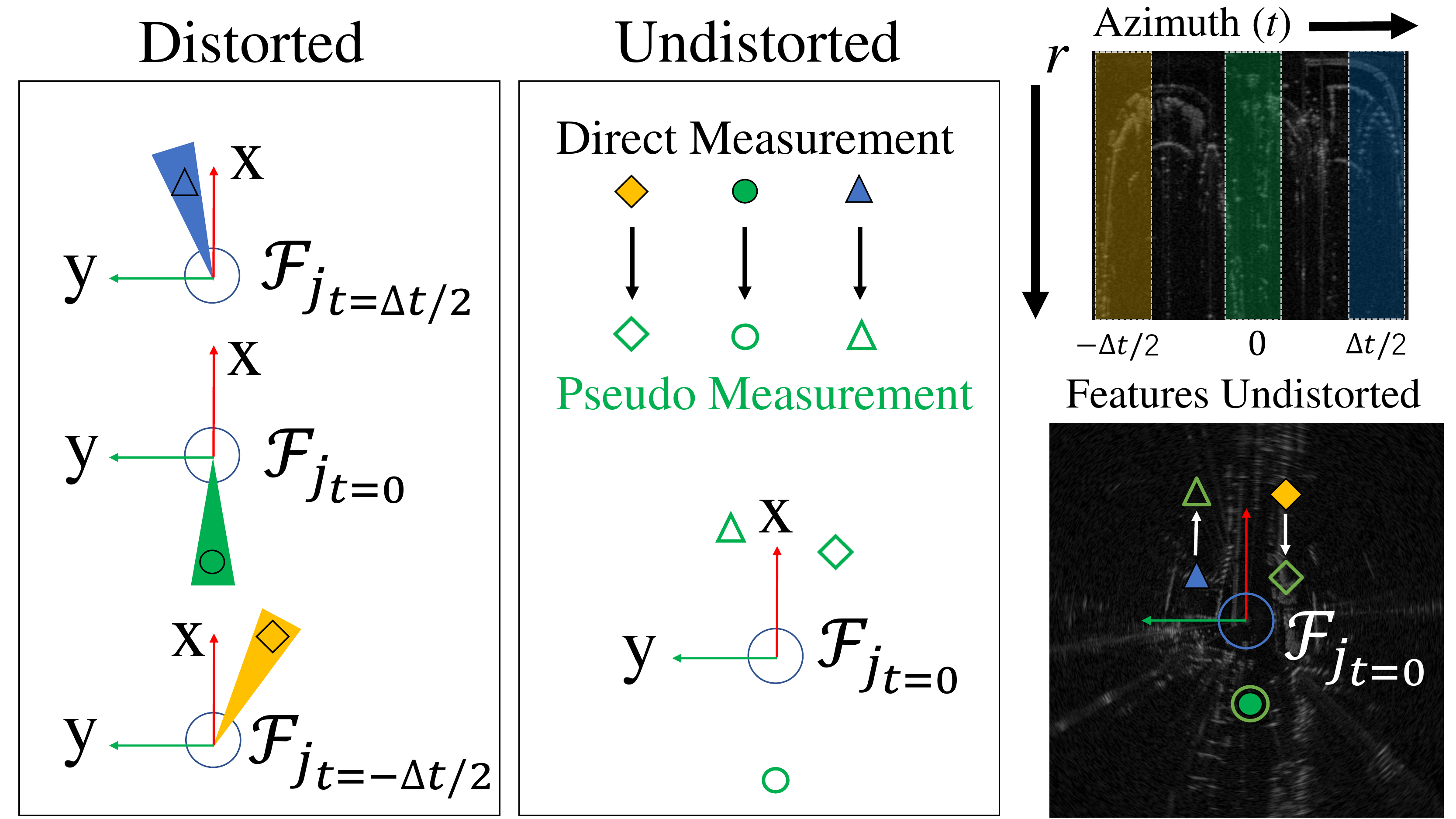}
    \caption{Motion modelling to remove distortion. (Left): For each azimuth angle in a single scan, a radar detection is observed within frame $\mathcal{F}_{j_t}$ while moving. (Middle): All the detections are projected onto frame $\mathcal{F}_{j_0}$ using the optimized motion model, compensating for motion distortion. (Right): Corresponding azimuth scans of the detections shown on a polar image and the positional changes of detections without distortion in a Cartesian image. All these compensated detections are within frame $\mathcal{F}_{j_{0}}$.}
    \label{fig:motion_undistortion_model}
\end{figure}

Assume a full polar radar scan $\mathbf{S}_j$ takes $\Delta t$ seconds to finish. Denote $\mathbf{T}_{w, j}$ as the pose of radar scan $\mathbf{S}_j$ in the world coordinate frame $\mathcal{F}_w$ and $\mathbf{T}_{j,j_t}$ as the pose of the radar scan in the local frame $\mathcal{F}_{j}$ while capturing its azimuth beam at time $t \in [-\Delta t/2, \Delta t/2]$. Without losing generality, we compensate the motion distortion relative to the central azimuth beam at $t=0$, i.e., $\mathcal{F}_{j_{0}}$ defines the local coordinate frame $\mathcal{F}_{j}$ of $\mathbf{S}_j$. The motion distortion model is designed to correct detections on each beam of a radar scan, i.e. optimally estimating the detections on an undistorted radar image as shown in Fig. \ref{fig:motion_undistortion_model}.

The radar pose in the world coordinate frame while capturing an azimuth scan at time $t$ can be obtained by
\begin{equation} \label{eq:compensate_point}
 \mathbf{T}_{w,j_t} = \mathbf{T}_{w,j}  \mathbf{T}_{j,j_t}.
\end{equation}
Consider a constant velocity model in a full scan, we can compute the relative transformation $\mathbf{T}_{j,j_t}$ given the velocity $\mathbf{v}_j$, i.e
\begin{equation}
    \mathbf{T}_{j,j_t} = \exp\left((\mathbf{v}_jt)^{\wedge} \right) = \left[\begin{array}{ccc}
    \cos(v_{\theta}  t) &  -\sin(v_{\theta}  t) & v_x  t \\
    \sin(v_{\theta}  t) &   \cos(v_{\theta}  t) & v_y  t \\
    0          &  0          &   1
    \end{array}\right]
\end{equation}
where $\exp()$ is the matrix exponential map and $\wedge$ is the operation to transform a vector to a matrix.
If the $i$th keypoint $\mathbf{p}^i_w$ in the world frame $\mathcal{F}_w$ is observed as $\mathbf{p}^i_{j_t}$ in the azimuth scan at time $t$, its motion compensated location is then
\begin{equation}
    \mathbf{p}^i_{j_0} =  \mathbf{T}_{j,j_t}  \mathbf{p}^i_{j_t}.
\end{equation}
In other words, $\mathbf{p}^i_{j_0}$ is the compensated location of $\mathbf{p}^i_w$ in the local frame $\mathcal{F}_{j_{0}}$.
Therefore, the feature residual between the locally observed and estimated (after motion compensation) locations of this $i$th keypoint can be computed as:
\begin{equation}
    \mathbf{e}^i_{p} = \rho_c{(\mathbf{T}^{-1}_{w,j}  \mathbf{p}^i_w - \mathbf{p}^i_{j_0}}
\end{equation}
where $\rho_c$ is the Cauchy robust cost function used to account for perspective changes described in \ref{ssc:persepectiveChanges}.

\begin{figure}[h]
    \centering
    \begin{subfigure}{0.49\linewidth}
    \includegraphics[width=\linewidth]{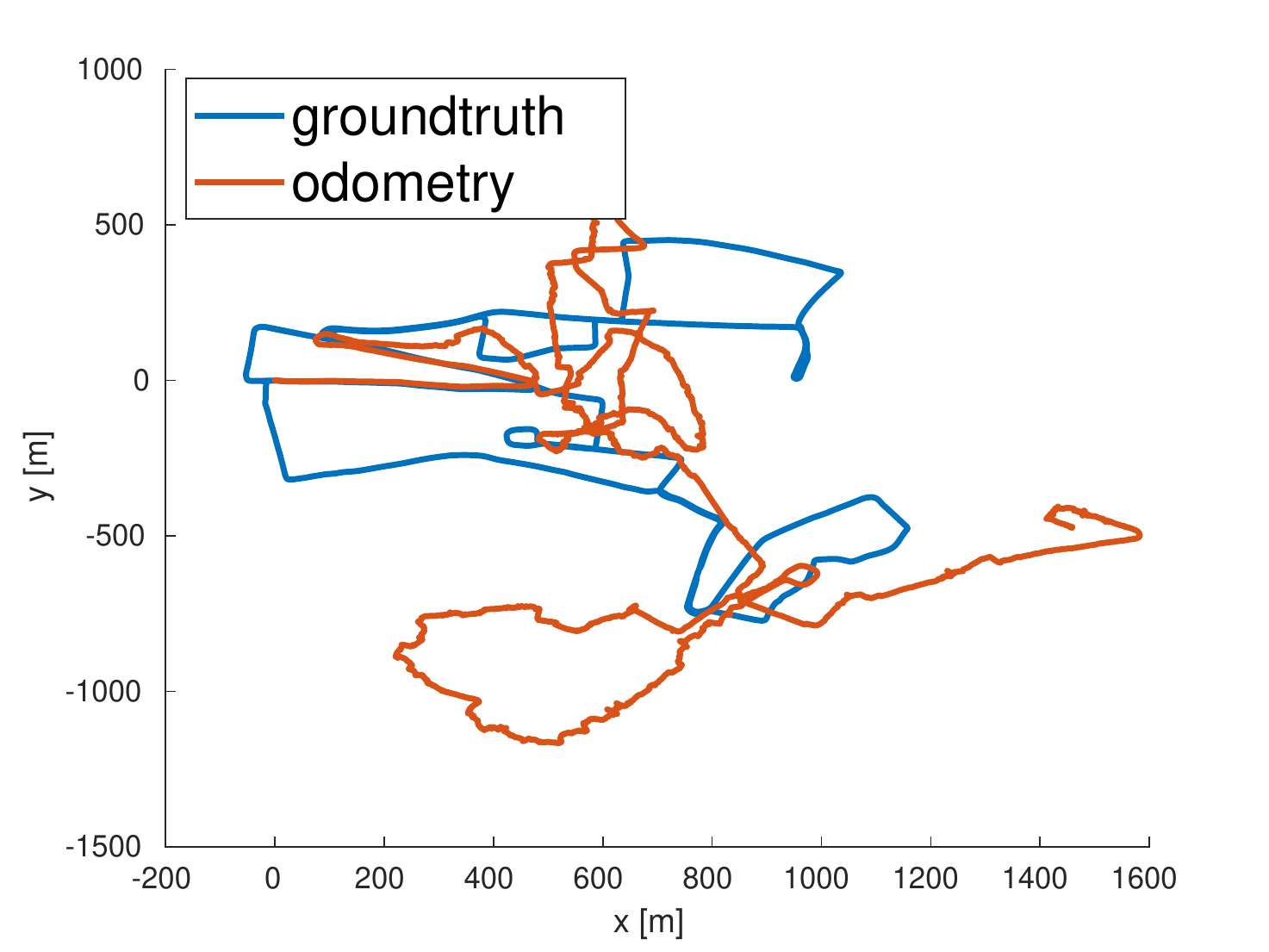}
    \caption{Without $\mathbf{e}_v$}
    \end{subfigure}
    \begin{subfigure}{0.49\linewidth}
    \includegraphics[width=\linewidth]{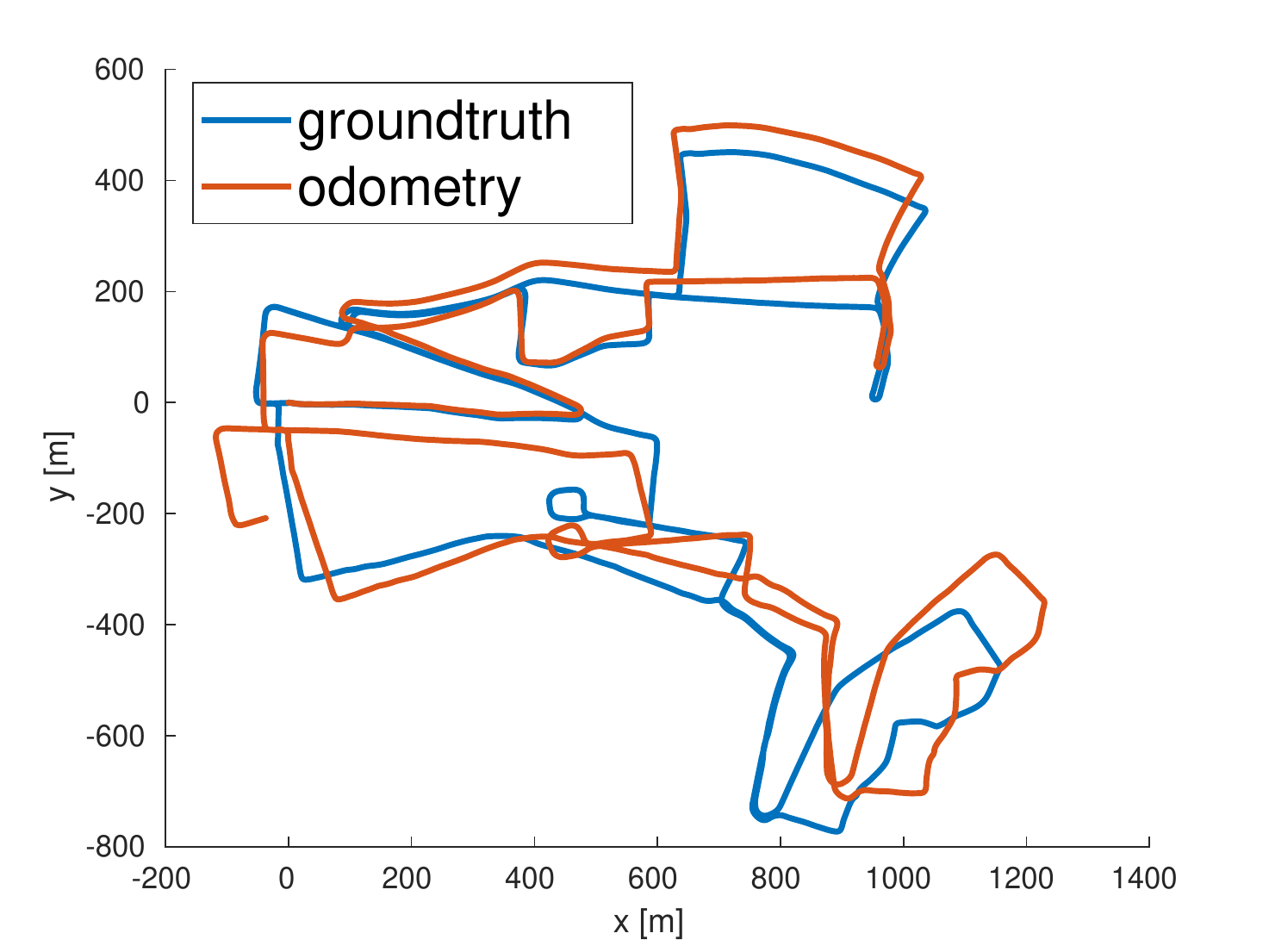}
    \caption{With $\mathbf{e}_v$}
    \end{subfigure}
    \caption{Odometry trajectories without and with the residual term $e_v$ in the optimization. Without the residual term $e_v$ to correlate pose with velocity, we might obtain an arbitrary solution for pose and velocity. To stabilize the optimization, $e_v$ is needed.}
    \label{fig:velocity_prior}
\end{figure}



\subsection{Optimal Motion Compensated Radar Pose Tracking}
Radar pose tracking aims to find the optimal radar pose $\mathbf{T}_{w,j}$ and the current velocity $\mathbf{v}_j$ while considering the motion distortion. In order to ensure smooth motion dynamics, a velocity error $\mathbf{e}_{v}$ is also introduced as a velocity prior term:
\begin{equation}
    \mathbf{e}_{v} = \mathbf{v}_j - \mathbf{v}_{j,prior}
\end{equation}
where $\mathbf{v}_{j,prior}$ is a prior on the current velocity which is parameterized as
\begin{equation}
    \mathbf{v}_{j,prior} = \frac{\log(\left(\mathbf{T}_{w,j-1}\right)^{-1}\mathbf{T}_{w,j}^{\vee}}{\Delta t}
\end{equation}
Here, $\vee$ is the operation to convert a matrix to a vector. This velocity prior term establishes a constraint on velocity changes by considering the previous pose $\mathbf{T}_{w,j-1}$. This prior is crucial to stabilize the optimization. The results with and without this prior are compared in Fig. \ref{fig:velocity_prior}. Therefore, the pose tracking optimizes the velocity $\mathbf{v}_j$ and the current radar pose $\mathbf{T}_{w,j}$ by minimizing the cost function including the feature residuals of all the $N$ keypoints tracked in $\mathbf{S}_j$ and the velocity prior, i.e.,
\begin{equation}
    \mathbf{v}_j^*, \mathbf{T}_{w,j}^* = \argmin_{\mathbf{v}_j, \mathbf{T}_{w,j}} \sum_{i=1}^N {\mathbf{e}^i_{p}}^T\mathbf{\Lambda}^i_p \mathbf{e}^i_{p} + \mathbf{e}_{v}^T \mathbf{\Lambda}_v\mathbf{e}_{v}
    \label{eq:opt_cost}
\end{equation}
where $\mathbf{\Lambda}^i_p$ and $\mathbf{\Lambda}_v$ are the information matrices of the keypoint $i$ and the velocity. 

\begin{figure}
    \includegraphics[width=\linewidth]{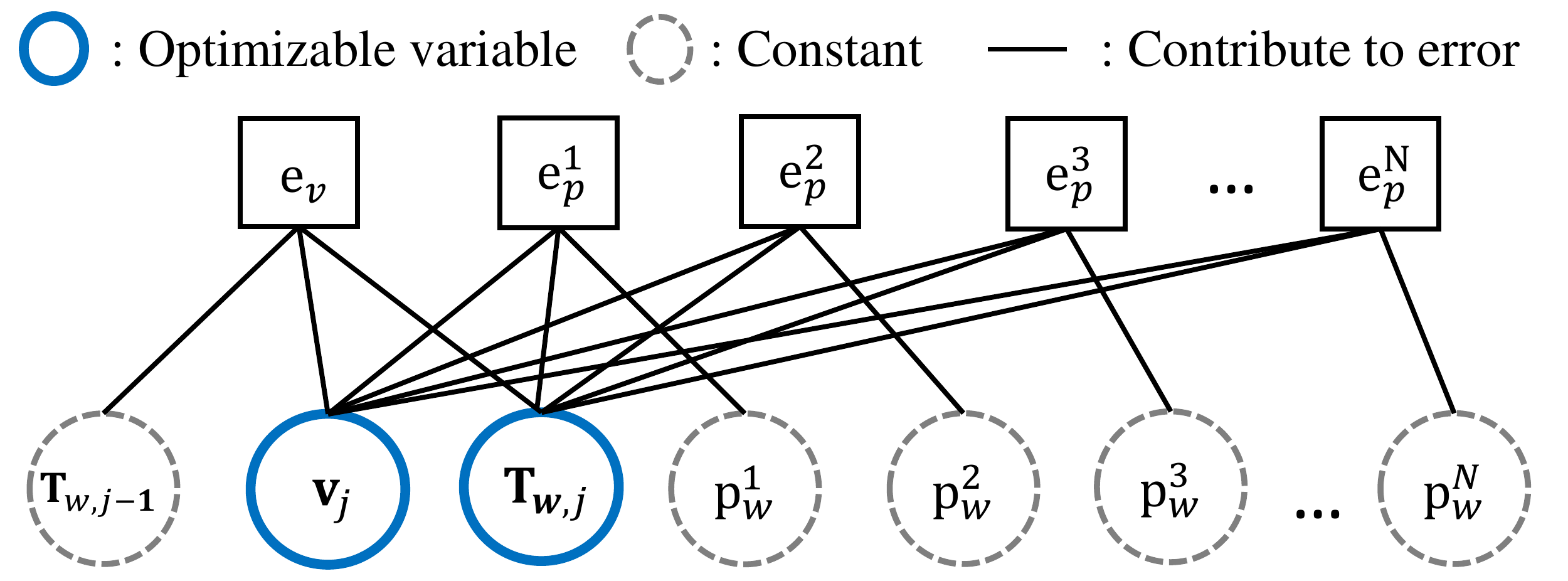}
    \caption{Factor graph for the motion compensation model.}
    \label{fig:factor_graph}
\end{figure}

By formulating a state variable $\Theta$ containing all the variables to be optimized, i.e. the velocity $\mathbf{v}_j$ and the current radar pose $\mathbf{T}_{w,j}$, and denoting $\mathbf{e}(\Theta)$ as a generic residual block of either $\mathbf{e}^i_{p}$ or $\mathbf{e}_{v}$, $\Theta$ can be solved in an optimization whose factor graph representation is shown in Fig. \ref{fig:factor_graph}. The optimization problem is then framed as finding the minimum of the weighted Sum of Squared Errors cost function:
\begin{equation} \label{eq:total_loss}
    \mathbf{F}(\Theta) = \sum_{i}\mathbf{e}^T_i(\Theta) \mathbf{W}_i\mathbf{e}_i(\Theta)
\end{equation}
\begin{equation}
    \Theta^* = \argmin_{\Theta} \mathbf{F}(\Theta)
\end{equation}
where $\mathbf{W}_i$ is a symmetric positive definite weighting matrix.



The total cost in \ref{eq:total_loss} is minimized using the Levenberg–Marquardt algorithm. With the initial guess $\Theta$, the residual $\mathbf{e}_i(\Delta\Theta \boxplus \Theta)$ is approximated by its first order Taylor expansion around the current guess $\Theta$:
\begin{equation} \label{eq:taylor_expansion}
    \mathbf{e}_i(\Delta\Theta \boxplus \Theta) \simeq \mathbf{J}_i\Delta\Theta \boxplus \mathbf{e}_i
\end{equation}
where $\mathbf{J}_i$ is the Jacobian of $\mathbf{e}_i$ evaluated with the current parameters $\Theta$. $\boxplus$ represents an addition operator for the velocity variable $\mathbf{v}_j$ and a $\oplus$ operation for the pose $\mathbf{T}_{w,j}$, as defined in Eq. \ref{eq:lie_compounding}. $\mathbf{J}_i$ can be decomposed into two parts with respect to the velocity and the pose:
\begin{equation}
    \mathbf{J}_i = [\mathbf{J_{v}},\mathbf{J_{p}}]
\end{equation}
i.e., the Jacobian part with respect to the velocity as
\begin{equation}
    \mathbf{J_{v}} = \frac{\partial \mathbf{e}_i(\Theta)}{\partial  (\mathbf{v}_jt)}  \frac{\partial  (\mathbf{v}_jt)}{\partial  \mathbf{v}_j}
\end{equation}
and the Jacobian with respect to the pose as
\begin{equation}
    \mathbf{J_{p}} = \frac{\partial \mathbf{e}_i(\Theta)}{\partial  (\mathbf{T}_{w,j}}
\end{equation}
which is equivalent to computing the Jacobian with respect to zero perturbation $\Delta\Theta_\mathbf{p}$ around the pose \cite{macroLie}, i.e.
\begin{equation}
     \frac{\partial \mathbf{e}_i(\Theta)}{\partial  (\mathbf{T}_{w,j}} =
     \frac{\partial \mathbf{e}_i(\Delta\Theta \boxplus \Theta)}{\partial \Delta\Theta_\mathbf{p}}
\end{equation}
The Levenberg–Marquardt algorithm computes a solution $\Delta\Theta$ at each iteration such that it minimizes the residual function $\mathbf{e}_i(\Delta\Theta \boxplus \Theta)$. We refer the reader to \cite{kummerle2011g} for further details on the optimization process.

\subsection{New Point Generation}
After tracking the current radar scan $\mathbf{S}_j$, the total number of successfully tracked keypoints is checked to decide whether new keypoints should be generated. If it is below a certain threshold, new keypoints are extracted and added for tracking if they are located in image grids whose total numbers of keypoints are low.
Once a new keypoint $\mathbf{p}^i_{j_t}$ is associated with the current frame $\mathbf{S}_j$, its global position $\mathbf{p}_w$ can be derived from
\begin{equation}
    \mathbf{p}_w = \mathbf{T}^*_{w,j}\exp\left((\mathbf{v}_j^* t)^{\wedge} \right)\mathbf{p}^i_{j_t}
\end{equation}
using its direct observation of $\mathbf{p}^i_{j_t}$ (with distortion). $\mathbf{T}^*_{w,j}$ and $\mathbf{v}_j^*$ are the optimal radar pose and velocity derived in Eq. \ref{eq:opt_cost}.

\subsection{New Keyframe Generation}
To scale the system in a large-scale environment, we use a pose-graph representation for the map with each node parameterized by a keyframe. Each keyframe which contains a velocity and a pose is connected with its neighbouring keyframes using its odometry derived from the motion tracking. The keyframe generation criterion is similar to that introduced in \cite{murORB2}, i.e., the current frame is created as a new keyframe if its distance with respect to the last keyframe is larger than a certain threshold or its relative yaw angle is larger than a certain threshold. A new keyframe is also generated if the number of tracked points is less than a certain number.

\section{Loop Closure and Pose Graph Optimization} \label{sec:loop_pose_graph}

Robust loop closure detection is critical to reduce drift in a SLAM system. Although the Bag-of-Words model has proved efficient for visual SLAM algorithms, it is not adequate for radar based loop closure detection due to three main reasons: first, radar images have less distinctive pixel-wise characteristics compared to optical images, which means similar feature descriptors can repeat widely across radar images causing a large number of incorrect feature matches; second, the multi-path reflection problem in radar can introduce further ambiguity for feature description and matching; third, a small rotation of the radar sensor may produce tremendous scene changes, significantly distorting the histogram distribution of the descriptors. On the other hand, radar imagery encapsulates valuable absolute metric information, which is inherently missing for an optical image. Therefore, we propose a loop closure technique which captures the geometric scene structure and exploits the spatial signature of reflection density from radar point clouds.

\subsection{Point Cloud Generation}

\begin{algorithm}[!t]
  \SetAlgoLined
  \textbf{Input:} Radar polar scan $\mathbf{S} \in \mathbb{R}^{m \times n} $\;
  \textbf{Output:} Point Cloud $\mathbf{C} \in \mathbb{R}^{z \times 2}$\;
  \textbf{Parameters:} Minimum peak prominence $\delta_{p}$ and minimum peak distance $\delta_{d}$\;
  Initialize empty point cloud set $\mathbf{C}$\;
      \For{$i \gets 1$ to $m$}{
      $Q^{k \times 1}$ $\leftarrow$ findPeaks($\mathbf{S}[i,:]$, $\delta_{p}$,  $\delta_{d}$)\;
      $(\mu, \sigma)$ $\leftarrow$ meanAndStandardDeviation($Q^{k \times 1}$)\;
          \For{each peak $q$ in $Q$}{
              \If{$q \geq (\mu + \sigma)$}
              {
                  $p$ $\leftarrow$ transformPeakToPoint($q$, $i$)\;
                  Add the point $p$ to $\mathbf{C}$\;
              }
          }
      }
      \caption{Radar Polar Scan to Point Cloud Conversion}
      \label{algo:point_cloud_generation}
\end{algorithm}
\begin{figure}
    
    \begin{subfigure}{0.32\linewidth}
    {\includegraphics[width=\linewidth]{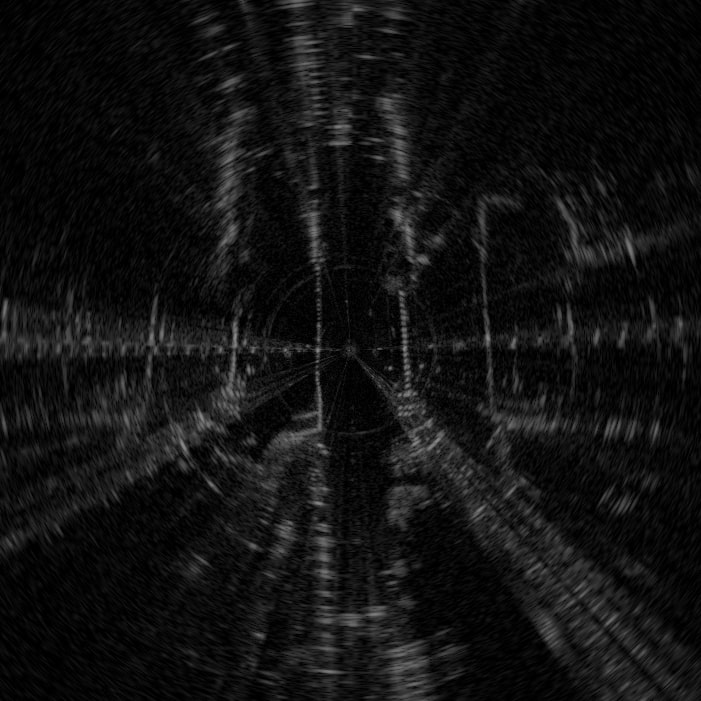}}
    \end{subfigure}
    \begin{subfigure}{0.32\linewidth}
    {\includegraphics[width=\linewidth]{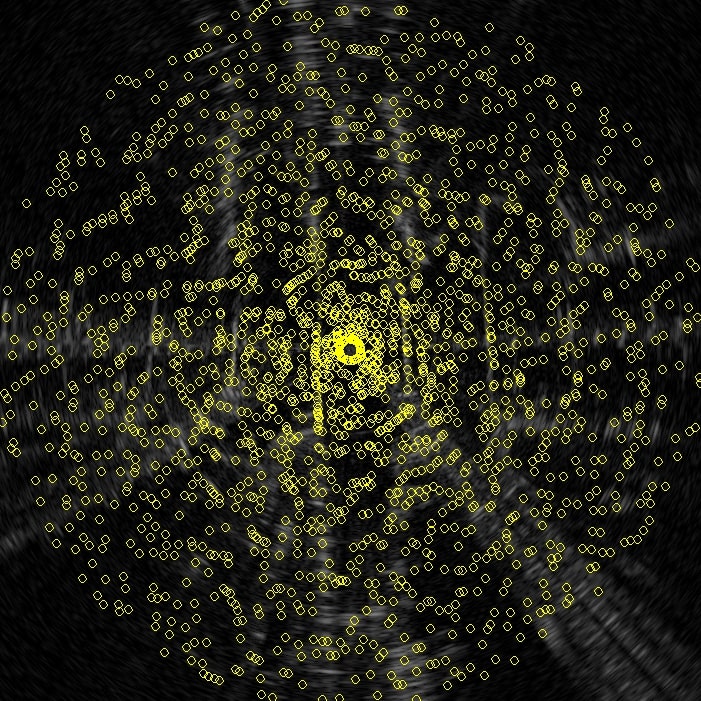}}
    \end{subfigure}
    \begin{subfigure}{0.32\linewidth}
    {\includegraphics[width=\linewidth]{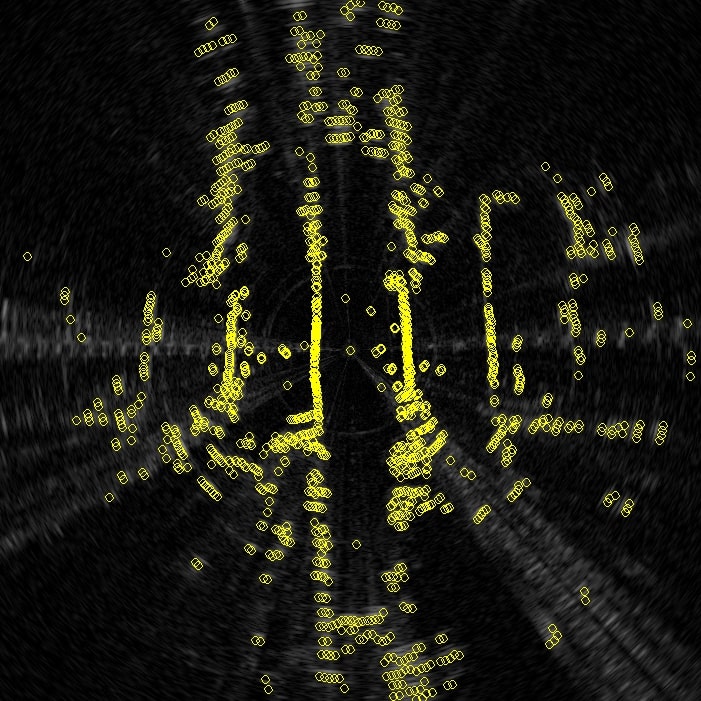}}
    \end{subfigure}

    \caption{Peak detection in a radar scan. Left: Original Cartesian image. Middle: Peaks (in yellow) detected using a local maxima algorithm. Note that a great amount of peaks detected are due to speckle noise. Right: Peaks detected using the proposed point cloud extraction algorithm which preserves the environmental structure and suppresses detections from multi-path reflection and speckle noise.}
    \label{fig:loop_closure}
\end{figure}

Considering the challenges of radar sensing in Section \ref{sec:radar_challenges}, we want to separate true targets from the noisy measurements on a polar scan. An intuitive and naive way would be to detect peaks by finding the local maxima from each azimuth reading. However, as shown in Fig. \ref{fig:loop_closure}, the detected peaks can be distributed randomly across the whole radar image, even for areas without a real object, due to the speckle noise described in Section \ref{ssc:noise_source}. Therefore, we propose a simple yet effective point cloud generation algorithm using adaptive thresholding. We denote the return power of a peak as $q$, we select peaks which satisfy the following inequality constraint:
\begin{equation}
    q \geq \mu + \sigma
\end{equation}
where $\mu$ and $\sigma$ are the mean and the standard deviation of the powers of the peaks in one azimuth scan. Estimating the power mean and standard deviation along one azimuth instead of the whole polar image can mitigate the effect of receiver saturation since the radar may be saturated at one direction while rotating. By selecting the peaks whose powers are greater than one standard deviation plus their mean power, the true detections tend to be separated from the false-positive ones. The procedure is shown in Algorithm \ref{algo:point_cloud_generation}. Once a point cloud $\mathbf{C}$ is generated from a radar image, M2DP \cite{he2016m2dp}, a rotation invariant global descriptor designed for 3D point clouds, is adapted for the 2D radar point cloud to describe it for loop closure detection. M2DP computes the density signature of the point cloud on a plane and uses the left and right singular vectors of these signatures as the descriptor.

\subsection{Loop Candidate Rejection with PCA}\label{sec:PCA}

\begin{figure}
    \begin{subfigure}{0.23\linewidth}
    {\includegraphics[width=\linewidth]{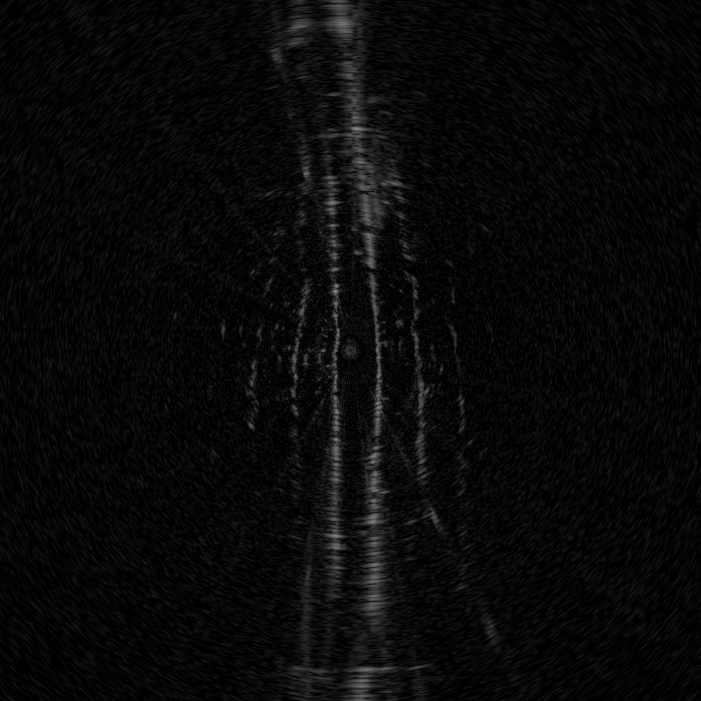}}
    \caption{Residential area A}
    \end{subfigure}
    \begin{subfigure}{0.23\linewidth}
    {\includegraphics[width=\linewidth]{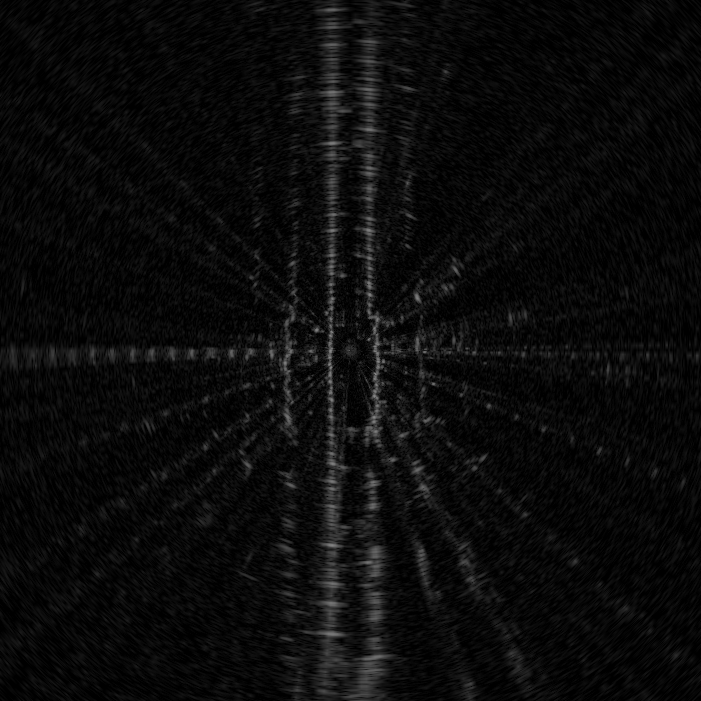}}
    \caption{Residential area A}
    \end{subfigure}
    \begin{subfigure}{0.23\linewidth}
    {\includegraphics[width=\linewidth]{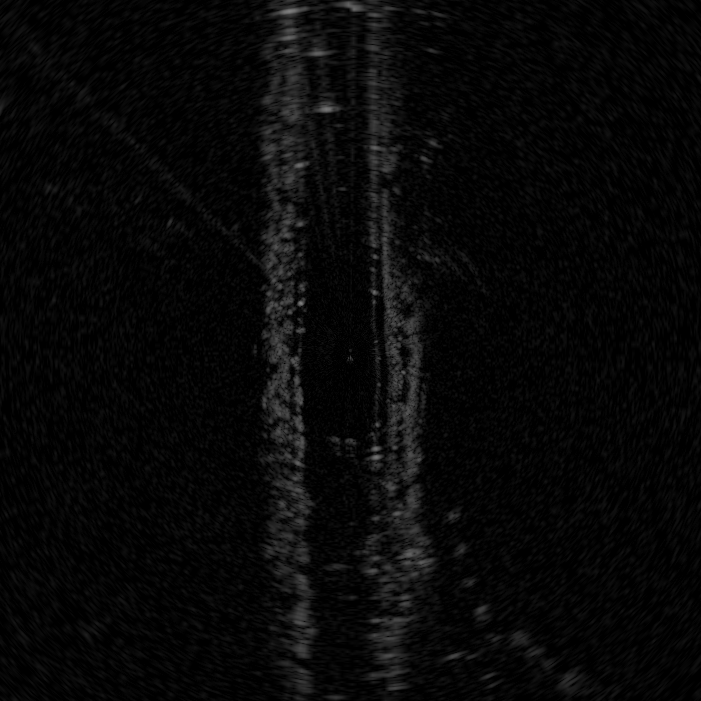}}
    \caption{Highway A}
    \end{subfigure}
    \begin{subfigure}{0.23\linewidth}
    {\includegraphics[width=\linewidth]{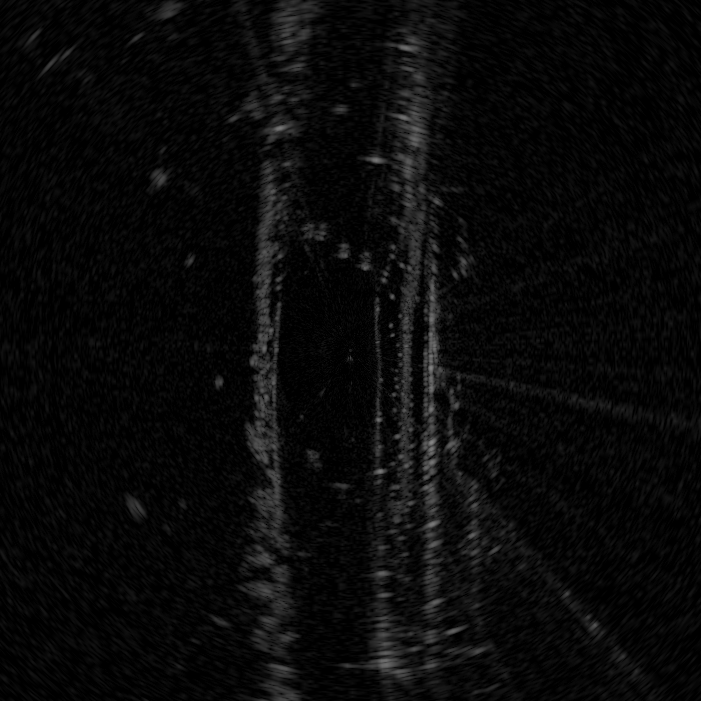}}
    \caption{Highway B}
    \end{subfigure}

    \caption{Radar images with large $r_{pca}$ tend to be ambiguous.}
    \label{fig:ambigious_places}
\end{figure}

We leverage principal component analysis (PCA) to determine whether the current frame can be a candidate to be matched with historical keyframes for loop closure detection. After performing PCA on the extracted 2D point cloud $\mathbf{C}$, we compute the ratio $r_{pca}=\gamma_1 / \gamma_2$ between the two eigen values $\gamma_1$ and $\gamma_2$ where $\gamma_1 \geq \gamma_2$.
The frame is selected for loop closure detection if its $r_{pca}$ is less than a certain threshold $\delta_{pca}$. The intuition behind this is to detect loop closure mainly on point clouds which have distinctive structural layouts, reducing the possibility of detecting false-positive loop closures. In other words, if $\gamma_1$ is dominant (i.e. $r_{pca}$ is big), it very likely that the radar imagery is collected in an environment, such as a highway or a country road, which exhibits less distinctive structural patterns and layouts and should be avoided for loop closure detection. Some examples are given in Fig. \ref{fig:ambigious_places}.

\subsection{Relative Transformation}
Once a loop closure is detected, the relative transformation $\mathbf{T}_{l,j}$ between the current radar image $\mathbf{I}_j$ and the matched radar image $\mathbf{I}_l$ is computed. Similar to Section \ref{sec:tracking}, we also associate keypoints of the two frames by using KLT tracker. The challenge here is that $\mathbf{I}_{l}$ might have a large rotation with respect to $\mathbf{I}_j$, causing the tracker to fail. To address this problem, we estimate firstly the relative rotation between the two frames and align them by using the eigenvectors from the PCA of their point clouds, similar to Section \ref{sec:PCA}. Then, the keypoints of $\mathbf{I}_{l}$ are tracked through the rotated version of $\mathbf{I}_j$. After obtaining the keypoint association, ICP is used to compute the relative transformation $\mathbf{T}_{l,j}$, which is added in the pose graph as a loop closure constraint for pose graph optimization.

\subsection{Pose Graph Optimization}
A pose graph is gradually built as the radar moves. Once a new loop closure constraint is added in the pose graph, pose graph optimization is performed. After successfully optimizing the poses of the keyframes, we update the global map points. The g2o \cite{kummerle2011g} library is used in this work for the pose graph optimization.



\section{Experimental Results} \label{sec:result}
Both quantitative and qualitative experiments are conducted to evaluate the performance of the proposed radar SLAM method using three open radar datasets, covering large-scale environments and some adverse weather conditions.

\subsection{Evaluation Protocol}


We perform both quantitative and qualitative evaluation using different datasets. Specifically, the quantitative evaluation is to understand the pose estimation accuracy of the SLAM system. For Relative/Odometry Error (RE), we follow the popular KITTI odometry evaluation criteria, i.e., computing the mean translation and rotation errors from length $100$ to $800$ meters with a $100$ meters increment. Absolute Trajectory Error (ATE) is also adopted to evaluate the localization accuracy of full SLAM, in particular after loop closure and global graph optimization. The trajectories of all methods (see full list in Section \ref{sec:exp_setting_competing} are aligned with the ground truth trajectories using a 6 Degree-of-Freedom (DoF) transformation provided by the evaluation tool in \cite{Zhang18iros} for ATE evaluation. On the other hand, the qualitative evaluation focuses on how some challenging scenarios, e.g. in adverse weather conditions, influence the performance of various vision, LiDAR and radar based SLAM systems.

\subsection{Datasets}

So far there exist three public datasets that provide long-range radar data with dense returns: the Oxford RobotCar Radar Dataset \cite{RadarRobotCarDatasetICRA2020,RobotCarDatasetIJRR}, the MulRan Dataset \cite{gskim-2020-mulran} and the RADIATE Dataset \cite{sheeny2020radiate}. We choose the Oxford RobotCar and MulRan datasets for detailed quantitative benchmarking and our RADIATE dataset mainly for qualitative evaluation in our experiments.

\subsubsection{Oxford RobotCar Radar Dataset.}
\begin{figure}[t]
    \centering
    \includegraphics[width=\linewidth]{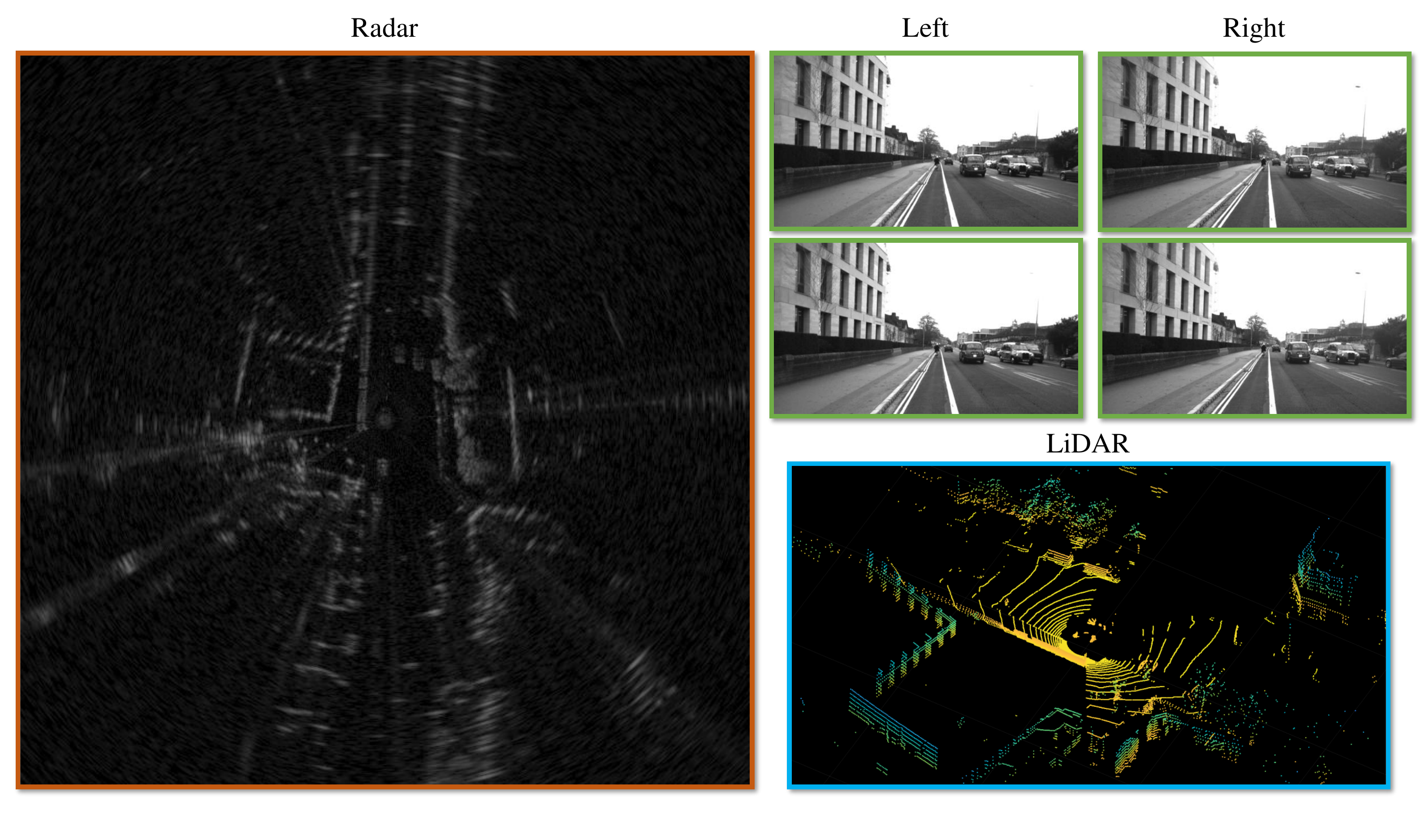}
    \caption{Synchronized radar, stereo, and LiDAR data from the Oxford Radar RobotCar Dataset \cite{RadarRobotCarDatasetICRA2020}.}
    \label{fig:oxford_data}
\end{figure}

The Oxford RobotCar Radar Dataset \cite{RadarRobotCarDatasetICRA2020,RobotCarDatasetIJRR} provides data from a Navtech CTS350-X Millimetre-Wave W radar for about $280$ km of driving in Oxford, UK, traversing the same route 32 times. It also provides stereo images from a Point Grey Bumblebee XB3 camera and LiDAR data from two Velodyne HDL-32E sensors with ground truth pose locations. The radar is configured to provide 4.38 cm and 0.9 degree resolutions in range and azimuth respectively, with a range up to 163 meters. The radar scanning frequency is 4 Hz. See Fig. \ref{fig:oxford_data} for some examples of data.

\begin{figure}[t]
    \centering
    \includegraphics[width=\linewidth]{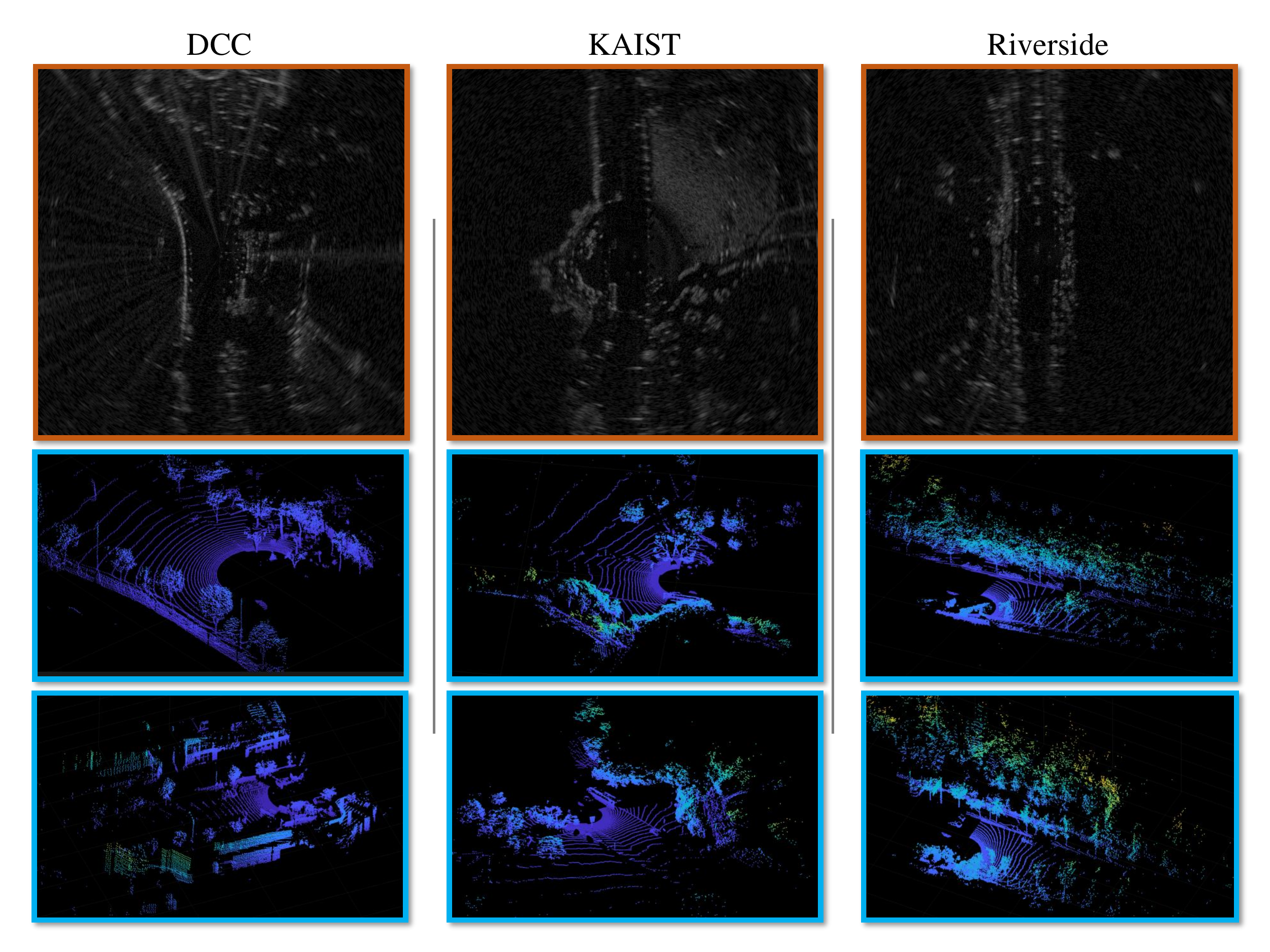}
    \caption{Radar and LiDAR data from the MulRan Dataset. In Riverside, we can see the repetitive structures of trees and bushes, which makes it challenging for LiDAR based odometry and mapping algorithms \cite{gskim-2020-mulran}.}
    \label{fig:mulran_data}
\end{figure}

\begin{table}[t]
    \footnotesize
    \centering
    \caption{Lengths of Collected Sequences in RADIATE Dataset.} \label{tab:radiate_dataset}
    \scalebox{0.83}{
    \begin{tabular}{c|c|c|c|c|c|c|c}
    \hline
    Sequence    & Fog 1 & Fog 2 & Rain &Normal & Snow 1 & Snow 2  & Night \\ \hline
    Length (km) & 4.7   & 4.8   & 3.3  & 3.3 & 8.7 & 3.3 & 5.6   \\ \hline
    \end{tabular}
    }
\end{table}

\begin{figure}[t]
    \centering
    \includegraphics[width=\linewidth]{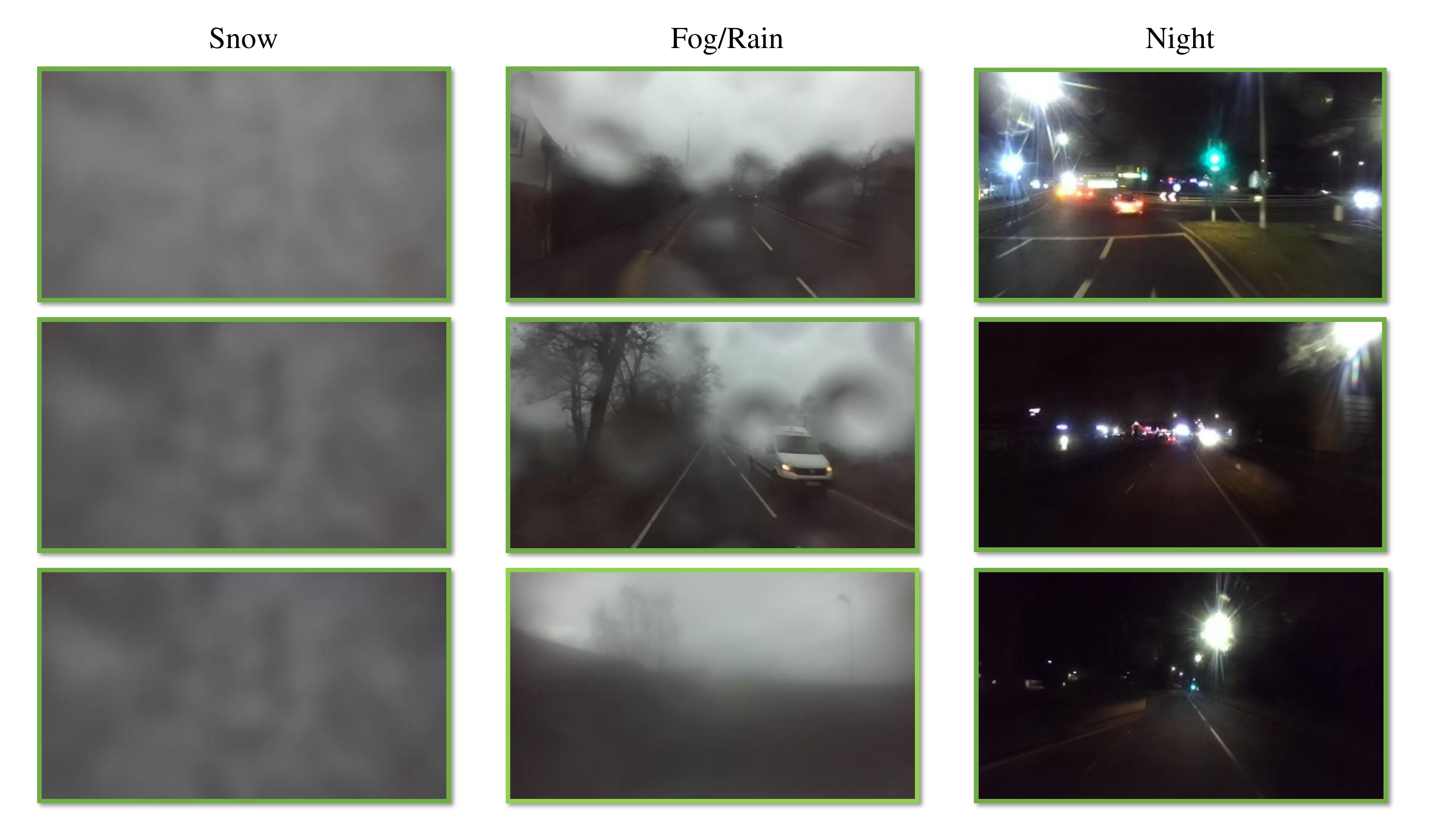}
    \caption{Images collected in Snow (left), Fog/Rain (middle), and Night (right). The image quality degrades in these conditions, making it extremely challenging for vision based odometry and SLAM algorithms. Note that for the snow sequence, the camera is completely covered by snow.}
    \label{fig:radiate_camera_data}
\end{figure}

\subsubsection{MulRun Dataset.}
The MulRan Dataset \cite{gskim-2020-mulran} provides radar and LiDAR range data, covering multiple cities at different times in a variety of city environments (e.g., bridge, tunnel and overpass). A Navtech CIR204-H Millimetre-Wave FMCW radar is used to obtain radar images with 6 cm range and 0.9 degree rotation resolutions with a maximum range of 200 m. The radar scanning frequency is also 4Hz. It also has an Ouster 64-channel LiDAR sensor operating at 10Hz with a maximum range of 120 m. Different routes are selected for our experiments, including Dajeon Convention Center (DCC), KAIST and Riverside. Specifically, DCC presents diverse structures while KAIST is collected while moving within a campus. Riverside is captured along a river and two bridges with repetitive features. Each route contains 3 traverses on different days. Some LiDAR and radar data examples are given in Fig. \ref{fig:mulran_data}.

\subsubsection{RADIATE Dataset.}
The RADIATE dataset is our recently released dataset which includes radar, LiDAR, stereo camera and GPS/IMU \cite{sheeny2020radiate}. One of its unique features is that it provides data in extreme weather conditions, such as rain and snow, as shown Fig. \ref{fig:radiate_camera_data}. A Navtech CIR104-X radar is used with 0.175 m range resolution and maximum range of 100 m at 4 Hz operation frequency. A 32-channel Velodyne HDL-32E LiDAR and a ZED stereo camera are set at 10Hz and 15 Hz, respectively. The 7 sequences used in this work include 2 fog, 1 rain, 1 normal, 2 snow and 1 night recorded in the City of Edinburgh, UK. Their sequence lengths are given in Table \ref{tab:radiate_dataset}. Note that only the rain, normal, snow and night sequences have loop closures and the GPS signal is occasionally lost in the snow sequence.


\begin{table*}[t!]
    \footnotesize
    \centering
    \caption{Relative error on Oxford Radar RobotCar Dataset}
    \label{tab:oxford_re_error}
    \begin{tabular}{ l|p{1.1cm}p{1.1cm}p{1.1cm}p{1.1cm}p{1.1cm}p{1.1cm}p{1.1cm}p{1.1cm}p{1.2cm}}
    \hline
    & \multicolumn{9}{c}{\textbf{Sequence}} \\
    \textbf{Method} &10-11-46-21 & 10-12-32-52 & 16-11-53-11 & 16-13-09-37 & 17-13-26-39 & 18-14-14-42  & 18-14-46-59 & 18-15-20-12 & Mean\\
    \hline
    ORB-SLAM2       &6.11/1.7    & 6.09/1.6    & 6.16/1.7    & 6.23/1.7    & 6.41/1.7    & 7.05/1.8     & 7.17/1.9    & 11.5/3.3    &7.09/3.1    \\
    \hline
    SuMa            & 1.1/0.3$\color{red}^{*12\%}$ & 1.1/0.3$\color{red}^{*20\%}$ & 0.9/0.3$\color{red}^{*27\%}$ & 1.2/0.4$\color{red}^{*29\%}$
                    & 1.1/0.3$\color{red}^{*23\%}$ & 0.9/0.1$\color{red}^{*10\%}$ & 1.0/0.1$\color{red}^{*10\%}$ & 1.0/0.2$\color{red}^{*20\%}$
    &1.03/0.3 \\
    \hline
    Cen Odometry    &N/A &N/A &N/A &N/A &N/A &N/A &N/A &N/A &3.71/0.95\\
    Barnes Odometry &N/A &N/A &N/A &N/A &N/A &N/A &N/A &N/A &2.7848/0.85\\
    Baseline Odometry &3.26/0.9  &2.98/0.8      &3.28/0.9 & 3.12/0.9    & 2.92/0.8  & 3.18/0.9   & 3.33/1.0 & 2.85/0.9  & 3.11/0.9 \\
    Baseline SLAM   &2.27/0.9   &2.16/0.6       &2.24/0.6 & 1.83/0.6    & 2.45/0.8  & 2.21/0.7   & 2.34/0.7 & 2.24/0.8  & 2.21/0.7 \\
    \hline
    Our Odometry    &2.16/0.6      & 2.32/0.7      & 2.49/0.7    & 2.62/0.7     & 2.27/0.6   & 2.29/0.7  & 2.12/0.6   &  2.25/0.7              & 2.32/0.7 \\
    Our SLAM        &1.96/0.7      & 1.98/0.6      & 1.81/0.6    & 1.48/0.5     & 1.71/0.5   & 2.22/0.7  & 1.68/0.5   &1.77/0.6           & 1.83/0.6 \\
    \hline
    \end{tabular}\par
    \smallskip
    Results are given as \textit{translation error} / \textit{rotation error}. Translation error is in \%, and rotation error is in degrees per 100 meters (deg/100m). For the Cen and Barnes odometry methods, only their mean errors are shown since individual sequence errors are not reported in their papers. $\color{red}^{*}xx\%$ indicates that the algorithm cannot finish the full sequence, and its result is reported up to the point ($xx\%$ of the full sequence) where it fails.
\end{table*}
\begin{table*}[h]
    \centering
    \caption{Absolute trajectory error for position (RMSE) on Oxford Radar RobotCar Dataset}
    \footnotesize
    \begin{tabular}{ p{2.5cm}|p{1.1cm}p{1.1cm}p{1.1cm}p{1.1cm}p{1.1cm}p{1.1cm}p{1.1cm}p{1.1cm}}
    \hline
    & \multicolumn{8}{c}{\textbf{Sequence}} \\
    \textbf{Method}   & 10-11-46-21 & 10-12-32-52 & 16-11-53-11 & 16-13-09-37 & 17-13-26-39& 18-14-14-42 & 18-14-46-59 &  18-15-20-12 \\
    \hline
    ORB-SLAM2 &     \textbf{7.301} & \textbf{7.961} & \textbf{3.539} & \textbf{7.590} & 7.609 & 24.632 & 9.715 &  12.174 \\
    SuMa   & N/A & N/A & N/A & N/A & N/A & N/A & N/A & N/A   \\
    Baseline SLAM &     58.138 & 14.598 & 12.933 & 12.829 & 10.898 & 49.599 & 23.270 &  56.422 \\
    Our SLAM &     13.784 & 9.593 & 7.136 & 11.182 & \textbf{5.835} & \textbf{21.206} & \textbf{6.011} &  \textbf{7.740} \\
    \hline
    \end{tabular}\par
    \smallskip
    The absolute trajectory error of position is in meters. N/A: SuMa fails to finish all eight sequences, no absolute trajectory error is applicable here.
    \label{tab:oxford_ate_error}
\end{table*}

\begin{figure*}[h]
    \centering
    \begin{subfigure}{0.24\textwidth}
    \includegraphics[width=\linewidth]{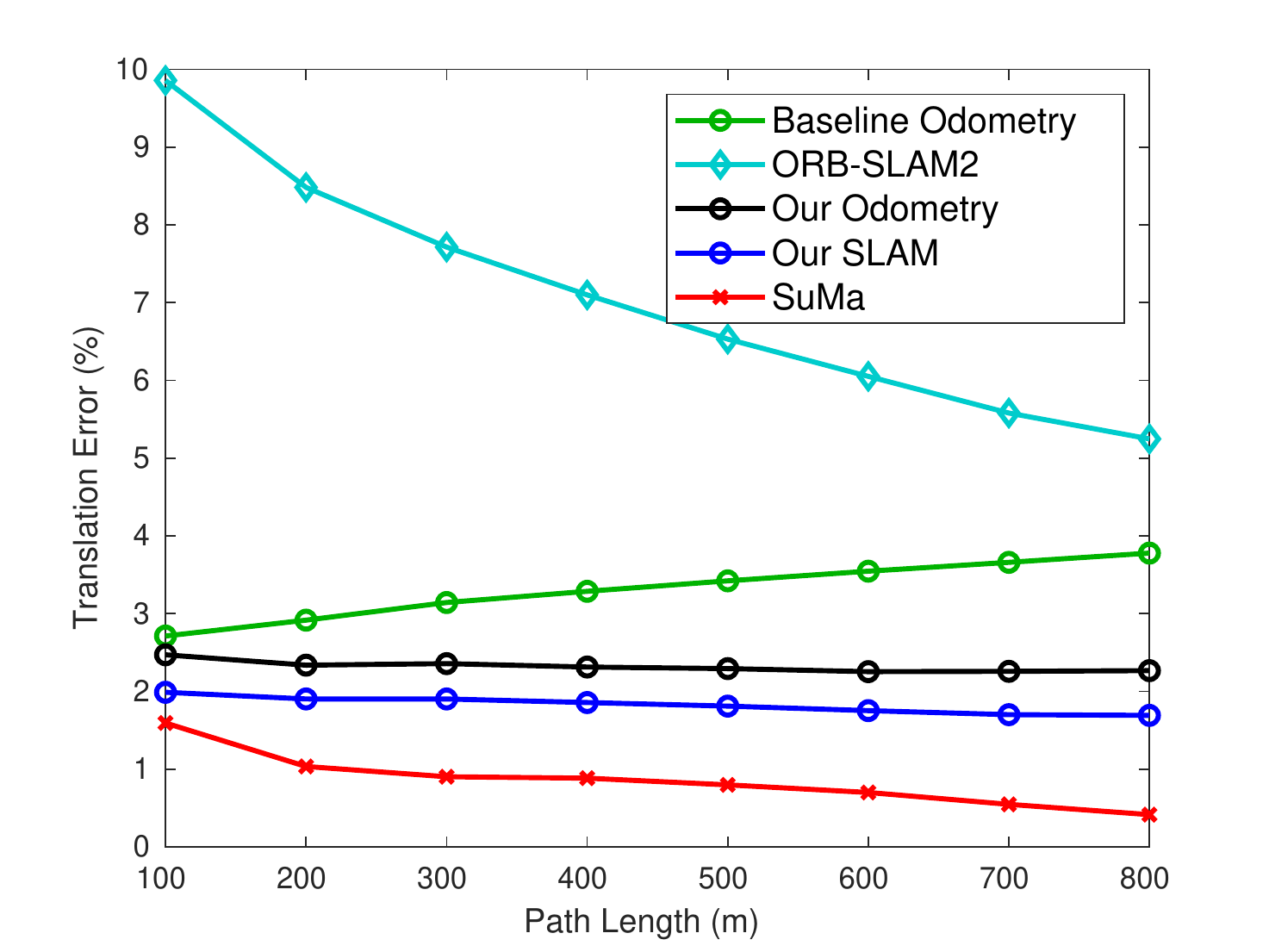}
    \caption{Translation against path length.}
    \end{subfigure}
    \begin{subfigure}{0.24\textwidth}
    \includegraphics[width=\linewidth]{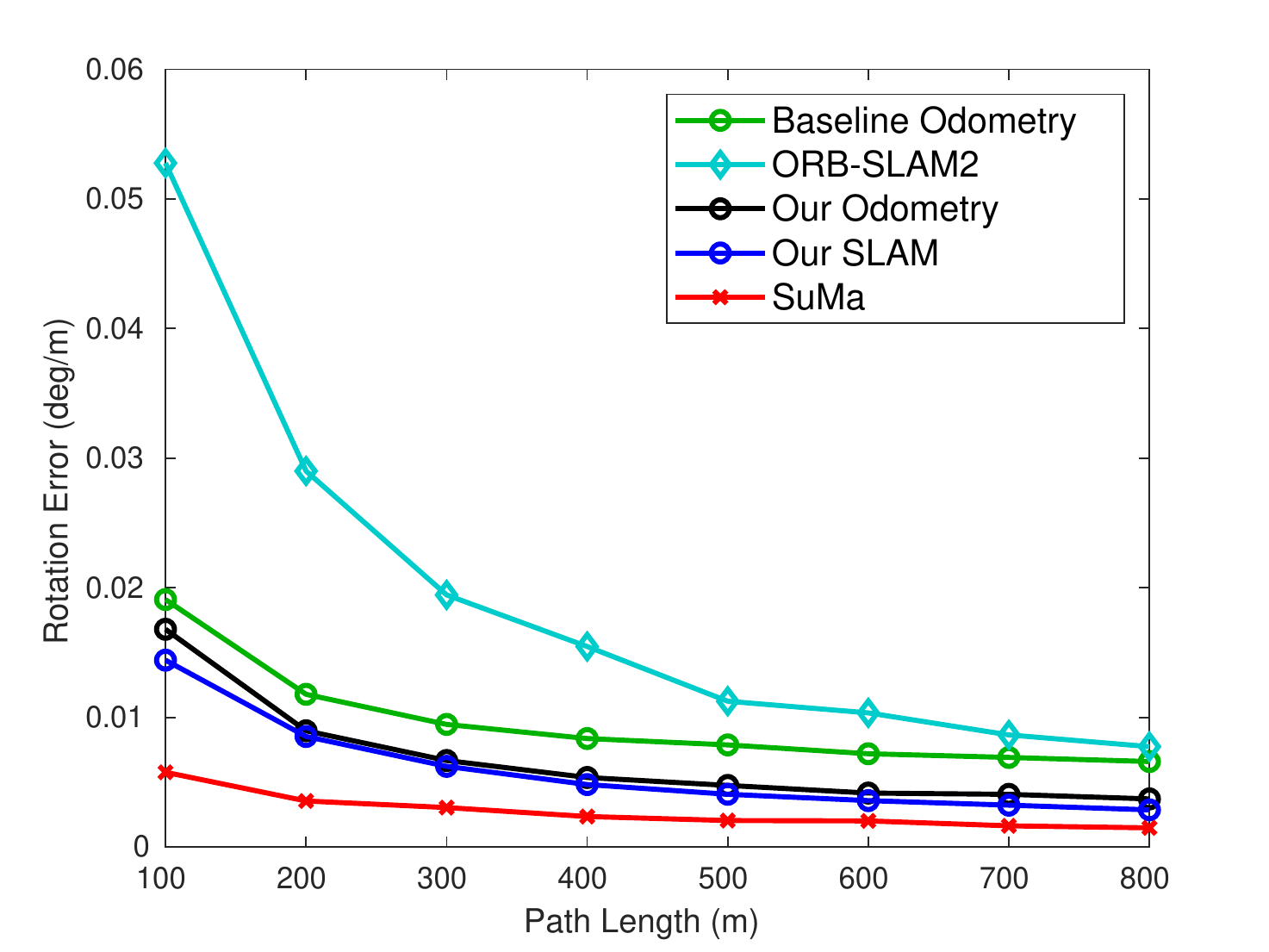}
    \caption{Rotation against path length.}
    \end{subfigure}
    \begin{subfigure}{0.24\textwidth}
    \includegraphics[width=\linewidth]{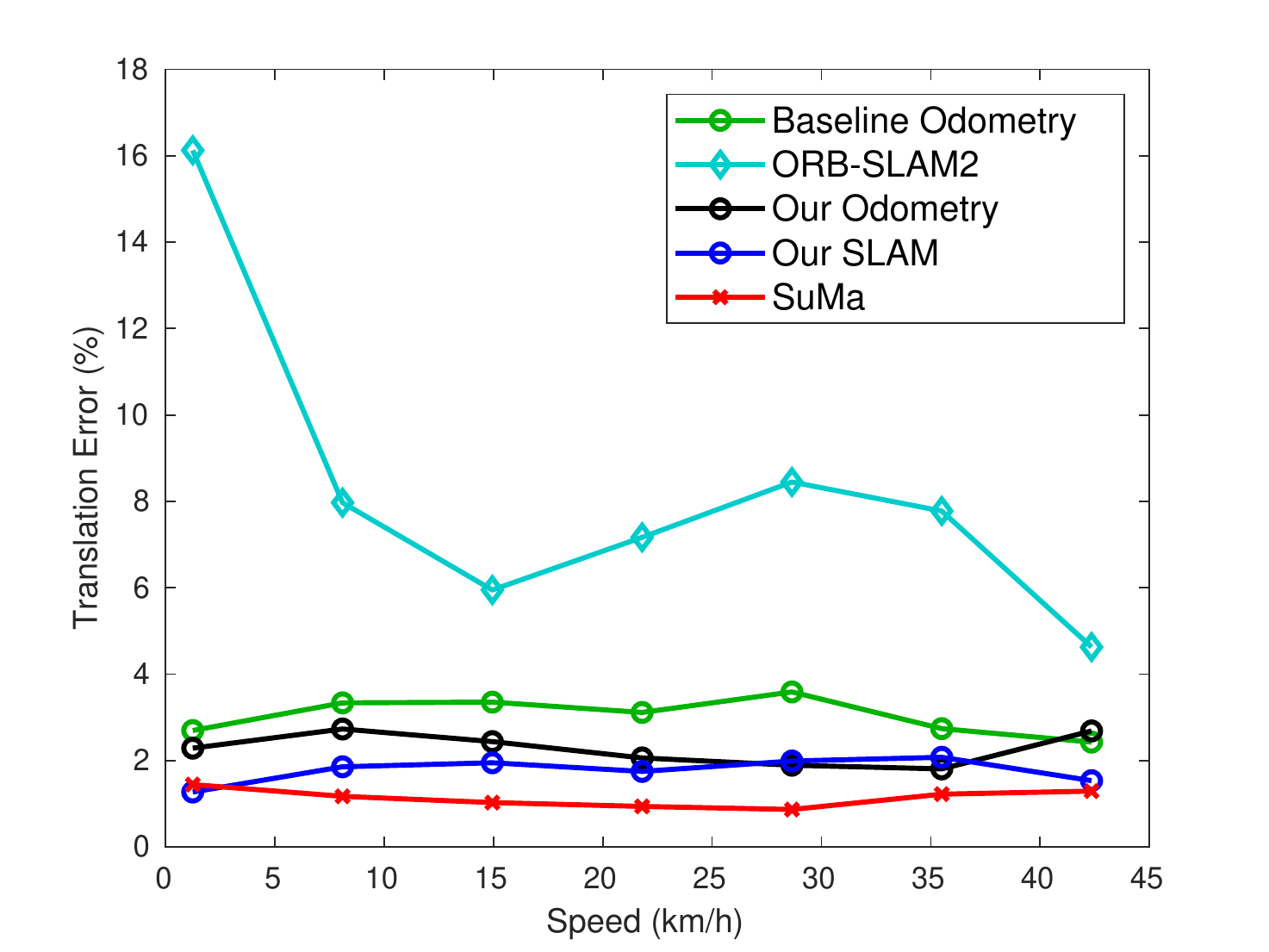}
    \caption{Translation against speed.}
    \end{subfigure}
    \begin{subfigure}{0.24\textwidth}
    \includegraphics[width=\linewidth]{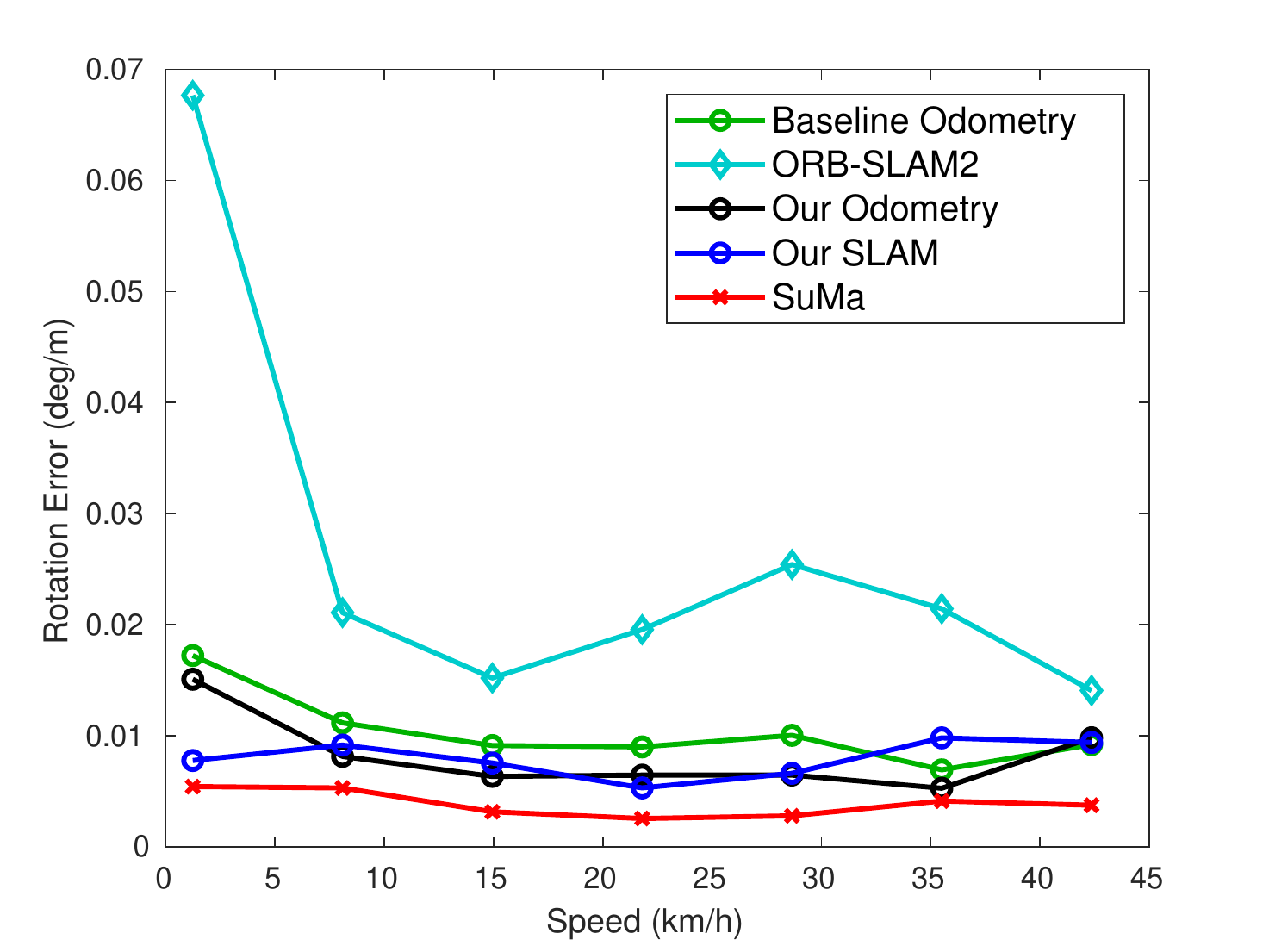}
    \caption{Rotation against speed.}
    \end{subfigure}
    \caption{Average error against different path lengths and speeds on Oxford Radar RobotCar Dataset. Note that SuMa's results are computed up to the point where it fails matching the $xx\%$ percentages in Table \ref{tab:oxford_re_error}.}
    \label{fig:error_lengths_speed}
\end{figure*}

\begin{figure*}[h]
    \centering
    \begin{subfigure}{0.49\linewidth}
    \includegraphics[width=\linewidth]{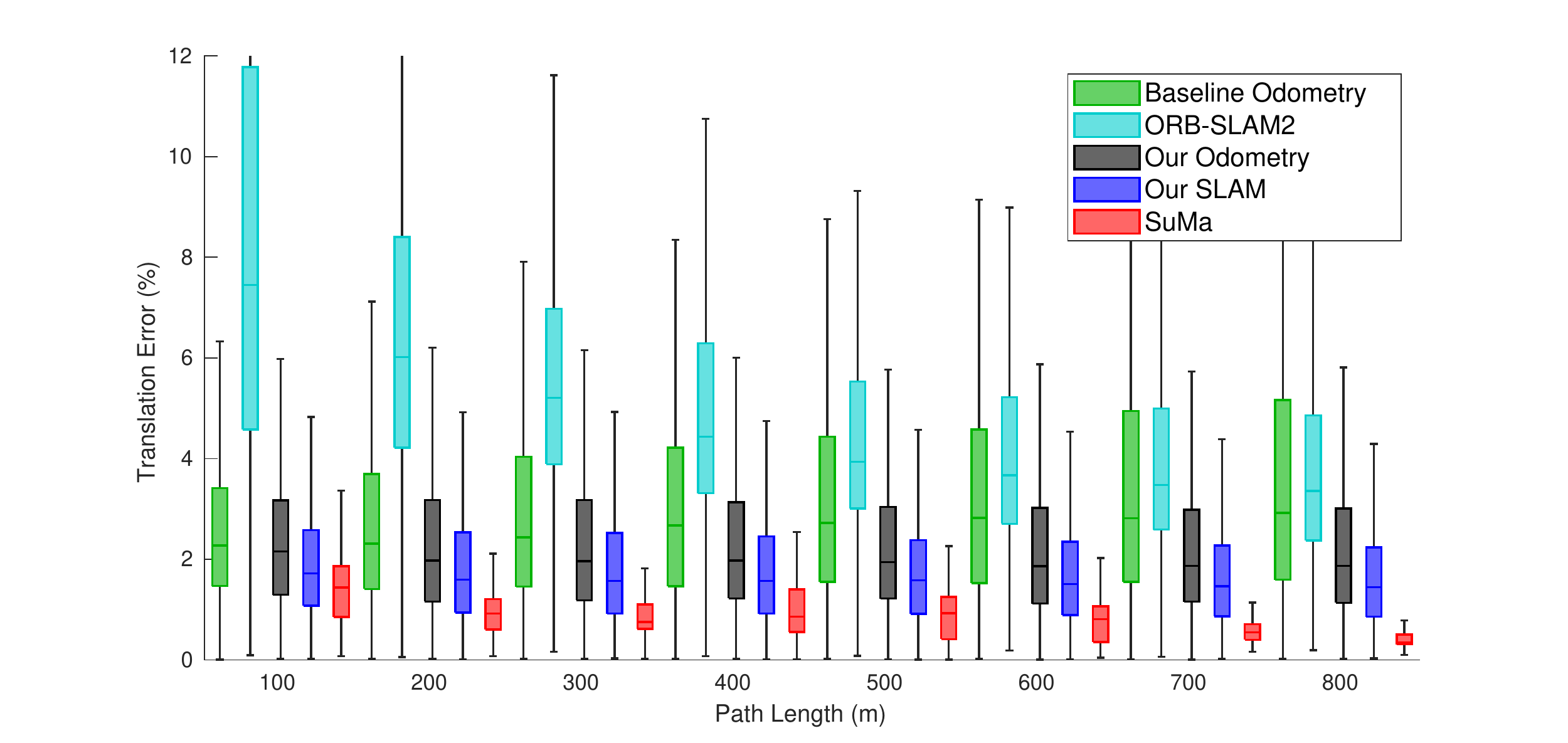}
    \caption{Low, median and high translation errors against path length.}    \label{fig:box_chart_translation}
    \end{subfigure}
    \begin{subfigure}{0.49\linewidth}
    \includegraphics[width=\linewidth]{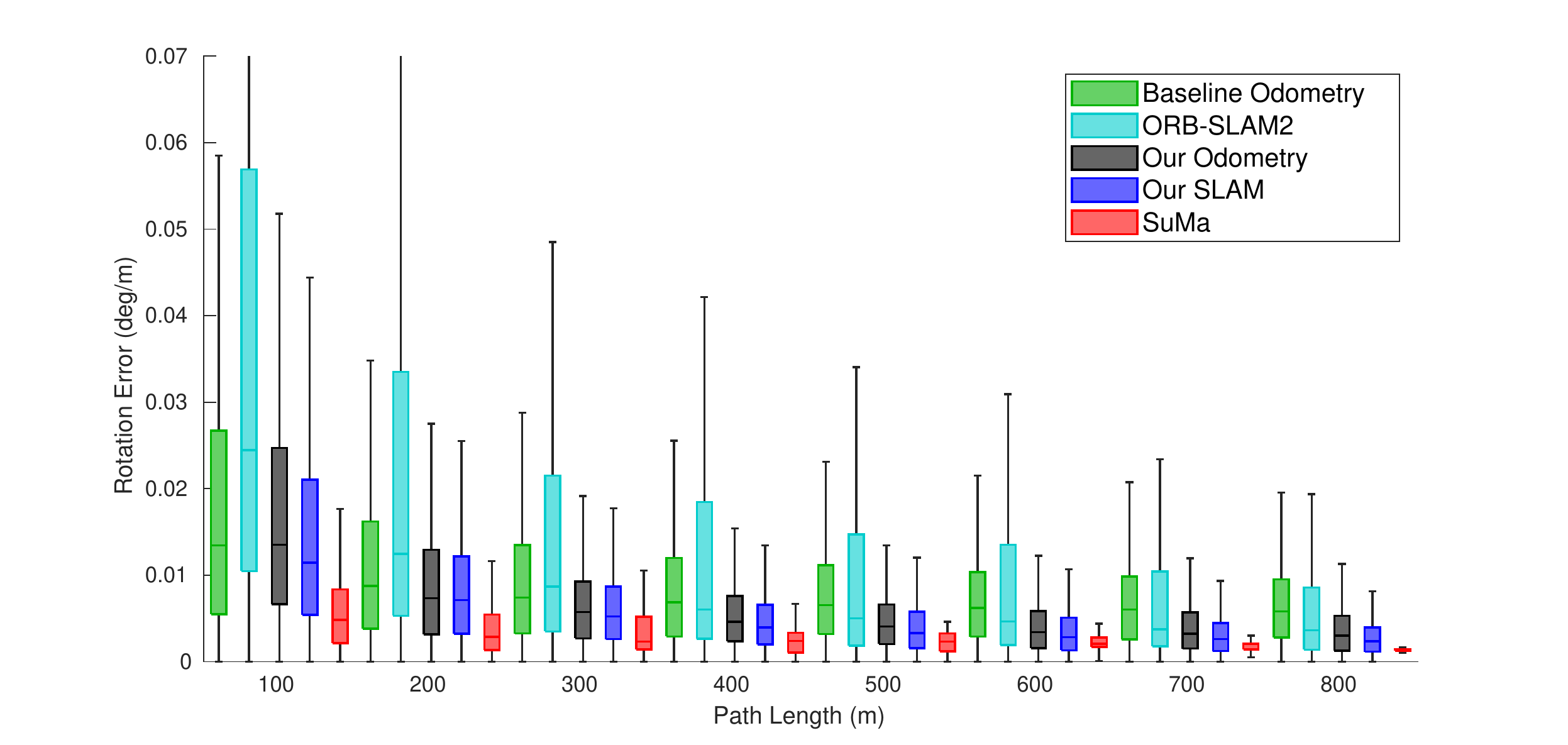}
    \caption{Low, median and high rotation errors against path length.} \label{fig:box_chart_rotation}
    \end{subfigure}
    \caption{Bar charts for translation and rotation errors against path length on Oxford Radar RobotCar Dataset.}
\end{figure*}

\subsection{Competing Methods and Their Settings}\label{sec:exp_setting_competing}

In order to validate the performance of our proposed radar SLAM system, state-of-the-art odometry and SLAM methods for large-scale environments using different sensor modalities (camera, LiDAR, radar) are chosen. These include ORB-SLAM2 \cite{murORB2}, SuMa \cite{behley2018rss} and our previous version of RadarSLAM \cite{hong2020radarslam}, as baseline algorithms for vision, LiDAR and radar based approaches, respectively. For the Oxford Radar RobotCar Dataset, the results reported in \cite{cen2018precise,barnes2019masking} are also included as a radar based method due to the unavailability of their implementations.

We would like to highlight that we use \textit{an identical set of parameters for our radar odometry and SLAM algorithm across all the experiments and datasets, without any parameter tuning.} We believe this is worthwhile to tackle the challenge that most existing odometry or SLAM algorithms require some levels of parameter tuning in order to reduce or avoid result degradation.

\subsubsection{Stereo Vision based ORB-SLAM2.}
OBR-SLAM2 \cite{murORB2} is a sparse feature based visual SLAM system which relies on ORB features. It also possesses loop closure and pose graph optimization capabilities. Local Bundle Adjustment is used to refine the map point position which boosts the odometry accuracy. Based on its official open-source implementation, we use its stereo setting in all experiments and loop closure is enabled.

\subsubsection{LiDAR based SuMa.}
SuMa \cite{behley2018rss} is one of the state-of-the-art LiDAR based odometry and mapping algorithms for large-scale outdoor environments, especially for mobile vehicles. It constructs and uses a surfel-based map to perform robust data association for loop closure detection and verification. We employ its open-source implementation and keep the original parameter setting used for KITTI dataset in our experiments.

\subsubsection{Radar based RadarSLAM.}
Our old version of RadarSLAM \cite{hong2020radarslam} extracts SURF features from Cartesian radar images and matches the keypoints based on their descriptors for pose estimation, which is different from the feature tracking technique in this work. It does not consider motion distortion although it includes loop closure detection and pose graph optimization to reduce drift and improve the map consistency.

\subsubsection{Cen's Radar Odometry.}
Cen's method \cite{cen2018precise} is one of the first attempts using the Navtech FMCW radar sensor to estimate ego-motion of a mobile vehicle. Landmarks are extracted from polar scans before performing data association by maximizing the overall compatibility with pairwise constraints. Given the associated pairs, SVD is used to find the relative transformation.

\subsubsection{Barnes' Radar Odometry.}
Barnes' method \cite{barnes2019masking} leverages deep learning to generate distraction-free feature maps and uses FFT cross correlation to find relative poses on consecutive feature maps. After being trained end-to-end, the system is able to mask out multipath reflection, speckle noise and dynamic objects. This facilitates the cross correlation stage and produces accurate odometry. The spatial cross-validation results in appendix of \cite{barnes2019masking} are chosen for fair comparison.

\subsection{Experiments on RobotCar Dataset}
Results of eight sequences of RobotCar Dataset are reported here for evaluation, i.e., 10-12-32-52, 16-13-09-37, 17-13-26-39,  18-15-20-12, 10-11-46-21, 16-11-53-11 and 18-14-46-59. The wide baseline stereo images are used for the stereo ORB-SLAM2 and the left Velodyne HDL-32E sensor is used for SuMa.

\subsubsection{Quantitative Comparison}
    
The RE and ATE results of each sequence are given in Tables \ref{tab:oxford_re_error} and \ref{tab:oxford_ate_error} respectively. Since the groundtruth poses provided by the Oxford Radar RobotCar Dataset are 3-DoF, only the $x, y$ and $yaw$ of the estimated 6-DoF poses of ORB-SLAM2 and SuMa are evaluated. \textit{Note that SuMa fails on all eight sequences at $10-30\%$ of the full lengths and all its results are reported until the point where it fails, and ATE is not applicable due to the lack of fully estimated trajectories.} Specifically, the stereo version of ORB-SLAM2 is able to complete all eight sequences, successfully close the loops and achieve superior localization accuracy. SuMa also performs accurately when it works although it is less robust using this dataset. This may be due to the large number of dynamic objects, e.g. surrounding moving cars and buses. Regarding radar based approaches, we can see that our proposed radar odometry/SLAM achieves less RE compared to the baseline radar odometry/SLAM and Cen's method and a similar mean AE to the learning based Barnes' method. It can also be seen that our proposed radar odometry and SLAM methods achieve better or comparable RE and ATE performance to ORB-SLAM2 and SuMa.
\begin{figure}[t!]
    \centering
    \includegraphics[width=\linewidth]{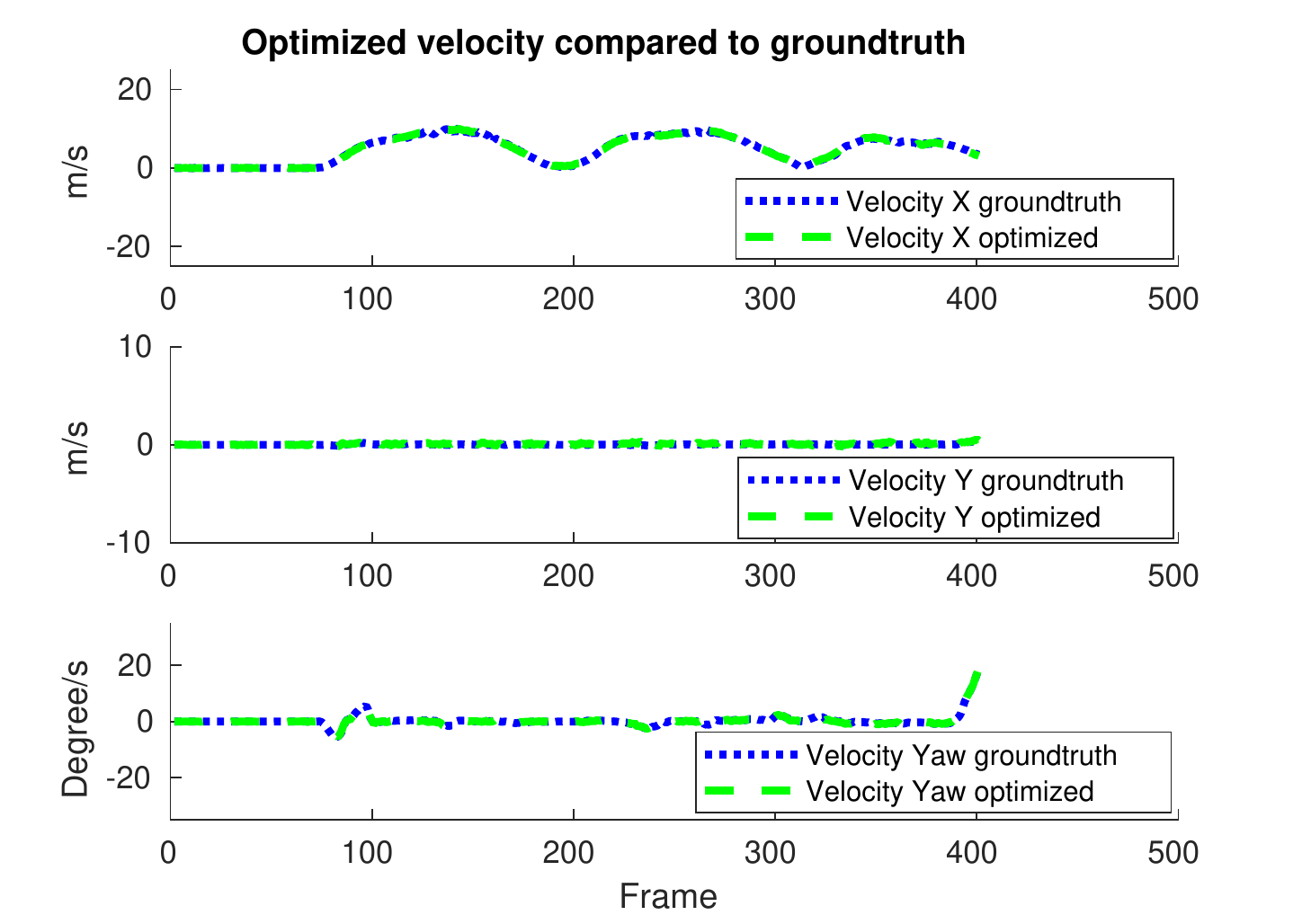}
    \caption{Optimized x, y and yaw velocities on RobotCar.}
    \label{fig:velocity_comparision}
\end{figure}
Fig. \ref{fig:error_lengths_speed} describes the REs of ORB-SLAM2, SuMa, baseline radar odometry and our radar odometry/SLAM algorithms using different path lengths and speeds, following the popular KITTI odometry evaluation protocol. SuMa has the lowest error on both translation and rotation against path lengths and speed until it fails, while our radar SLAM and odometry methods are the second and third lowest respectively.
The low, median and high translation and rotation errors are presented in Figs. \ref{fig:box_chart_translation} and \ref{fig:box_chart_rotation}. It can be seen that our SLAM and odometry achieve low values for both translation and rotation errors for different path lengths.

The optimized velocities of x, y and yaw are given in Fig \ref{fig:velocity_comparision} compared to the ground truth. The optimized velocities have very high accuracy, which verifies the superior performance of our proposed radar motion tracking algorithm.

\subsubsection{Qualitative  Comparison}

\begin{figure*}[h]
    \centering
    \begin{subfigure}{0.3\textwidth}
    \includegraphics[width=\linewidth]{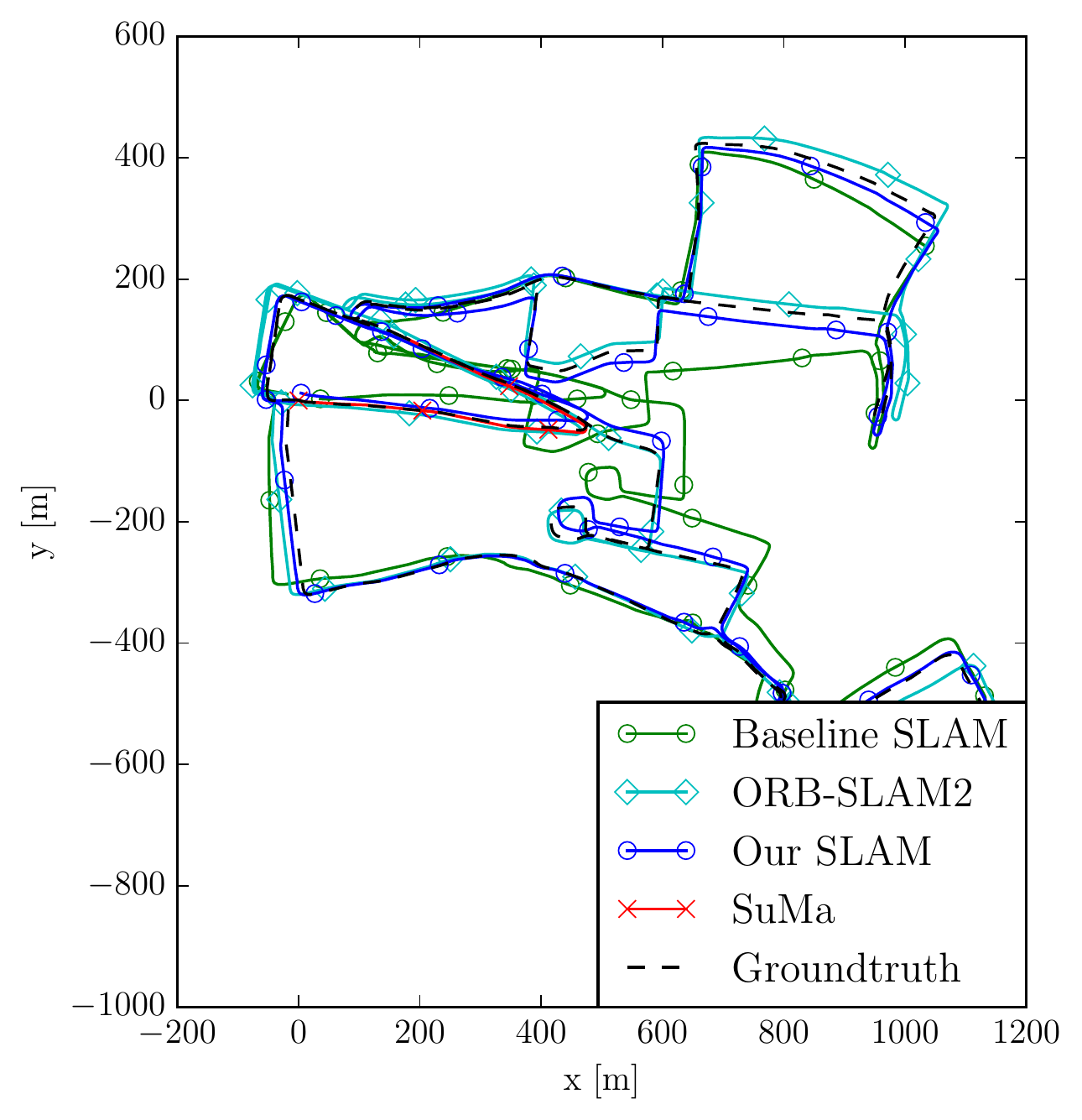}
    \caption{10-11-46-21}
    \end{subfigure}
    \begin{subfigure}{0.3\textwidth}
    \includegraphics[width=\linewidth]{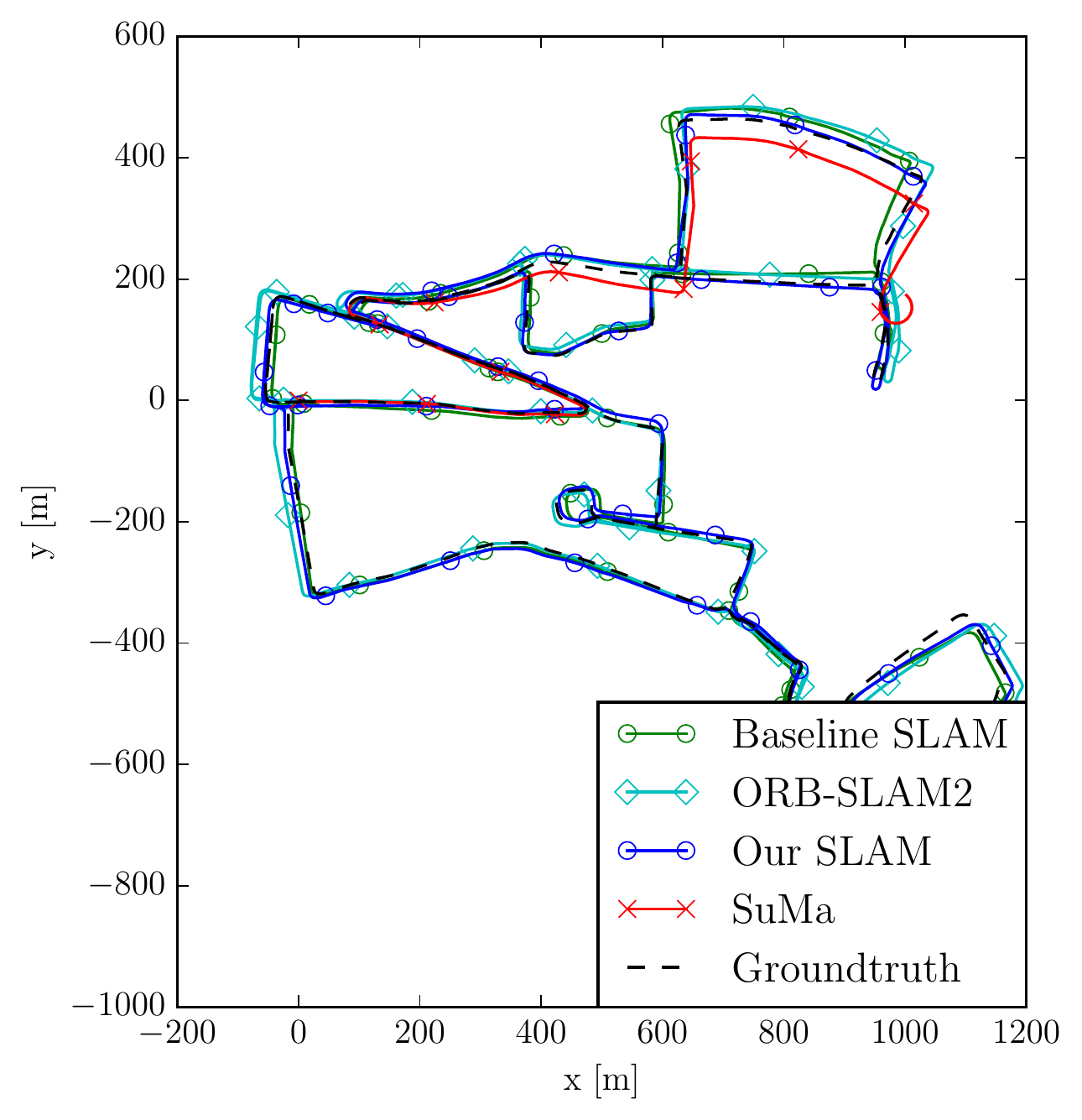}
    \caption{10-12-32-52}
    \end{subfigure}
    \begin{subfigure}{0.3\textwidth}
    \includegraphics[width=\linewidth]{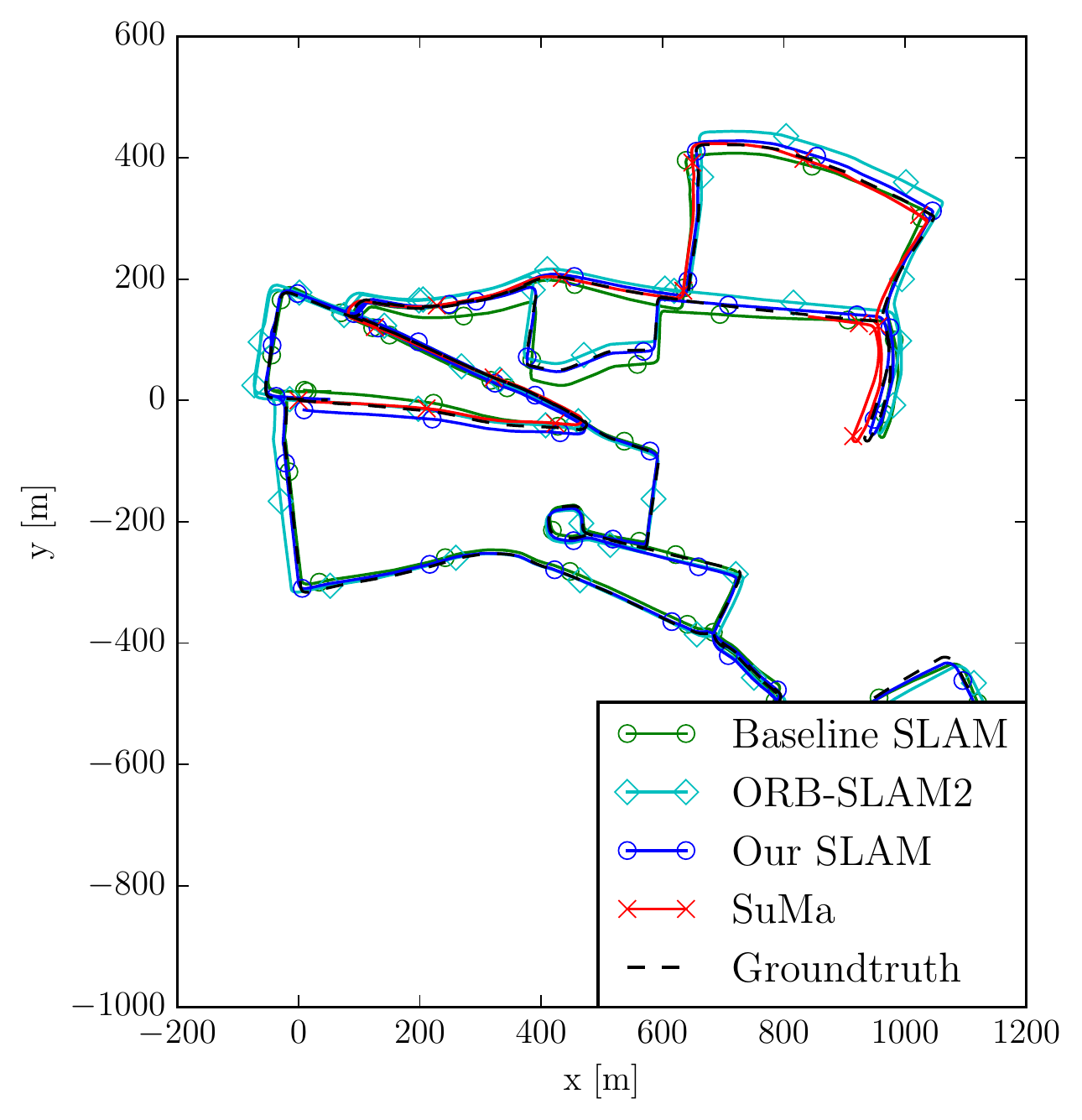}
    \caption{16-11-53-11}
    \end{subfigure}
    \hfill
    \begin{subfigure}{0.3\textwidth}
    \includegraphics[width=\linewidth]{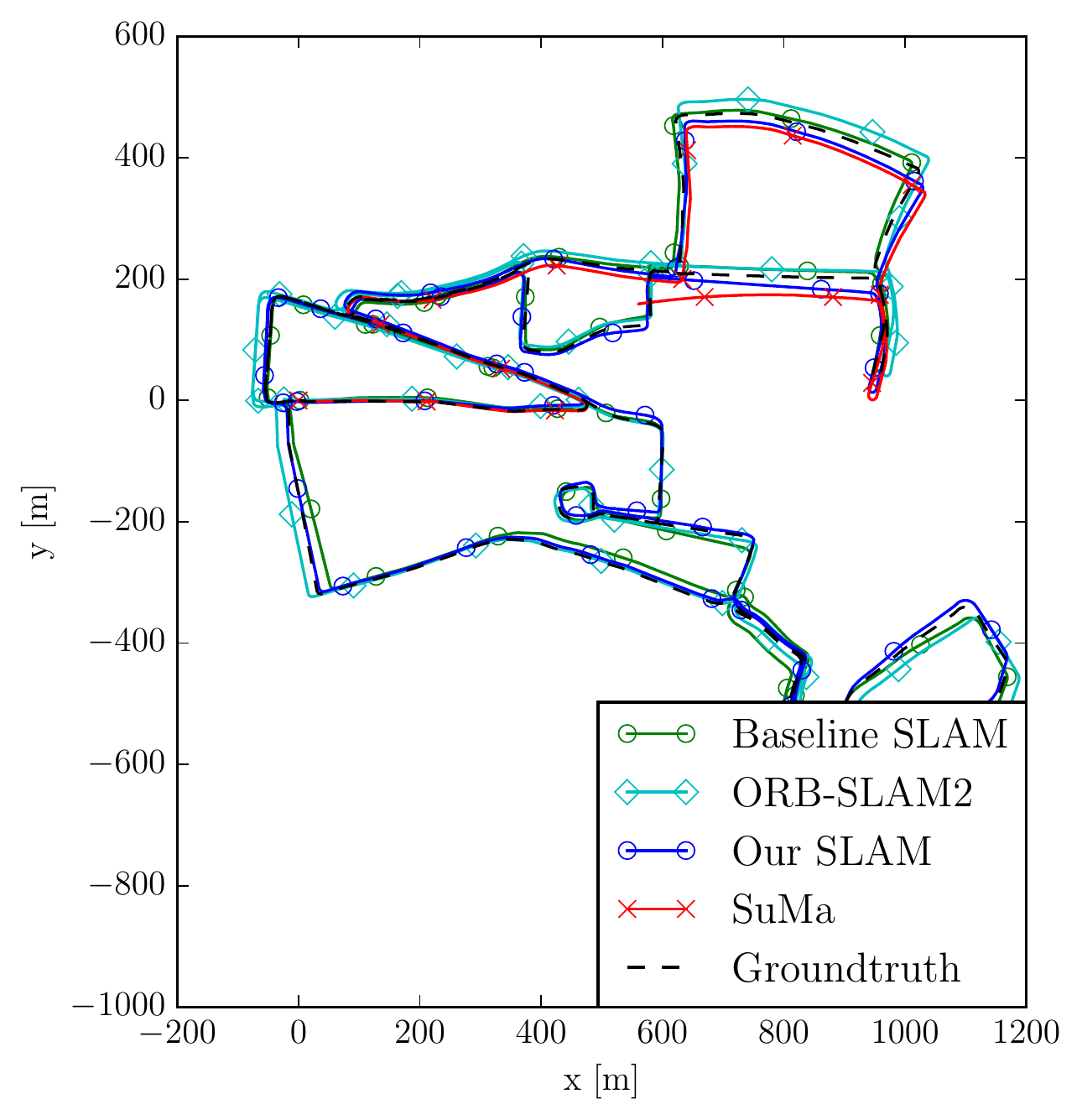}
    \caption{16-13-09-37}
    \end{subfigure}
    \begin{subfigure}{0.3\textwidth}
    \includegraphics[width=\linewidth]{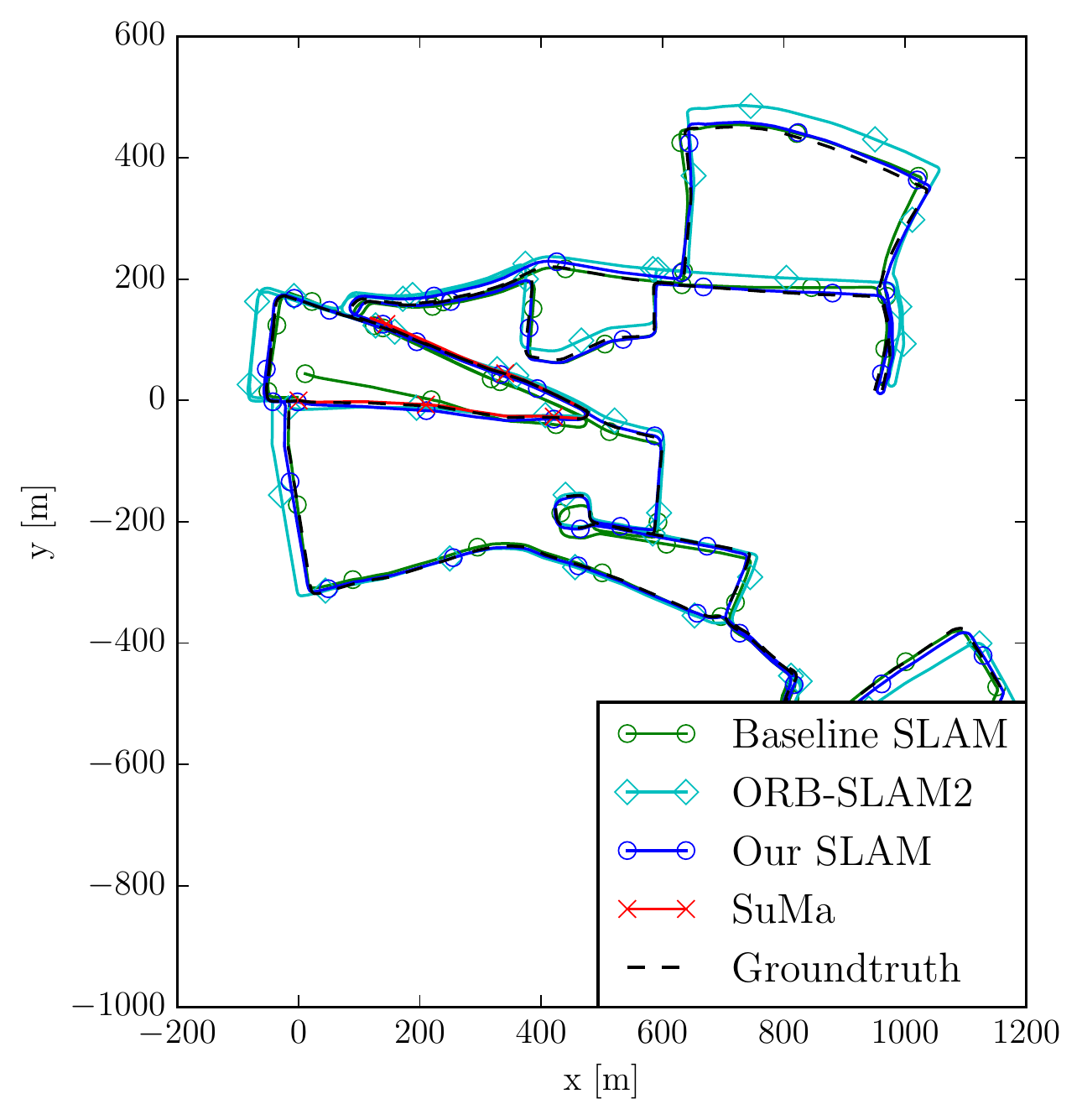}
    \caption{17-13-26-39}
    \end{subfigure}
    \begin{subfigure}{0.3\textwidth}
    \includegraphics[width=\linewidth]{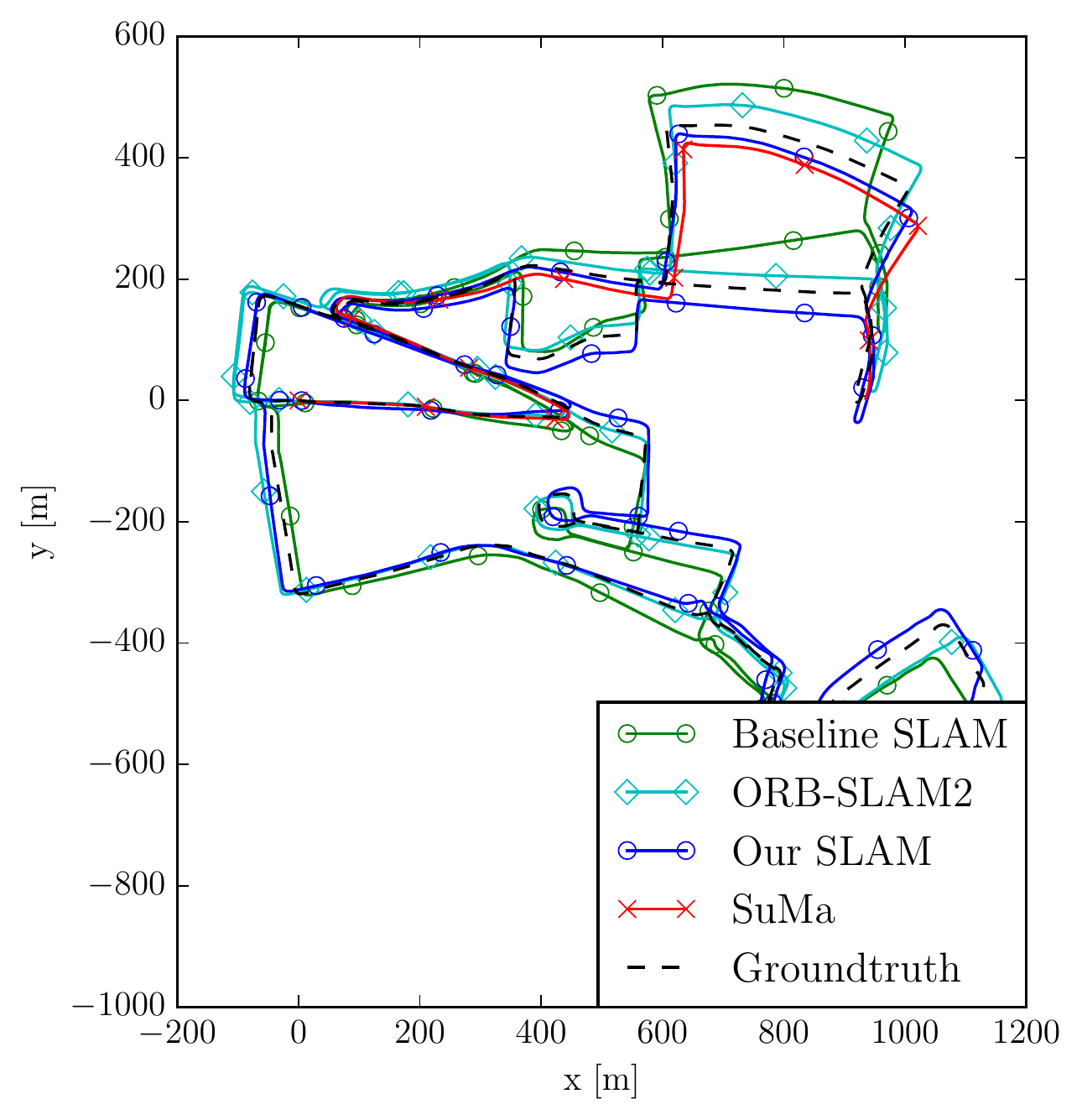}
    \caption{18-14-14-42}
    \end{subfigure}
    \caption{6 sequences trajectories results of different SLAM algorithms on Oxford Radar RobotCar Dataset.}
    \label{fig:oxford_trajectories_sequences}
\end{figure*}

\begin{figure*}[h]
    \centering
    \begin{subfigure}{0.24\linewidth}
    \includegraphics[width=\linewidth]{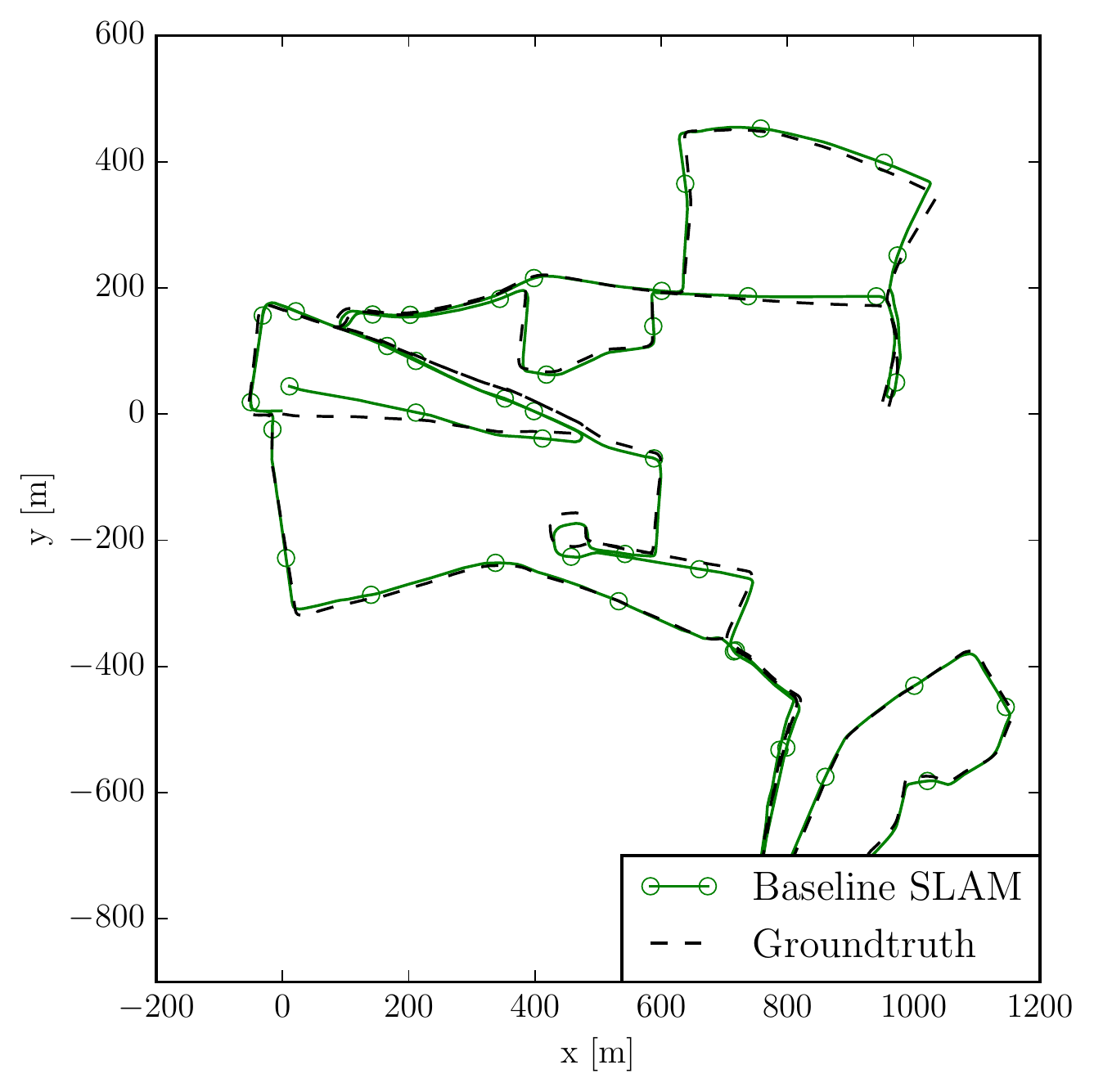}
    \caption{Baseline Radar SLAM}
    \end{subfigure}
    \begin{subfigure}{0.24\linewidth}
    \includegraphics[width=\linewidth]{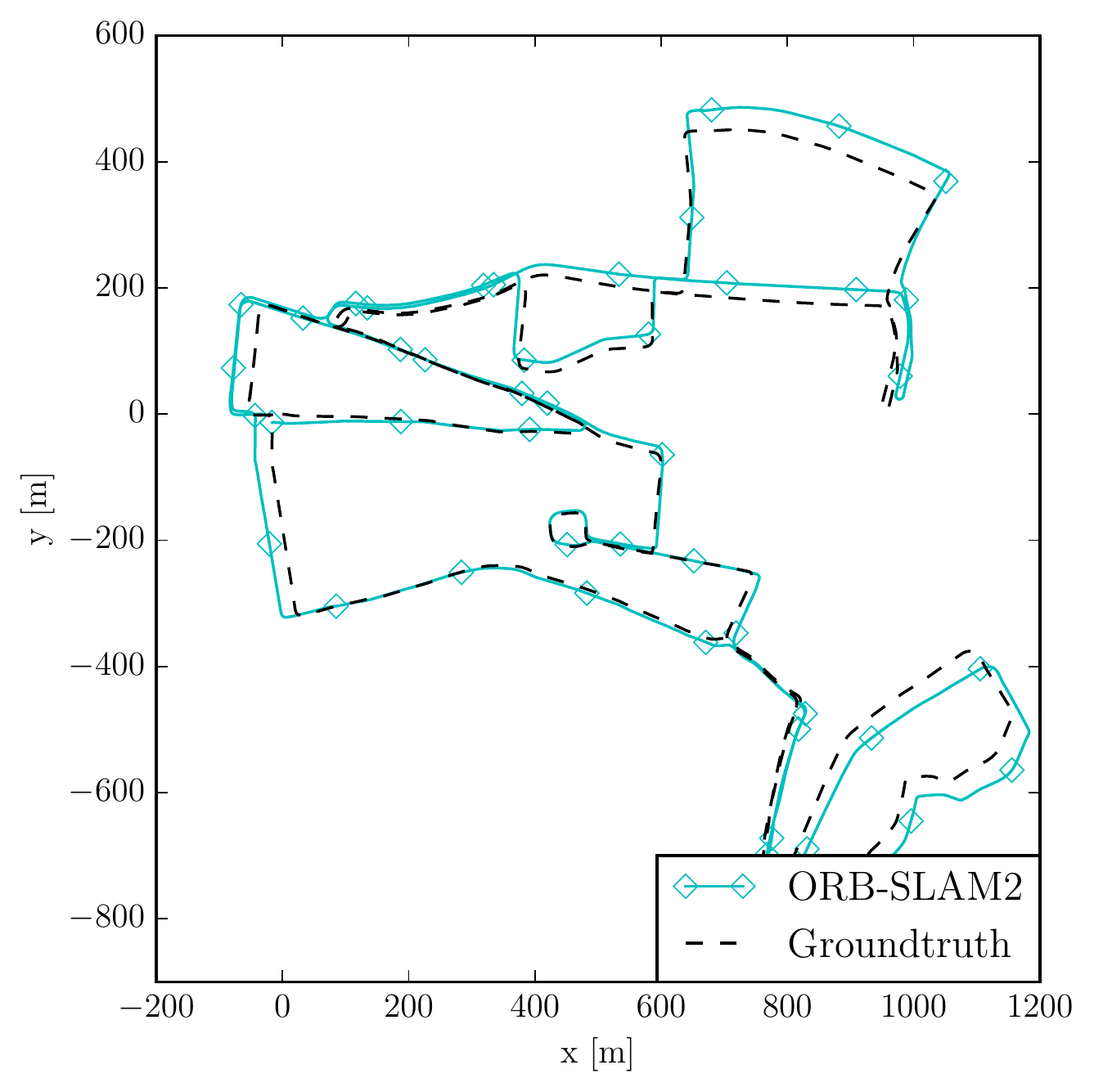}
    \caption{ORB-SLAM2 Stereo}
    \end{subfigure}
    \begin{subfigure}{0.24\linewidth}
    \includegraphics[width=\linewidth]{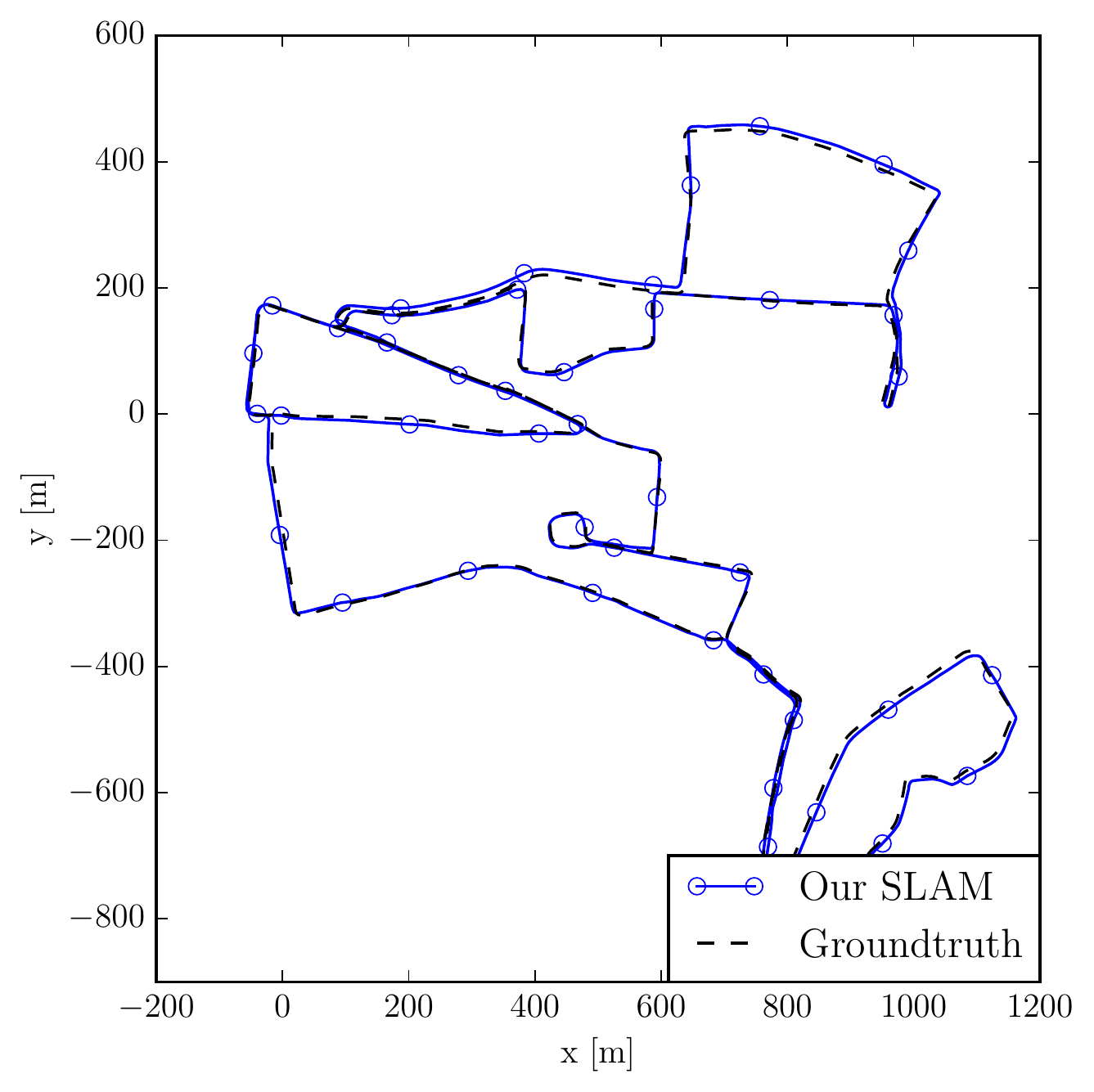}
    \caption{Our Radar SLAM}
    \end{subfigure}
    \begin{subfigure}{0.24\linewidth}
    \includegraphics[width=\linewidth]{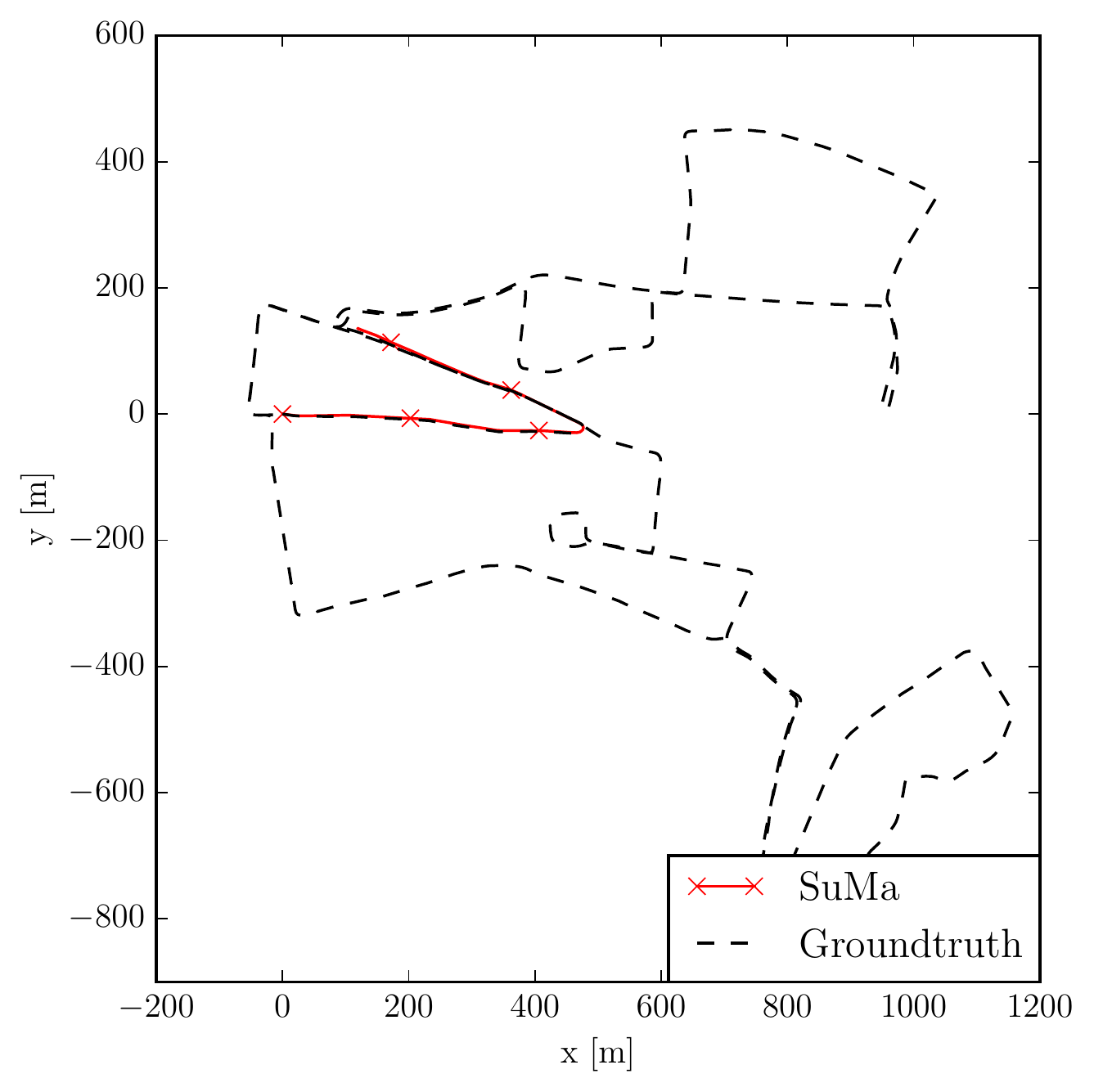}
    \caption{SuMa}
    \end{subfigure}
    \caption{Trajectories results of different SLAM algorithms on sequence 17-13-26-39 of Oxford Radar RobotCar Dataset.}
    \label{fig:oxford_trajectories}
\end{figure*}

We show the estimated trajectories of 6 sequences in Fig. \ref{fig:oxford_trajectories_sequences} for qualitative evaluation. For most of the sequences, our SLAM results are closest to the ground truth although the trajectories of baseline SLAM and ORB-SLAM2 are also accurate except for sequence 18-14-14-42. Fig. \ref{fig:oxford_trajectories} elaborate on the trajectory of each method on sequence 17-13-26-39 for qualitative performance.

\begin{figure*}[h]
    \centering
    \begin{subfigure}{0.32\linewidth}
    \includegraphics[width=\linewidth]{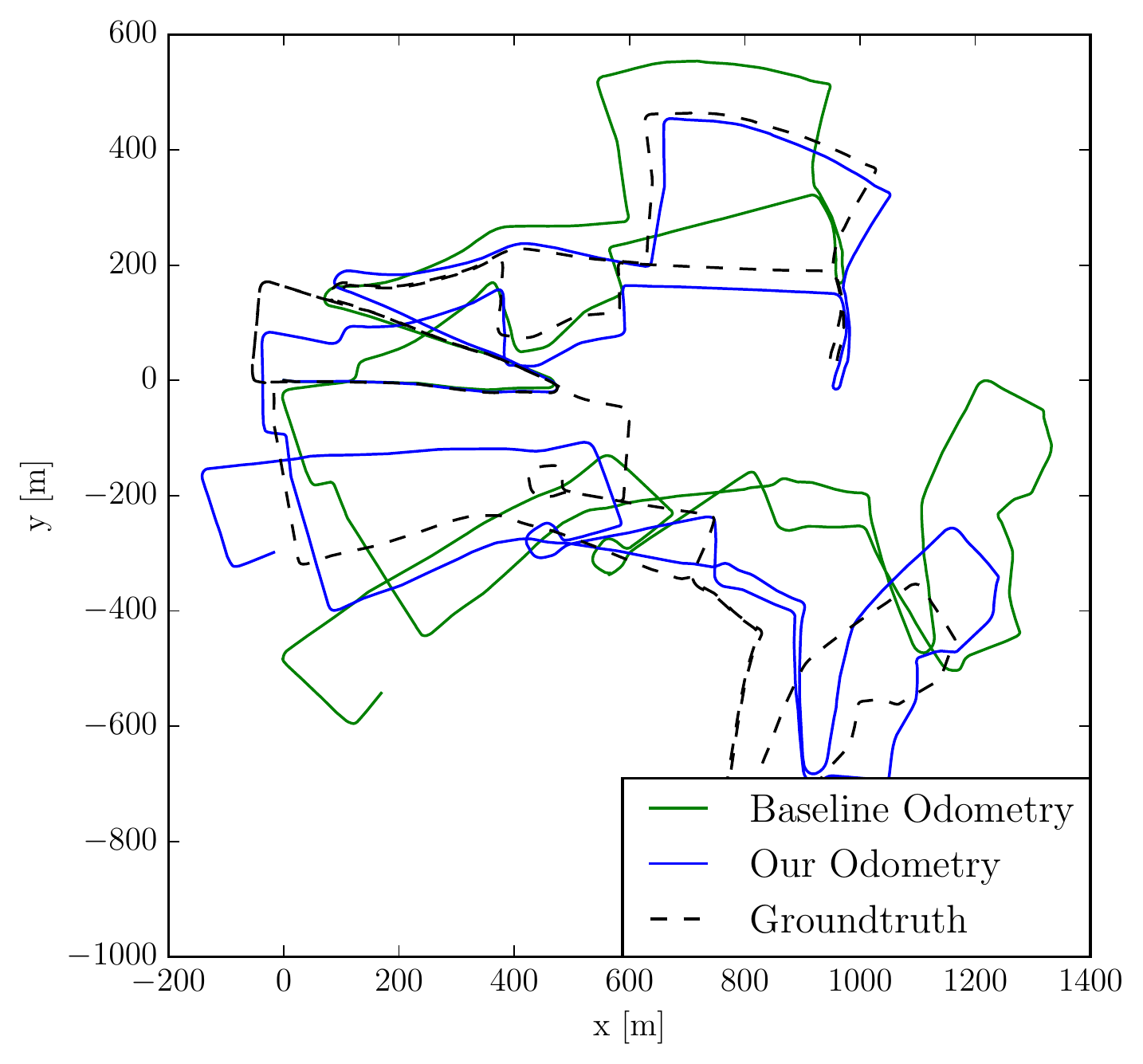}
    \caption{10-12-32-52}
    \end{subfigure}
    \begin{subfigure}{0.32\linewidth}
    \includegraphics[width=\linewidth]{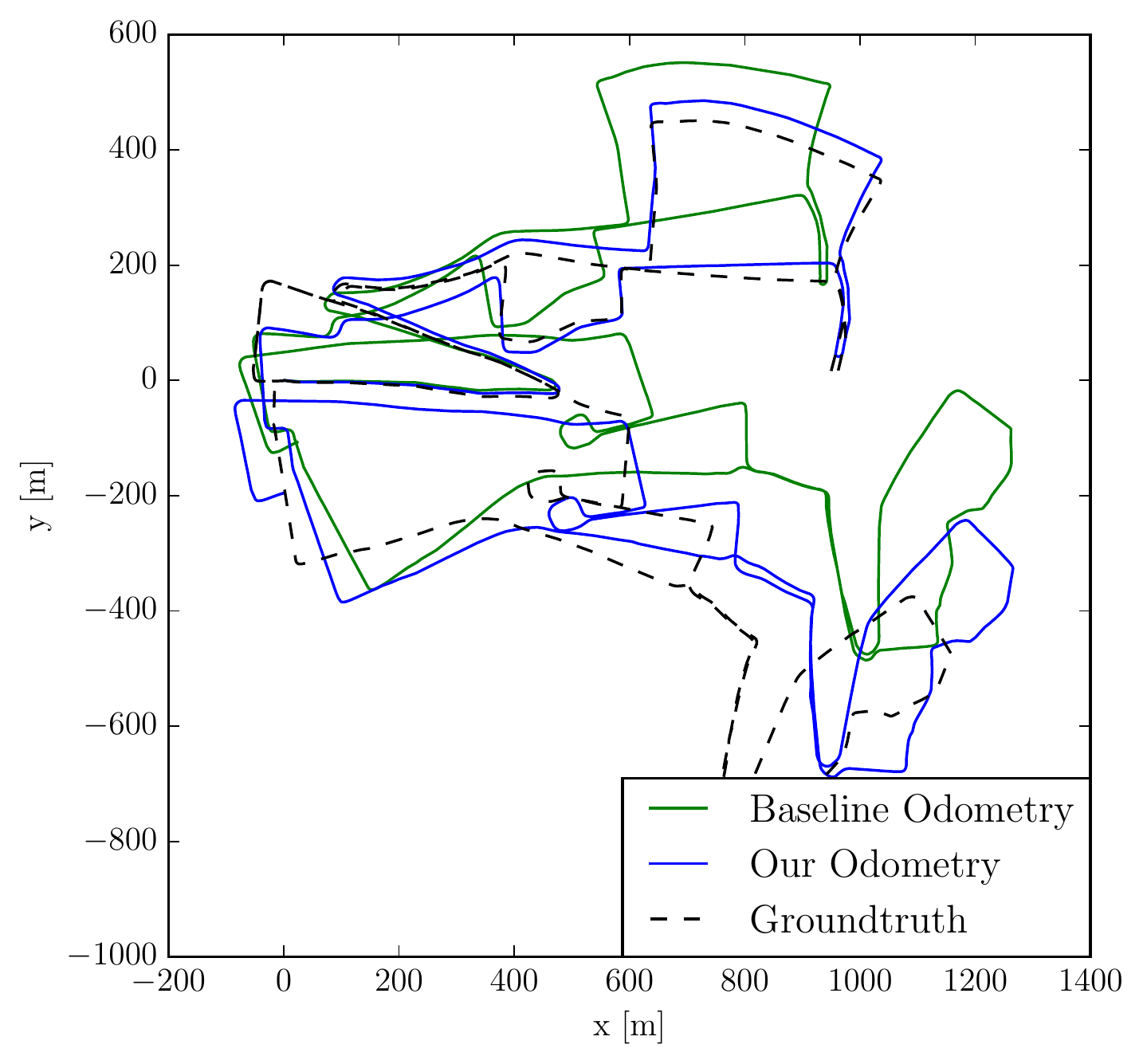}
    \caption{17-13-26-39}
    \end{subfigure}
    \begin{subfigure}{0.32\linewidth}
    \includegraphics[width=\linewidth]{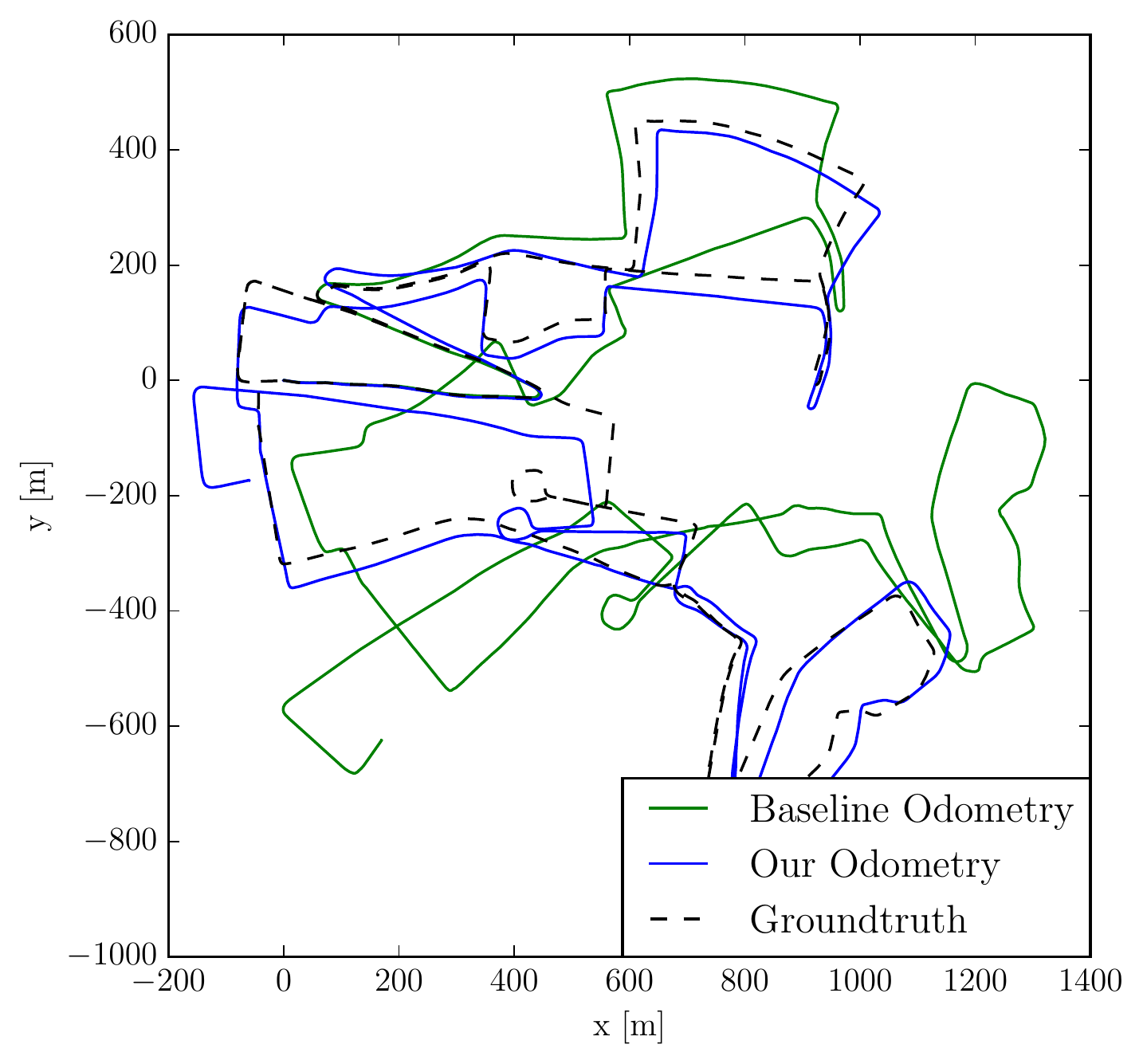}
    \caption{18-14-46-59}
    \end{subfigure}
    \caption{Trajectories of radar odometry results of different algorithms on Oxford Radar RobotCar Dataset.}
    \label{fig:oxford_odometry_baseline_our}
\end{figure*}

We further compare the proposed radar odometry with the baseline radar odometry \cite{hong2020radarslam}. Estimated trajectories of 3 sequences are presented in Fig. \ref{fig:oxford_odometry_baseline_our}. It is clear that our radar odometry drifts much slower than the baseline radar odometry method, validating the superior performance of the motion tracking algorithm with feature tracking and motion compensation. Therefore, our SLAM system also benefits from this improved accuracy.

\begin{table*}[t]
    \centering
    \caption{Relative error on MulRan Dataset}
    \footnotesize
    \begin{tabular}{ l|p{1.0cm}p{1.0cm}p{1.0cm}p{1.0cm}p{1.0cm}p{1.0cm}p{1.1cm}p{1.1cm}p{1.1cm}p{1.0cm}}
    \hline
    & \multicolumn{10}{c}{\textbf{Sequence}} \\
    \textbf{Method} & DCC01 & DCC02 & DCC03 & KAIST01 & KAIST02 & KAIST03 & Riverside01 & Riverside02 & Riverside03 & Mean \\
    \hline
    SuMa &2.71/0.4 & 4.07/0.9 &2.14/0.6 &2.9/0.8 &2.64/0.6 &2.17/0.6 & 1.66/0.6$\color{red}^{*30\%}$ & 1.49/0.5$\color{red}^{*23\%}$ & 1.65/0.4$\color{red}^{*5\%}$ & 2.38/0.5 \\
    Baseline Odometry &3.35/0.9 &2.12/0.6 &1.74/0.6 &2.32/0.8 &2.69/1.0 &2.62/0.8 &2.70/0.7 &3.09/1.1 &2.71/0.7 & 2.59/0.8\\
    Baseline SLAM     &3.81/0.9 &2.04/0.5 &1.90/5.5 &2.34/0.7 &1.95/0.6 &20.1/5.1 &3.56/0.9 &3.05/6.8 &152/0.175 &21.1/2.3\\
    Our Odometry &2.70/0.5 &1.90/0.4 &1.64/0.4 &2.13/0.7 &2.07/0.6 &1.99/0.5 &2.04/0.5 &1.51/0.5 &1.71/0.5 &1.97/0.5\\
    Our SLAM &2.39/0.4 &1.90/0.4 &1.56/0.2 &1.75/0.5 &1.76/0.4 &1.72/0.4 &3.40/0.9 &1.79/0.3 &1.95/0.5 &2.02/0.4\\
    \hline
    \end{tabular}\par
    \smallskip
    Results are given as \textit{translation error} / \textit{rotation error}. Translation error is in \%, and rotation error is in degrees per 100 meters (deg/100m). $\color{red}^{*}$ indicates the algorithm fails at the $xx\%$ of the sequence and its result is reported up to that point.
    \label{tab:mulran_relative_error}
\end{table*}

\begin{table*}[t]
    \centering
    \caption{Absolute trajectory error for position (RMSE) on MulRan Dataset}
    \footnotesize
    \begin{tabular}{ l|p{1.1cm}p{1.1cm}p{1.1cm}p{1.1cm}p{1.1cm}p{1.1cm}p{1.4cm}p{1.4cm}p{1.4cm}}
    \hline
    & \multicolumn{9}{c}{\textbf{Sequence}} \\
    \textbf{Method} & DCC01 & DCC02 & DCC03 & KAIST01 & KAIST02 & KAIST03 & Riverside01 & Riverside02 & Riverside03 \\
    \hline
    SuMa   &13.509 & 17.834 & 29.574 & 38.693 & 31.864 & 45.970 & N/A & N/A &  N/A   \\
    Baseline SLAM &     17.458 & 24.962 & 76.138 & 4.931 & 3.918 & 50.809 & 10.531 & 95.247 &  1091.605 \\
    Our SLAM & \textbf{12.886} & \textbf{9.878} & \textbf{3.917} & \textbf{6.873} & \textbf{6.028} & \textbf{4.109} & \textbf{9.029} & \textbf{7.049} &  \textbf{10.741}   \\
    \hline
    \end{tabular}\par
    \smallskip
    The absolute trajectory error of position is in meters. N/A: SuMa fails to finish Riverside01, 02 and 03 sequences.
    \label{tab:mulran_ate_error}
\end{table*}

\begin{figure*}[h]
    \centering
    \begin{subfigure}{0.3\textwidth}
    \includegraphics[width=\linewidth]{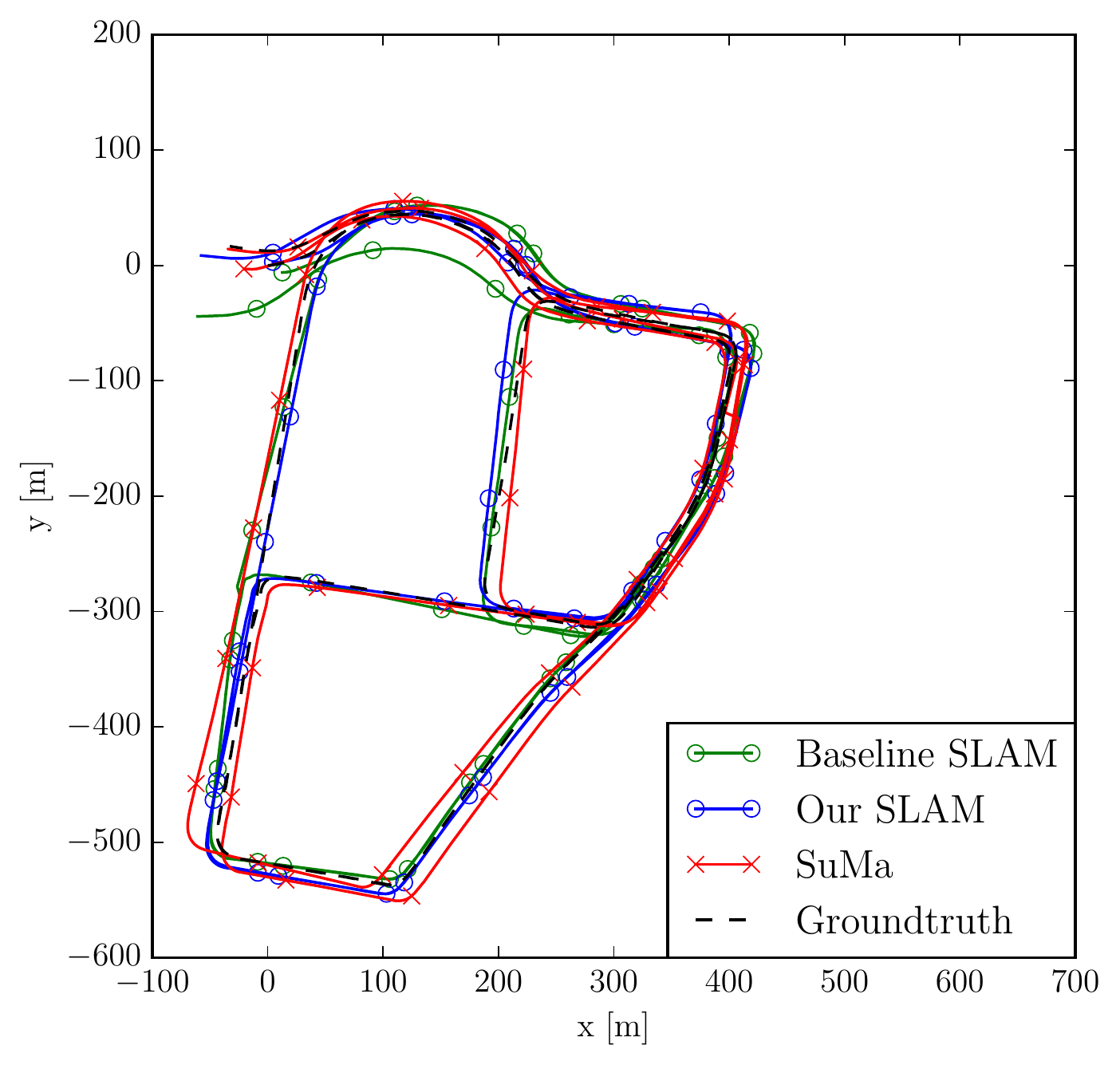}
    \caption{DCC01}
    \end{subfigure}
    \begin{subfigure}{0.3\textwidth}
    \includegraphics[width=\linewidth]{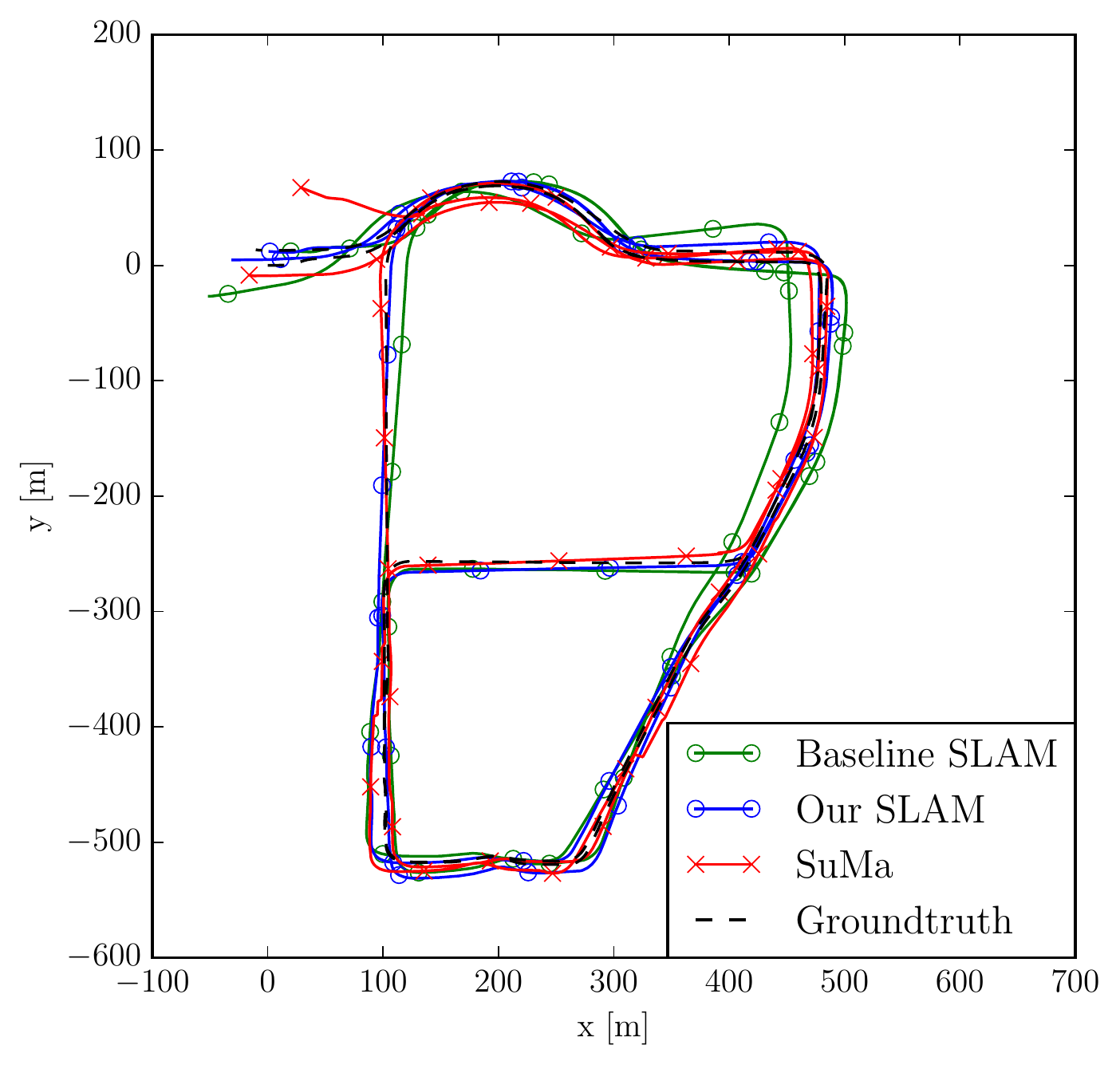}
    \caption{DCC02}
    \end{subfigure}
    \begin{subfigure}{0.3\textwidth}
    \includegraphics[width=\linewidth]{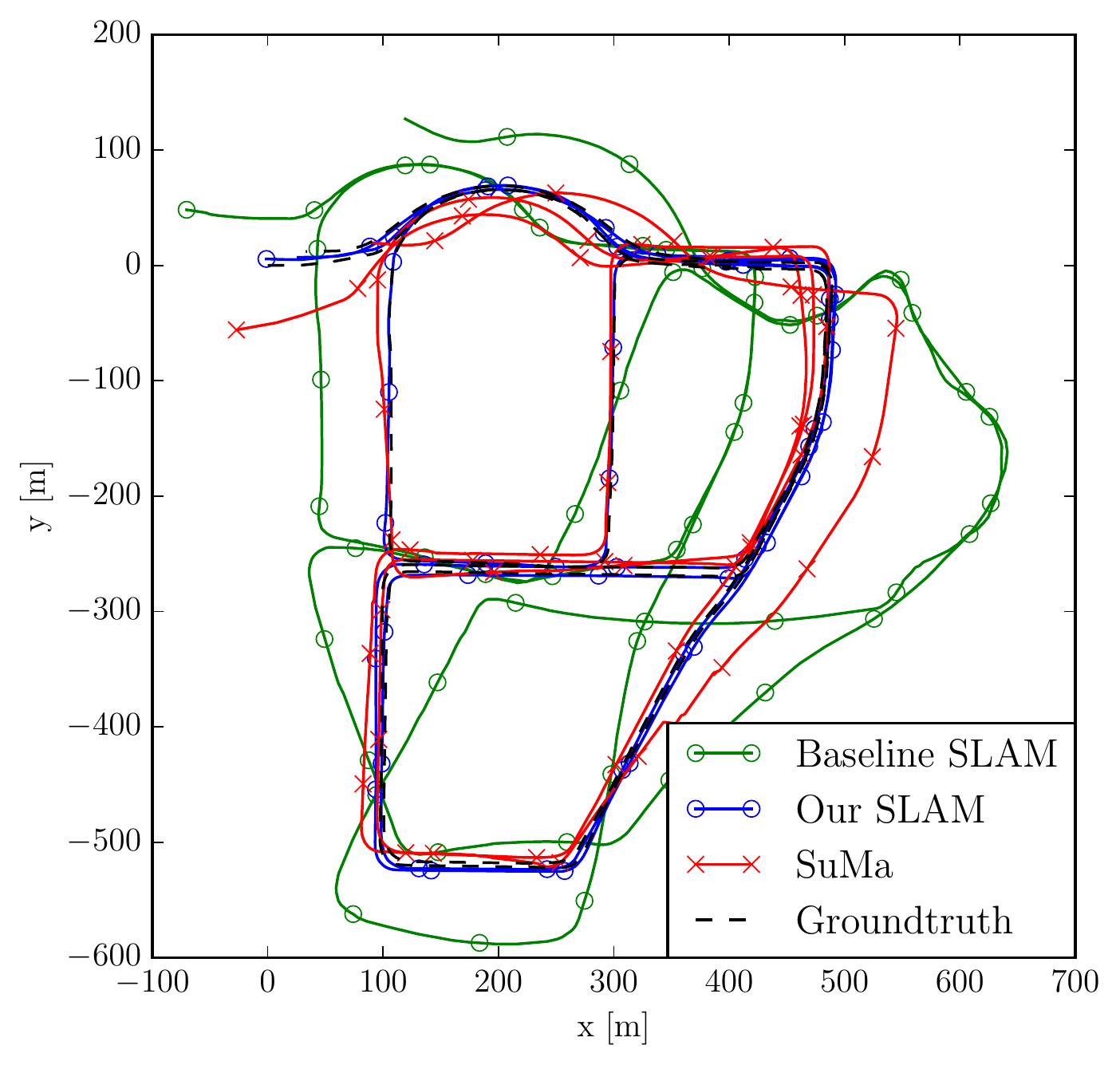}
    \caption{DCC03}
    \end{subfigure}
    \hfill
    \begin{subfigure}{0.3\textwidth}
    \includegraphics[width=\linewidth]{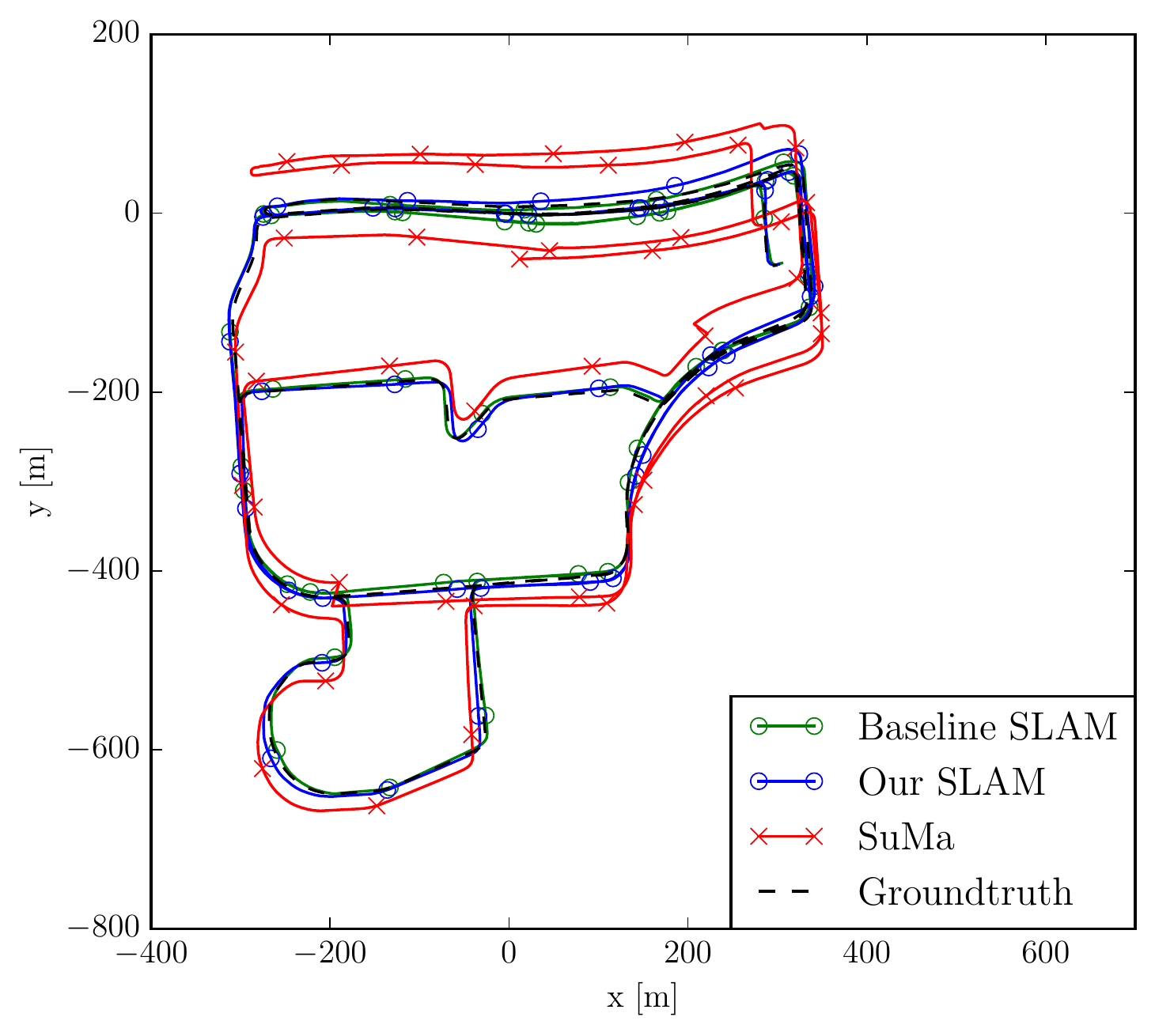}
    \caption{KAIST01}
    \end{subfigure}
    \begin{subfigure}{0.3\textwidth}
    \includegraphics[width=\linewidth]{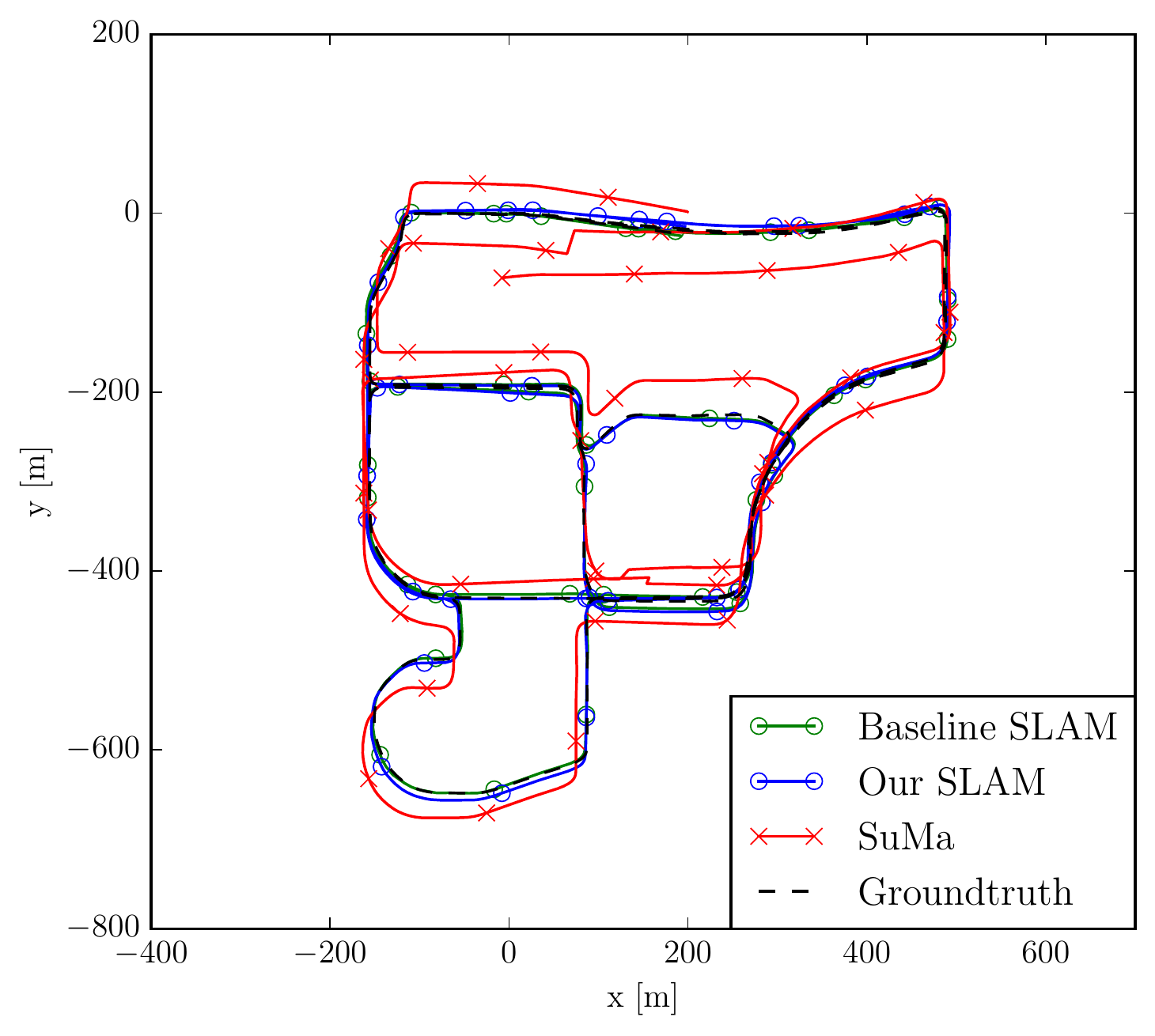}
    \caption{KAIST02}
    \end{subfigure}
    \begin{subfigure}{0.3\textwidth}
    \includegraphics[width=\linewidth]{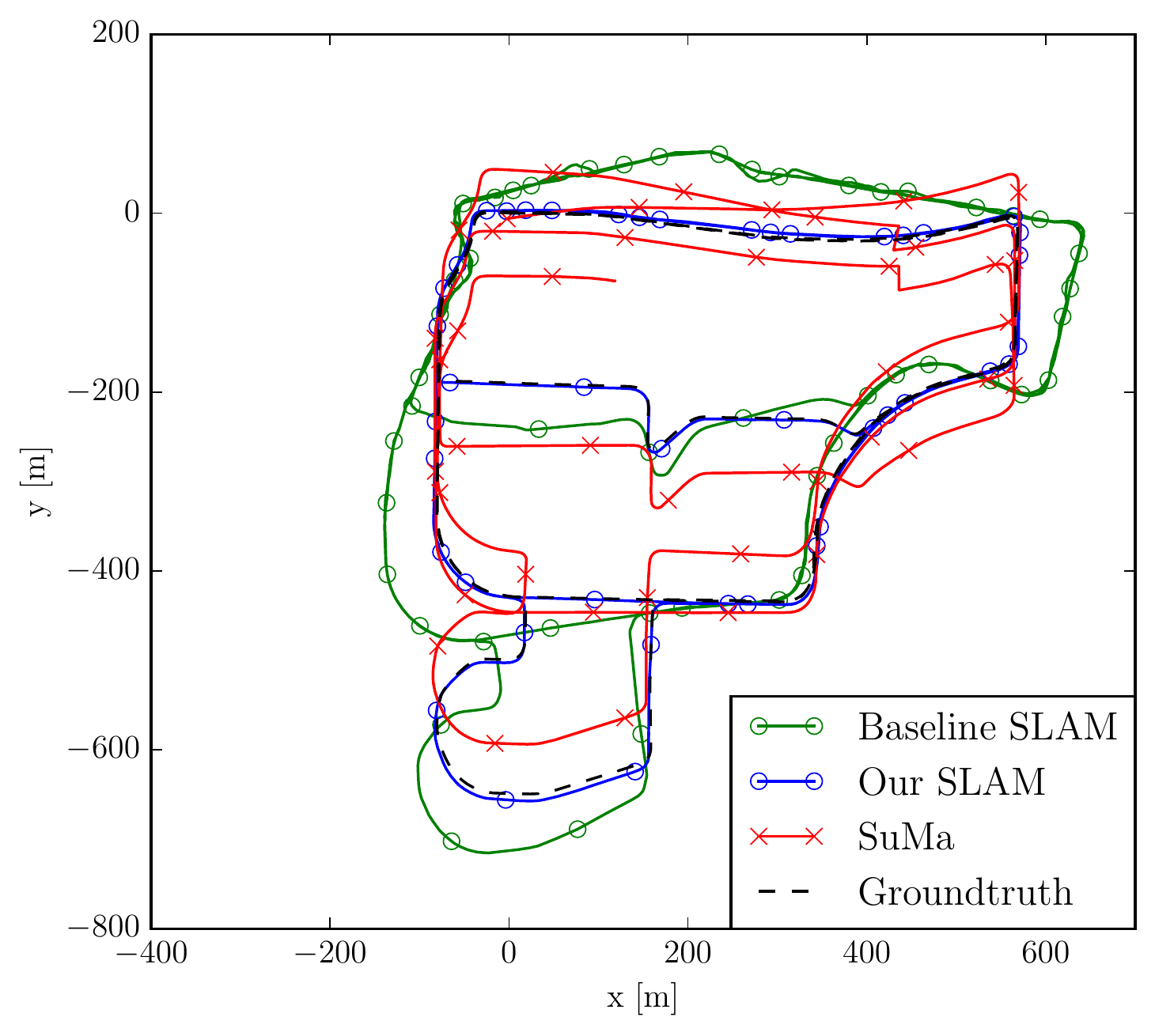}
    \caption{KAIST03}
    \end{subfigure}
    \caption{6 sequences trajectories results of different SLAM algorithms on MulRan Dataset.}
    \label{fig:mulran_sequences_trajectories}
\end{figure*}

\begin{figure*}[h]
    \begin{subfigure}{0.24\linewidth}
    \includegraphics[width=\linewidth]{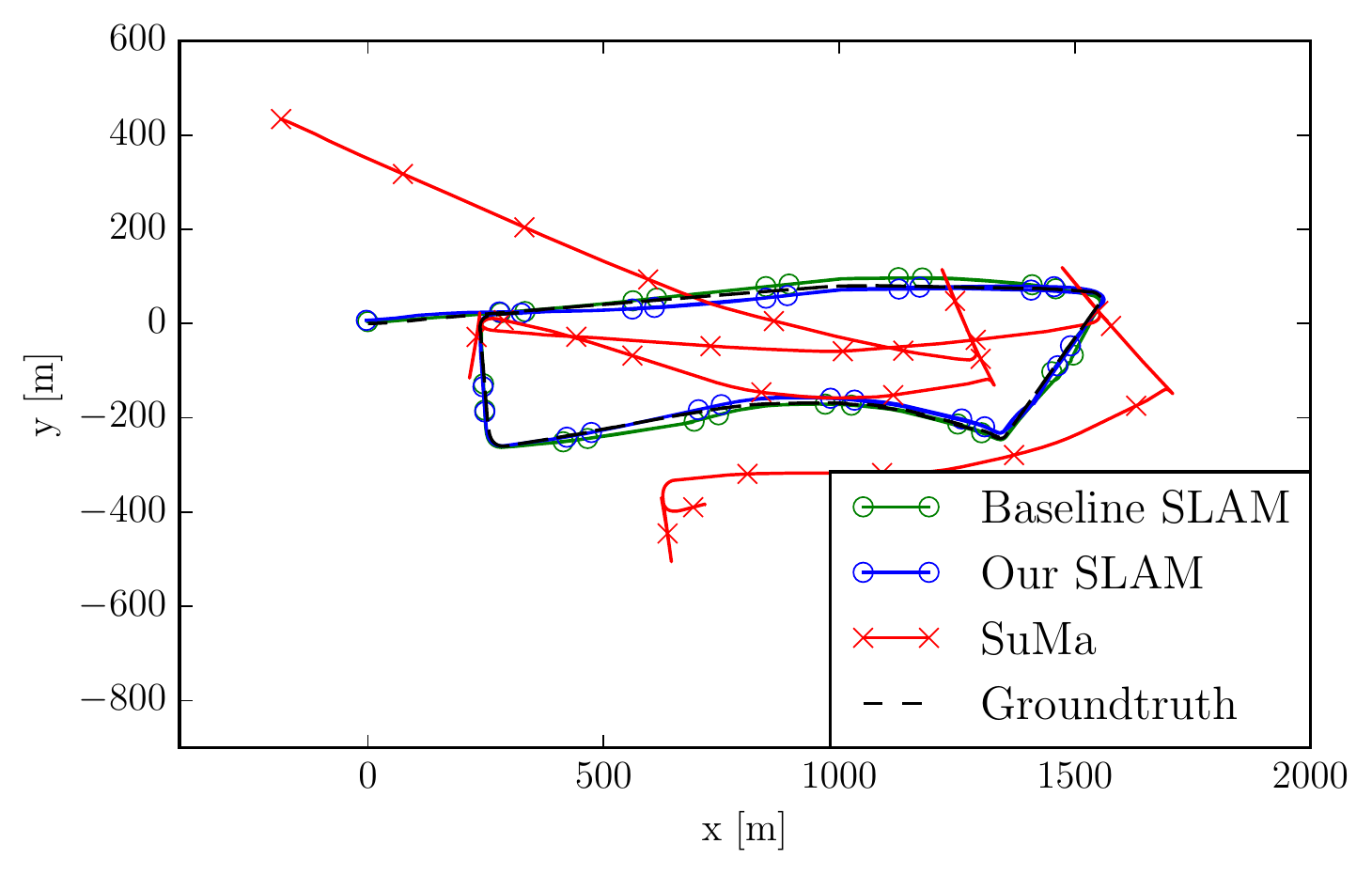}
    \caption{Comparision}
    \end{subfigure}
    \begin{subfigure}{0.24\linewidth}
    \includegraphics[width=\linewidth]{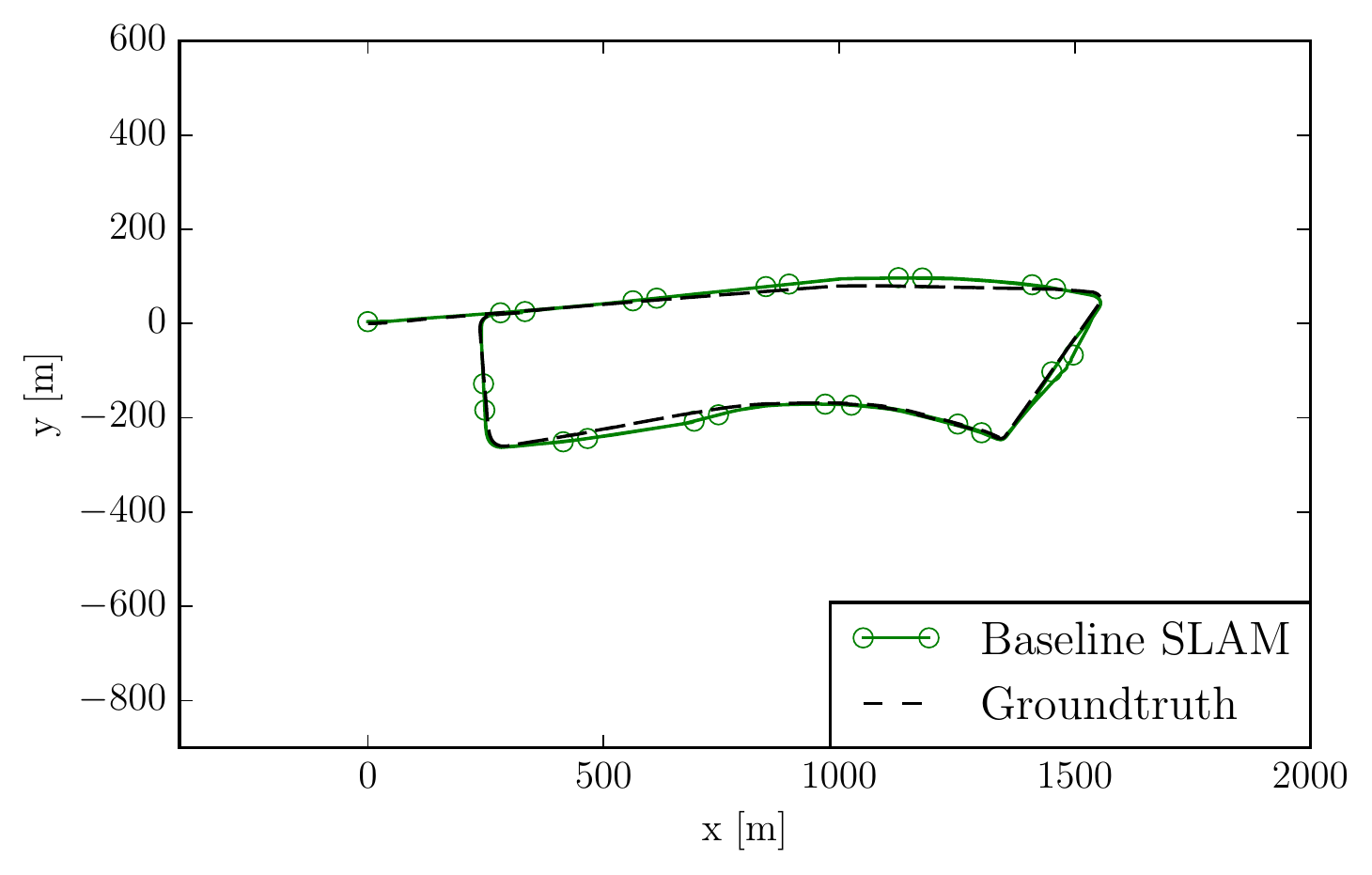}
    \caption{Baseline Radar SLAM}
    \end{subfigure}
    \begin{subfigure}{0.24\linewidth}
    \includegraphics[width=\linewidth]{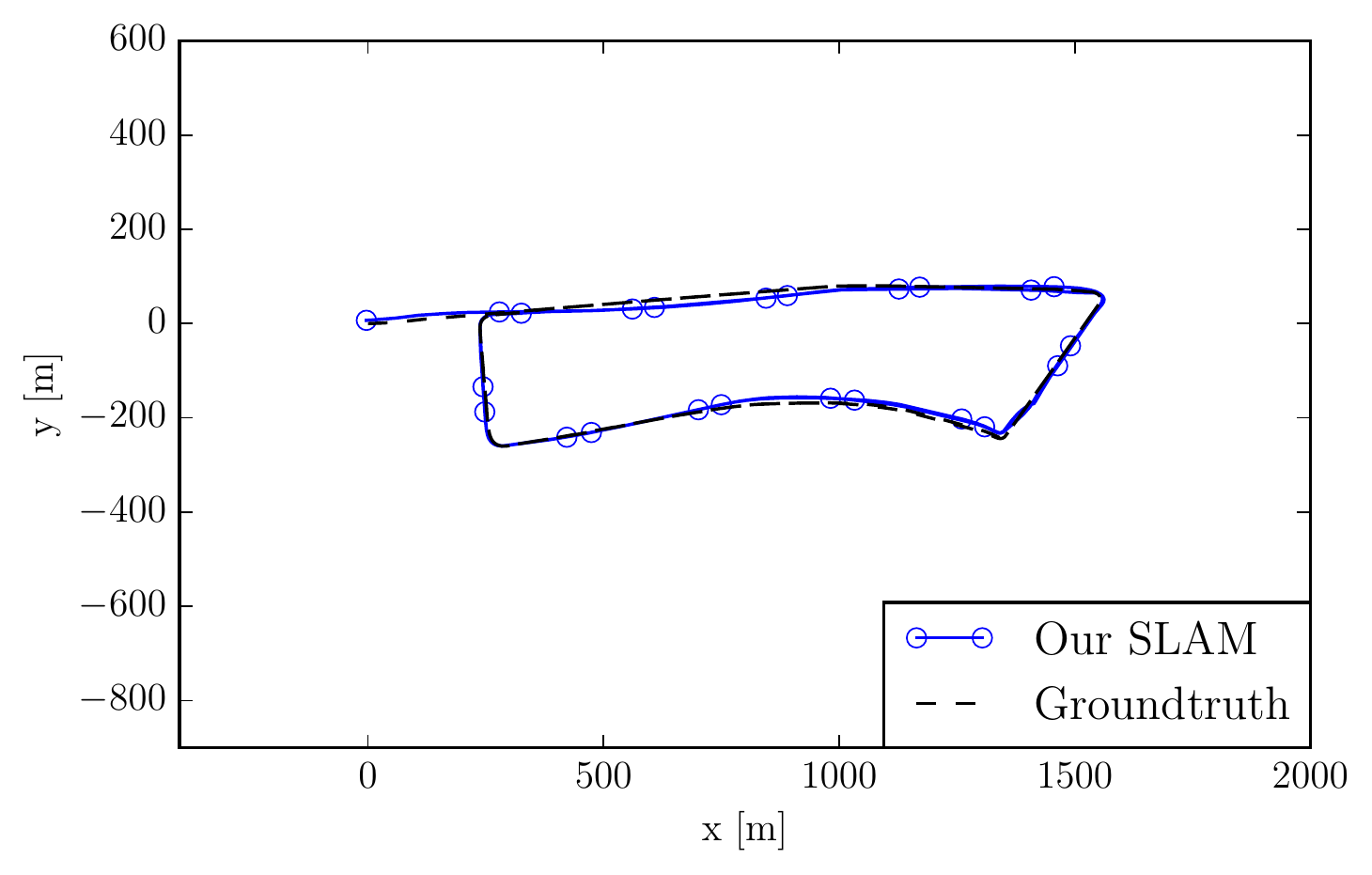}
    \caption{Our Radar SLAM}
    \end{subfigure}
    \begin{subfigure}{0.24\linewidth}
    \includegraphics[width=\linewidth]{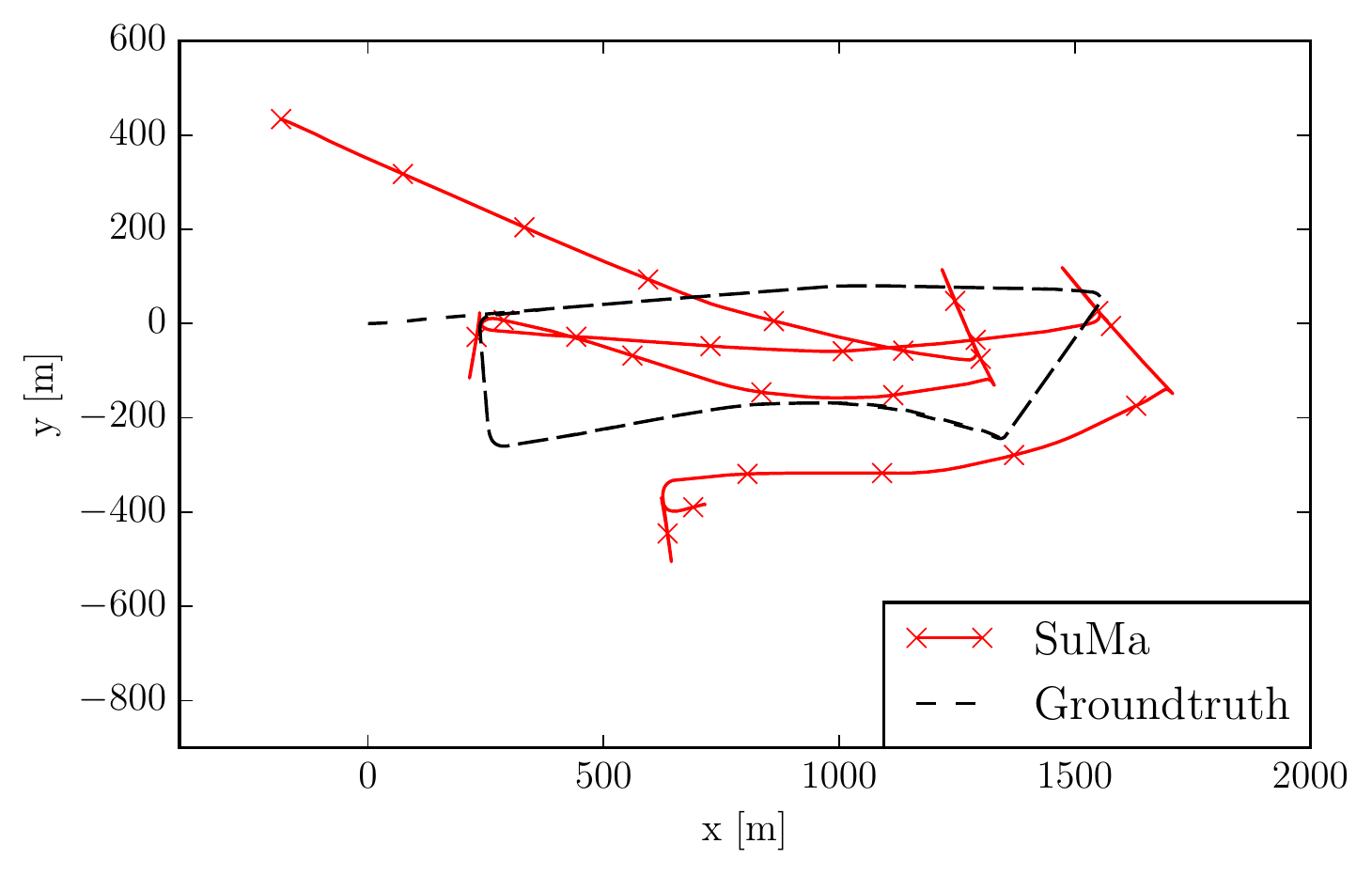}
    \caption{SuMa}
    \end{subfigure}
    \caption{Trajectories results of different SLAM algorithms on sequence Riverside01 of MulRan.}
    \label{fig:mulran_riverside01}
\end{figure*}

\subsection{Experiments on MulRun Dataset}

The RE and ATE of SuMa, baseline radar odometry/SLAM and our radar SLAM are shown in Table \ref{tab:mulran_relative_error} and Table \ref{tab:mulran_ate_error}. ORB-SLAM2 is not applicable here since MulRan only contains radar and LiDAR data. Similar to the RobotCat dataset, we again transform its provided 6-DoF ground truth poses into 3-DoF for evaluation. Both RE and ATE are evaluated on 9 sequences: DCC01, DCC02, DCC03, KAIST01, KAIST02, KAIST03, Riverside01, Riverside02 and Riverside03. In terms of RE, both our odometry and SLAM system achieve comparable or better performance on all sequences.

\begin{figure}[h]
    \centering
    \includegraphics[width=\linewidth]{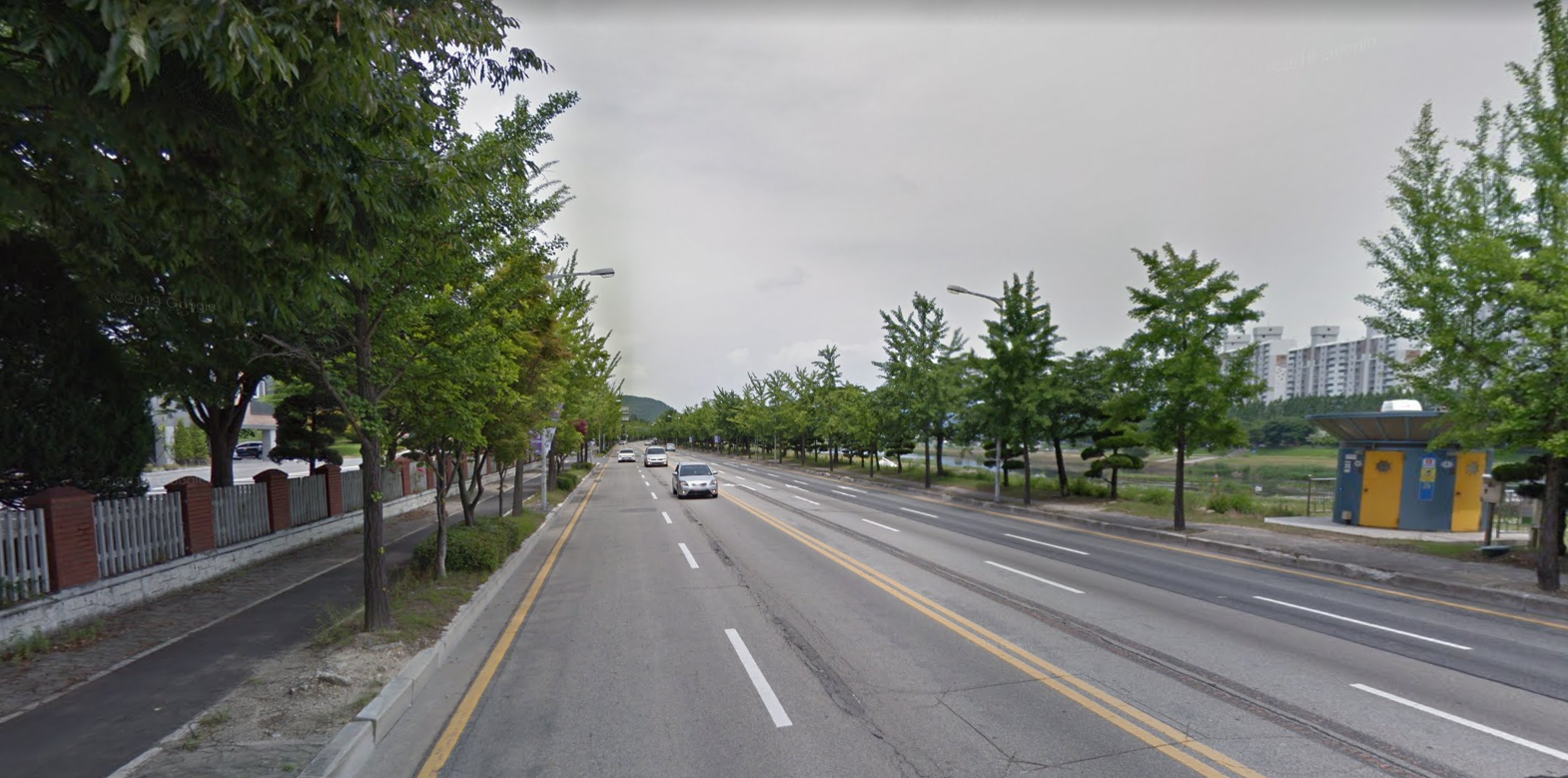}
    \caption{Riverside scenery of MulRan from Google Street View: repetitive structures are challenging to LiDAR based methods moving on high way.}
    \label{fig:mulran_scenery}
\end{figure}


Our odometry method reduces both translation error and rotation errors significantly compared to the baseline. Our SLAM system, to a great extent, outperforms both baseline SLAM and SuMa on ATE. More importantly, only our SLAM reliably works on all 9 sequences which cover diverse urban environments. Specifically, the baseline SLAM detects wrong loop closures on sequence KAIST03, Riverside02 and Riverside03 and fails to detect a loop in DCC03, which causes its large ATEs for these sequences. SuMa, on the other hand, fails to finish the sequences Riverside01, 02 and 03, likely due to the challenges of less distinctive structures along the rather open and long road as shown in Fig. \ref{fig:mulran_scenery}. It can be very challenging to register LiDAR scans accurately in this kind of environment.

The estimated trajectories on sequences DCC01, DCC02, DCC03, KAIST01, KAIST02, KAIST03 are shown in Figure \ref{fig:mulran_sequences_trajectories}. These qualitative results of the algorithms provide similar observations to the RE and ATE. For clarity, Figure \ref{fig:mulran_riverside01} presents trajectories of the SLAM algorithms on Riverside01 in separate figures.


\subsection{Experiments on the RADIATE Dataset}

\begin{figure}[t!]
    \centering
    \begin{subfigure}{0.49\linewidth}
    \includegraphics[width=\linewidth]{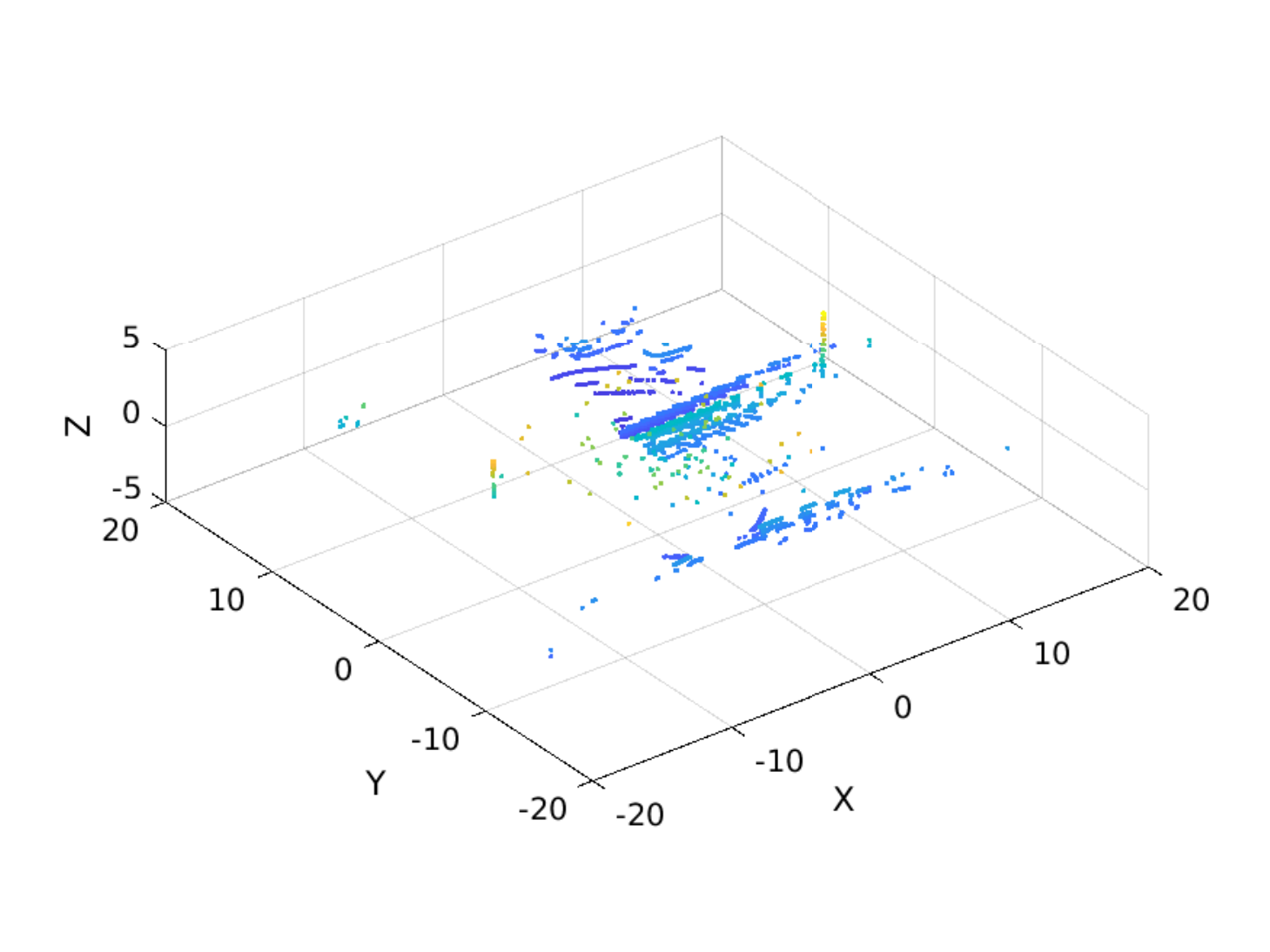}
    \caption{Place 1 snow}
    \end{subfigure}
    \begin{subfigure}{0.49\linewidth}
    \includegraphics[width=\linewidth]{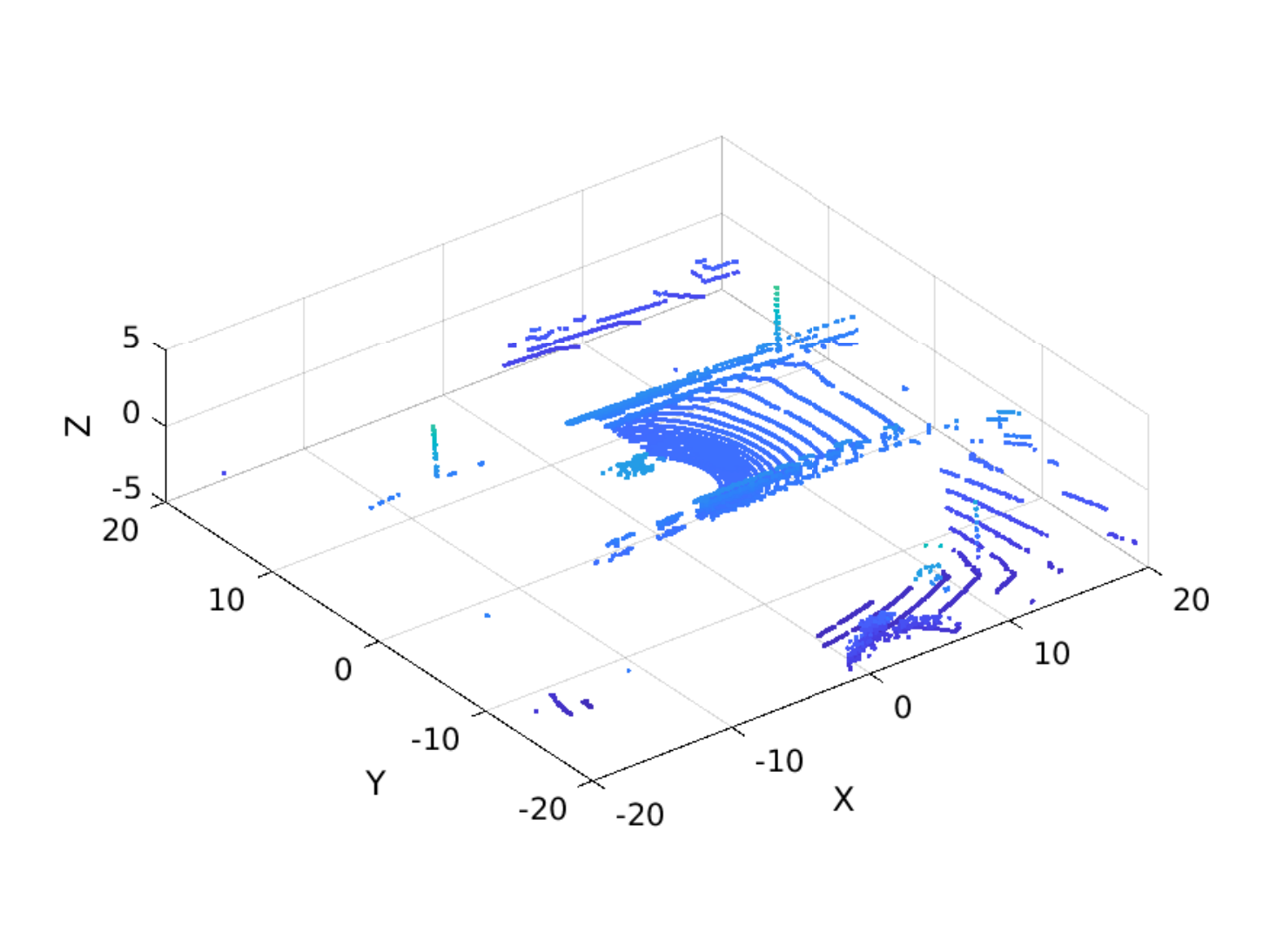}
    \caption{Place 1 normal}
    \end{subfigure}
    \begin{subfigure}{0.49\linewidth}
    \includegraphics[width=\linewidth]{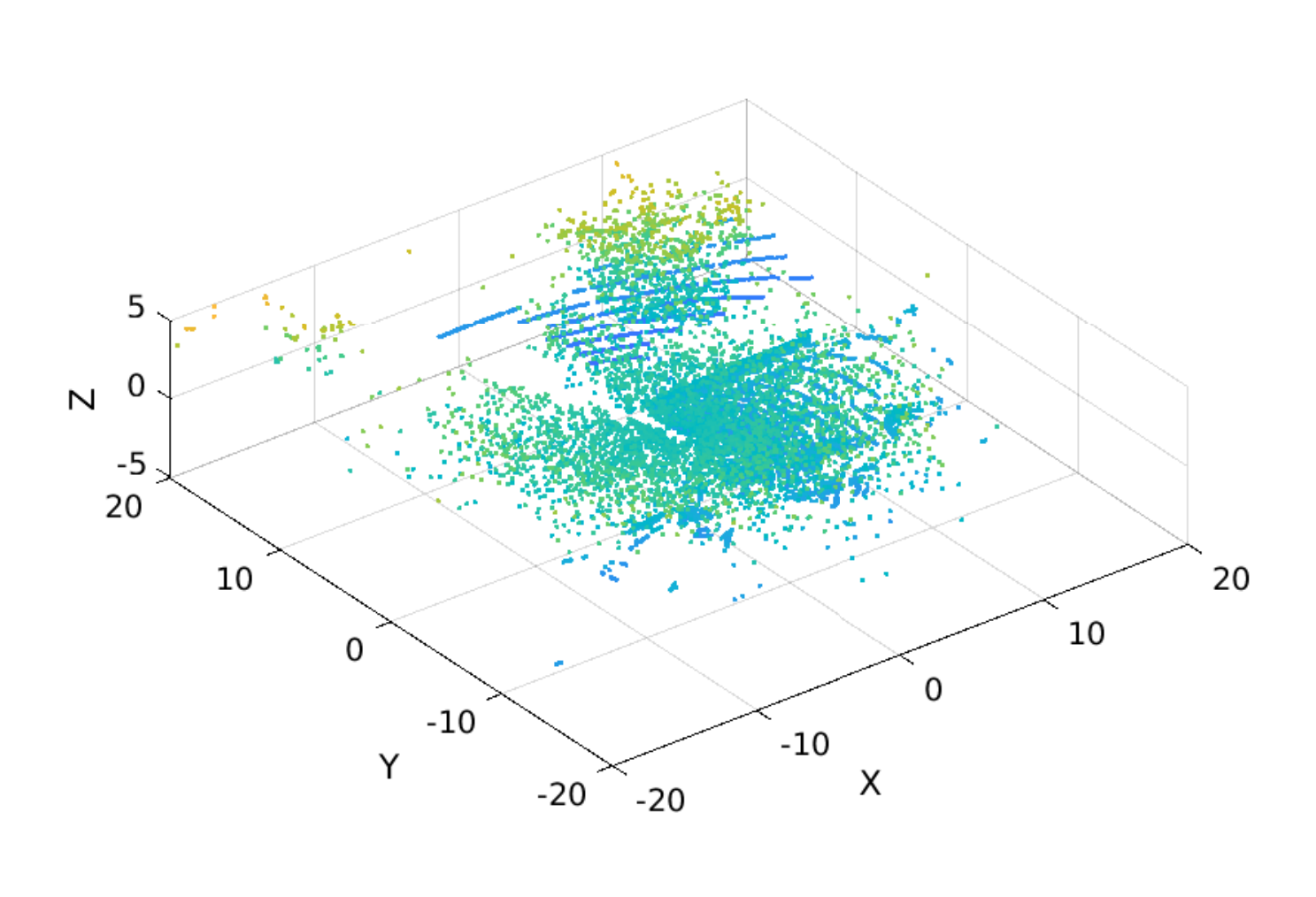}
    \caption{Place 2 snow}
    \end{subfigure}
    \begin{subfigure}{0.49\linewidth}
    \includegraphics[width=\linewidth]{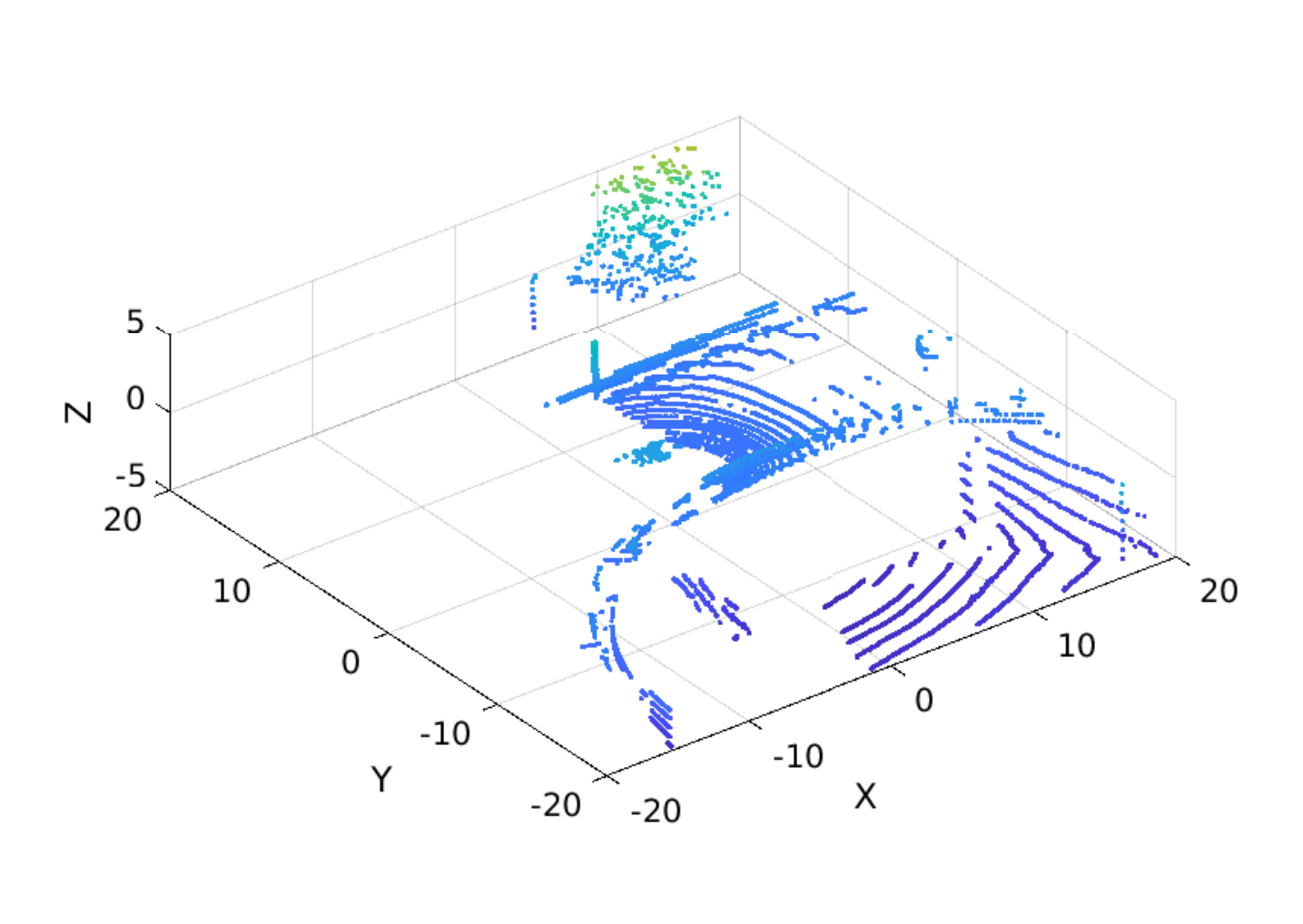}
    \caption{Place 2 normal}
    \end{subfigure}
    \caption{Two types of LiDAR degeneration. (a)-(b): Place 1 with less reflection from the scene. (c)-(d): Place 2 with many noisy detections from snowflakes around. }
    \label{fig:same_place_snow_norlmal}
\end{figure}
\begin{figure*}[t]
    \centering
    \begin{subfigure}{0.31\linewidth}
    \includegraphics[width=\linewidth]{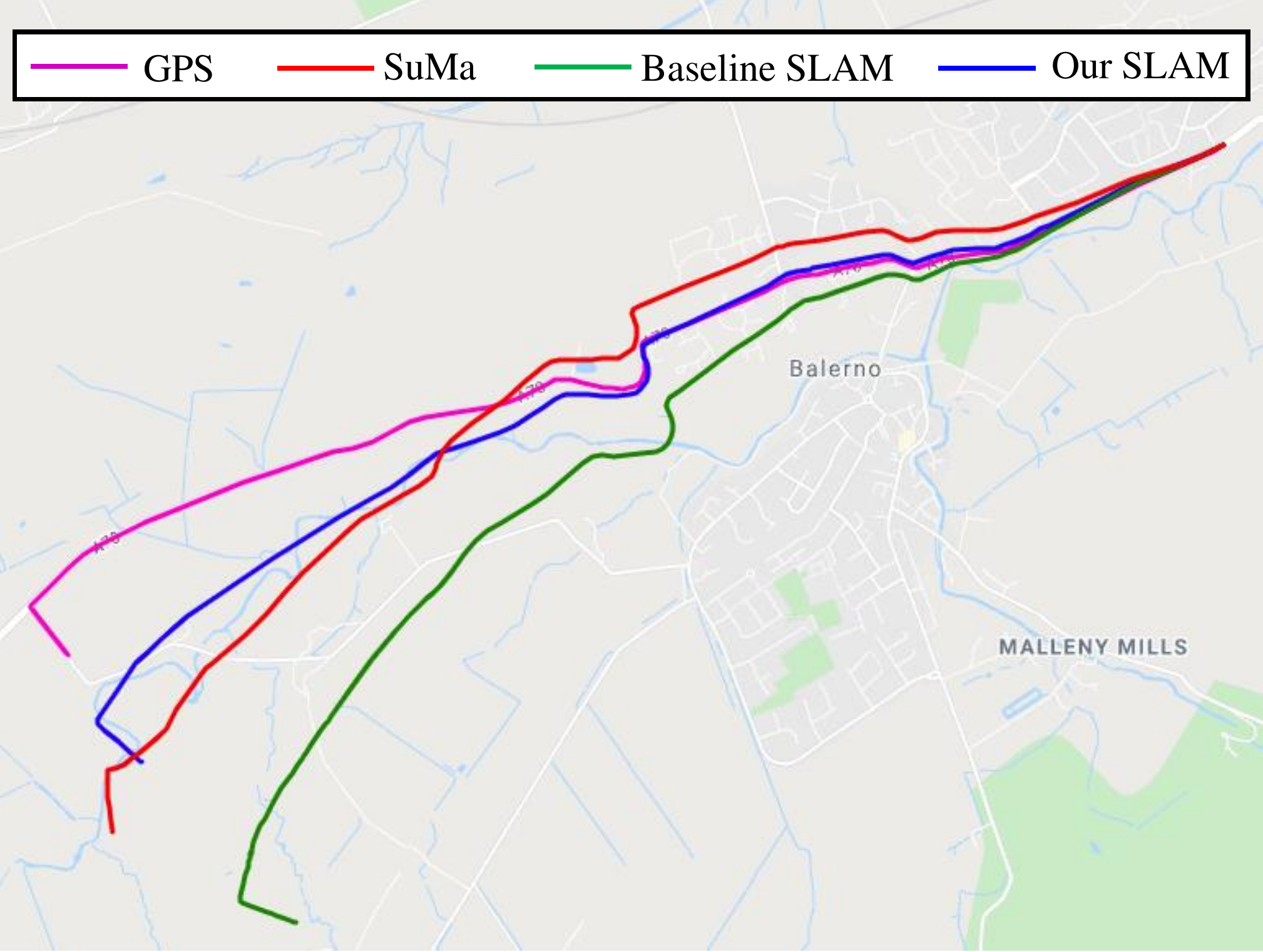}
    \caption{Fog 1}
    \label{fig:fog1_google}
    \end{subfigure}
    \begin{subfigure}{0.327\linewidth}
    \includegraphics[width=\linewidth]{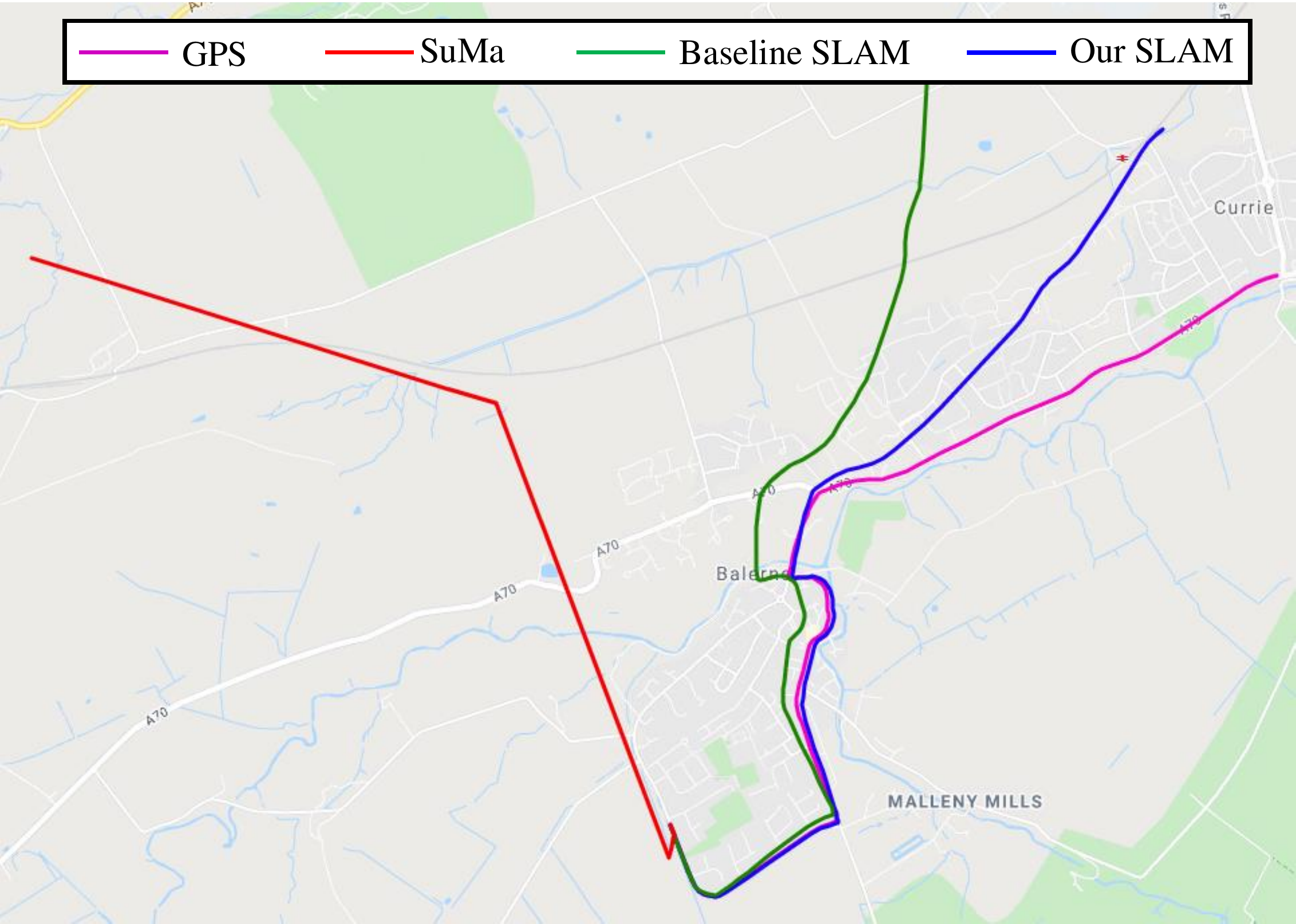}
    \caption{Fog 2}
    \label{fig:fog2_google}
    \end{subfigure}
    \begin{subfigure}{0.32\linewidth}
    \includegraphics[width=\linewidth]{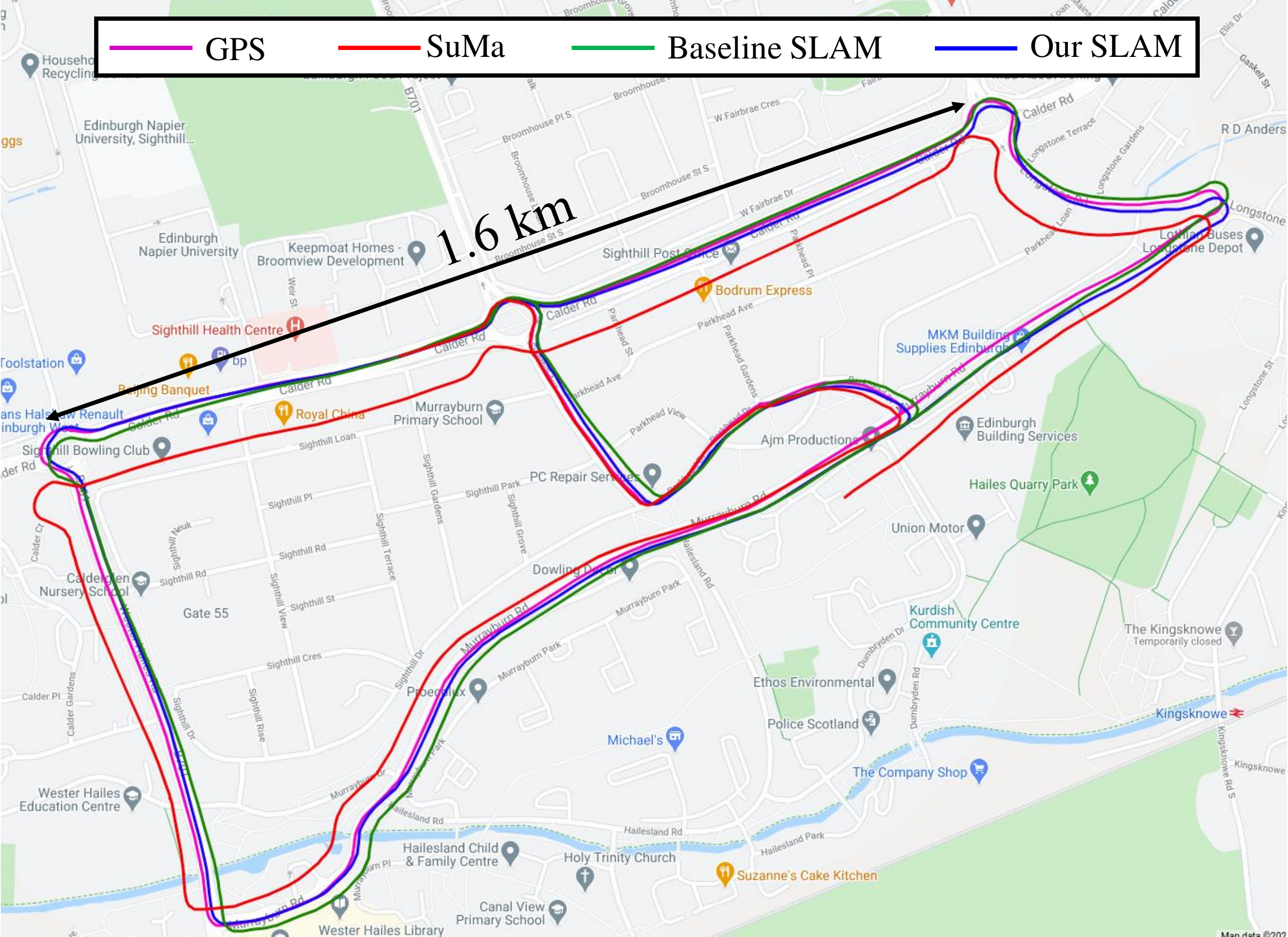}
    \caption{Night.}
    \label{fig:night_google}
    \end{subfigure}
    \caption{Estimated trajectories and groundtruth of sequence Fog 1, Fog 2 and Night.}

\end{figure*}

\begin{figure*}[t]
    \centering
    \includegraphics[width=\linewidth]{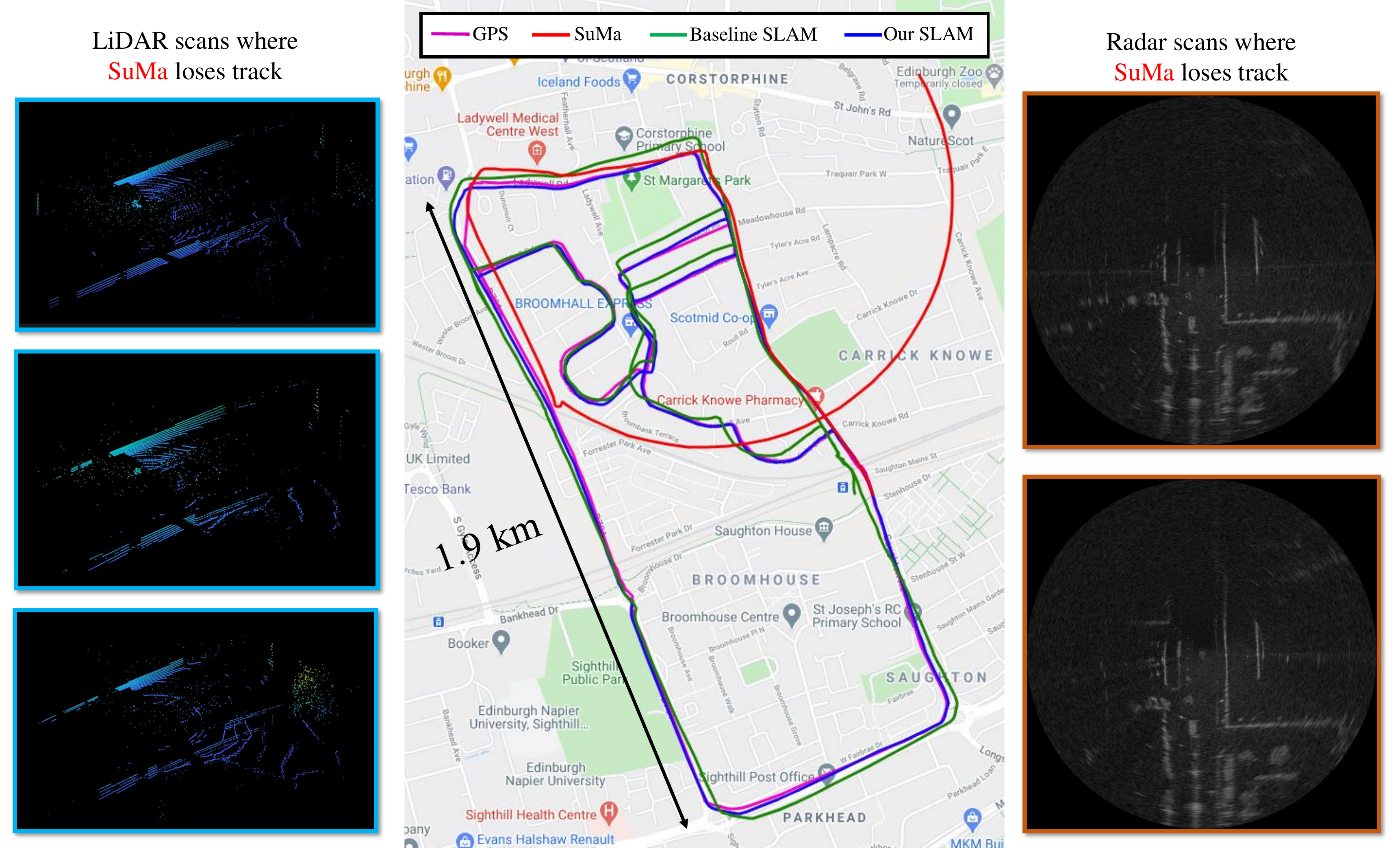}
    \caption{Results on the Snow sequence of the RADIATE dataset. Left: LiDAR scans when SuMa loses track. Note the noisy LiDAR reflection of snowflakes. Middle: GPS and estimated trajectories on Google Map.
    Right: Radar images when SuMa loses track.}
    \label{fig:snow_google}
\end{figure*}

\begin{figure*}[t]
    \centering

    \begin{subfigure}{0.28\linewidth}
    \includegraphics[width=\linewidth]{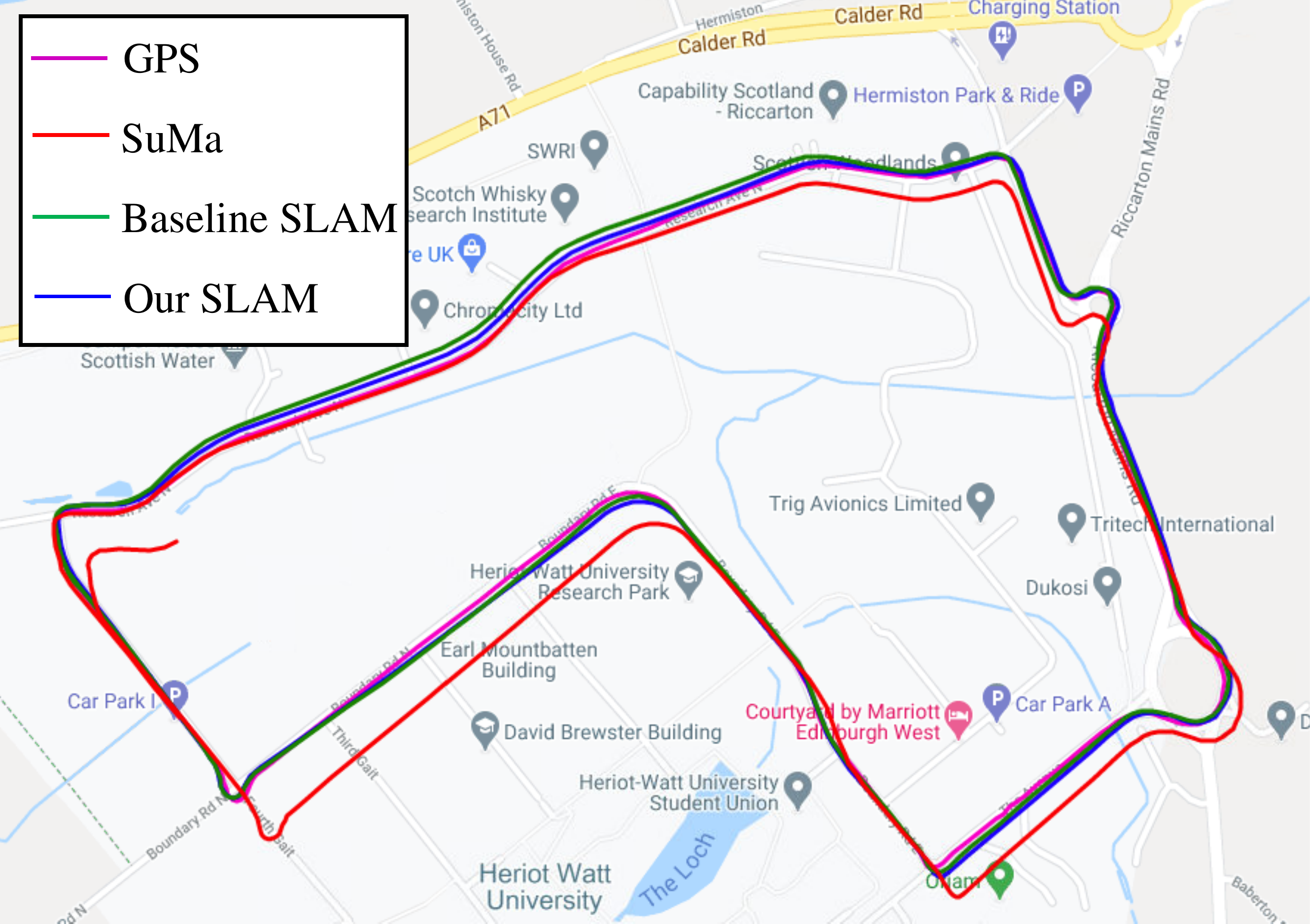}
    \caption{Normal weather sequence}
    \label{fig:good_weather_google}
    \end{subfigure}
    \begin{subfigure}{0.255\linewidth}
    \includegraphics[width=\linewidth]{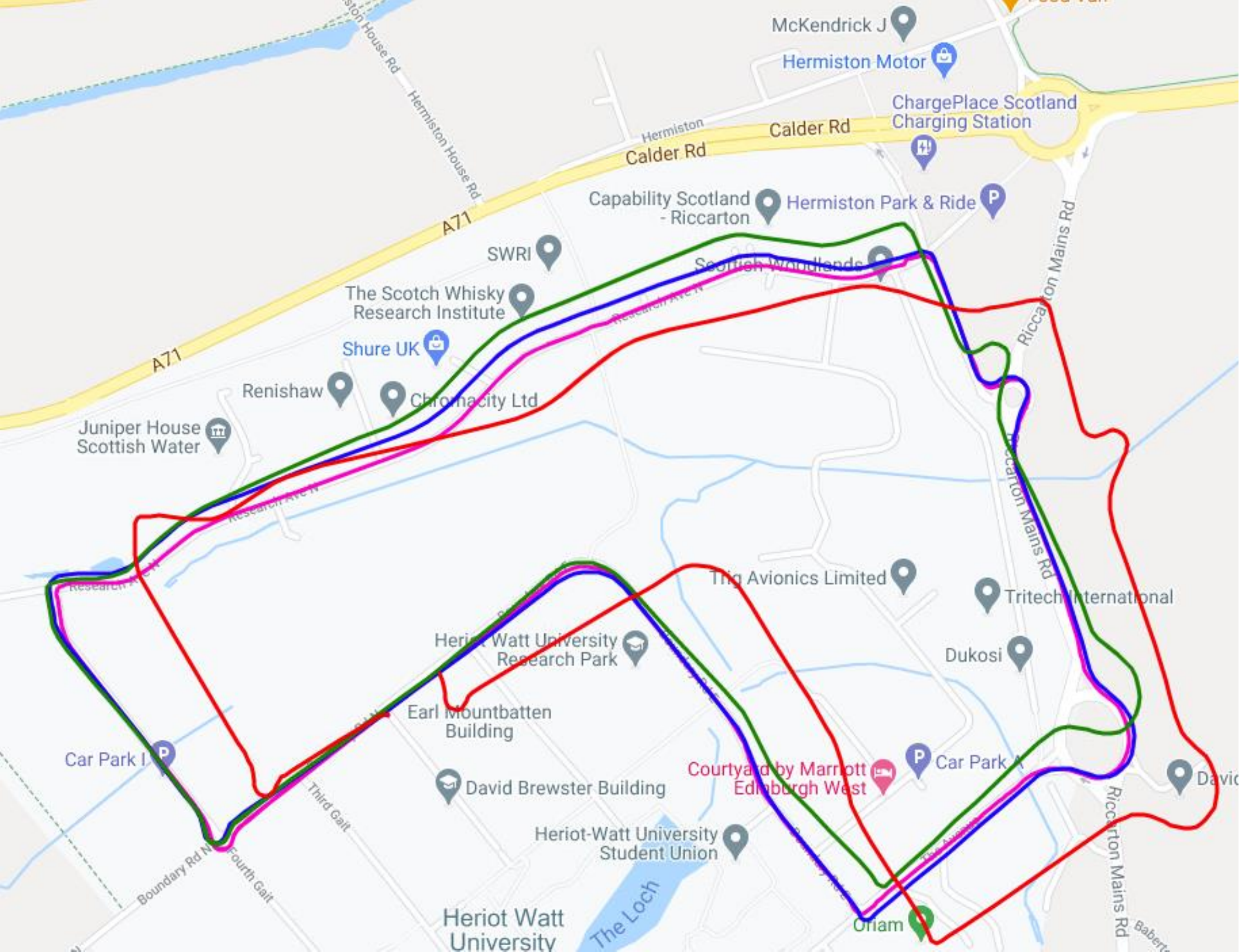}
    \caption{Rain sequence}
    \label{fig:rain_google}
    \end{subfigure}
    \begin{subfigure}{0.21\linewidth}
    \includegraphics[width=\linewidth]{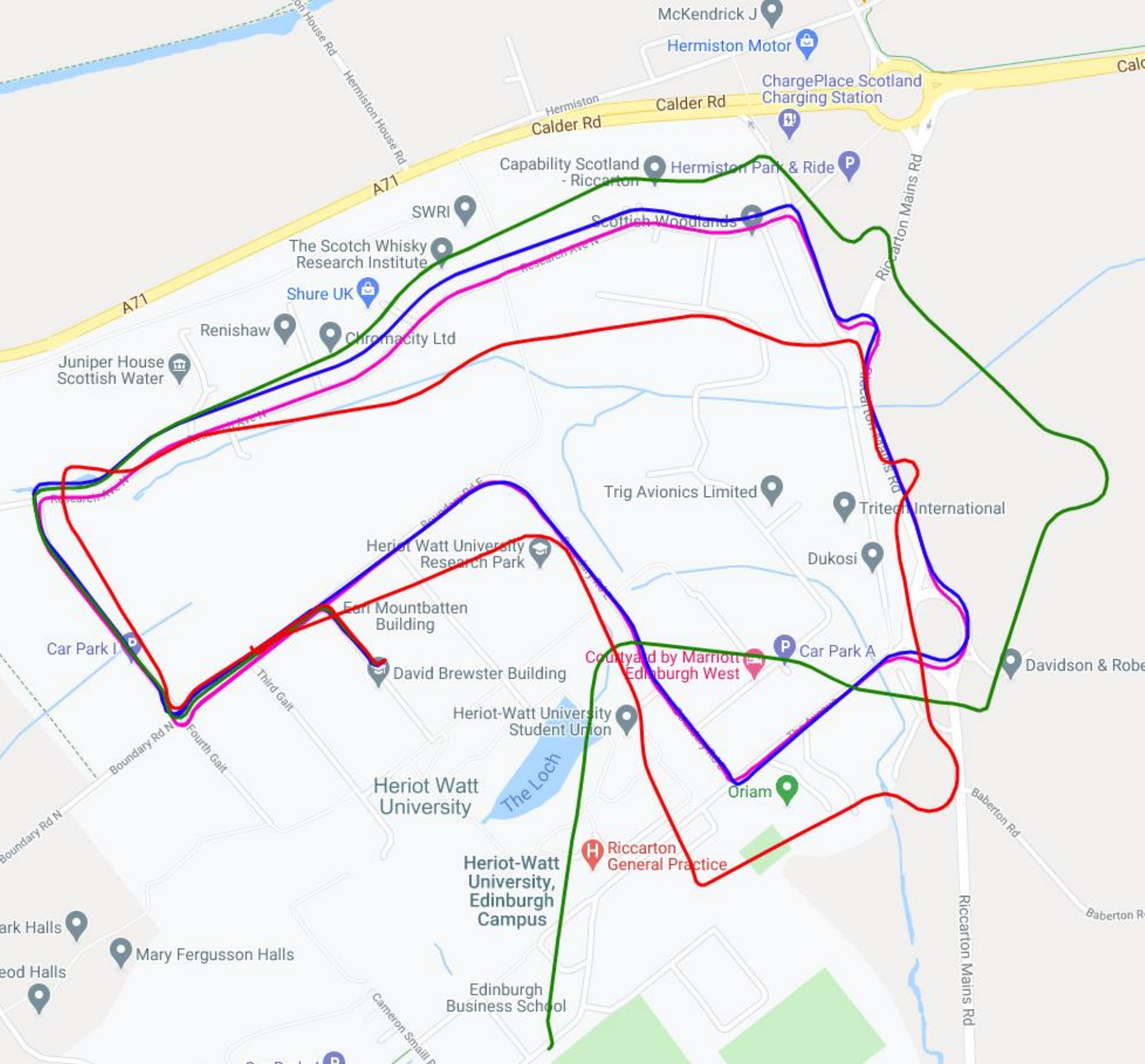}
    \caption{Snow 2 sequence}
    \label{fig:snow2_google}
    \end{subfigure}
    \begin{subfigure}{0.21\linewidth}
    \includegraphics[width=\linewidth]{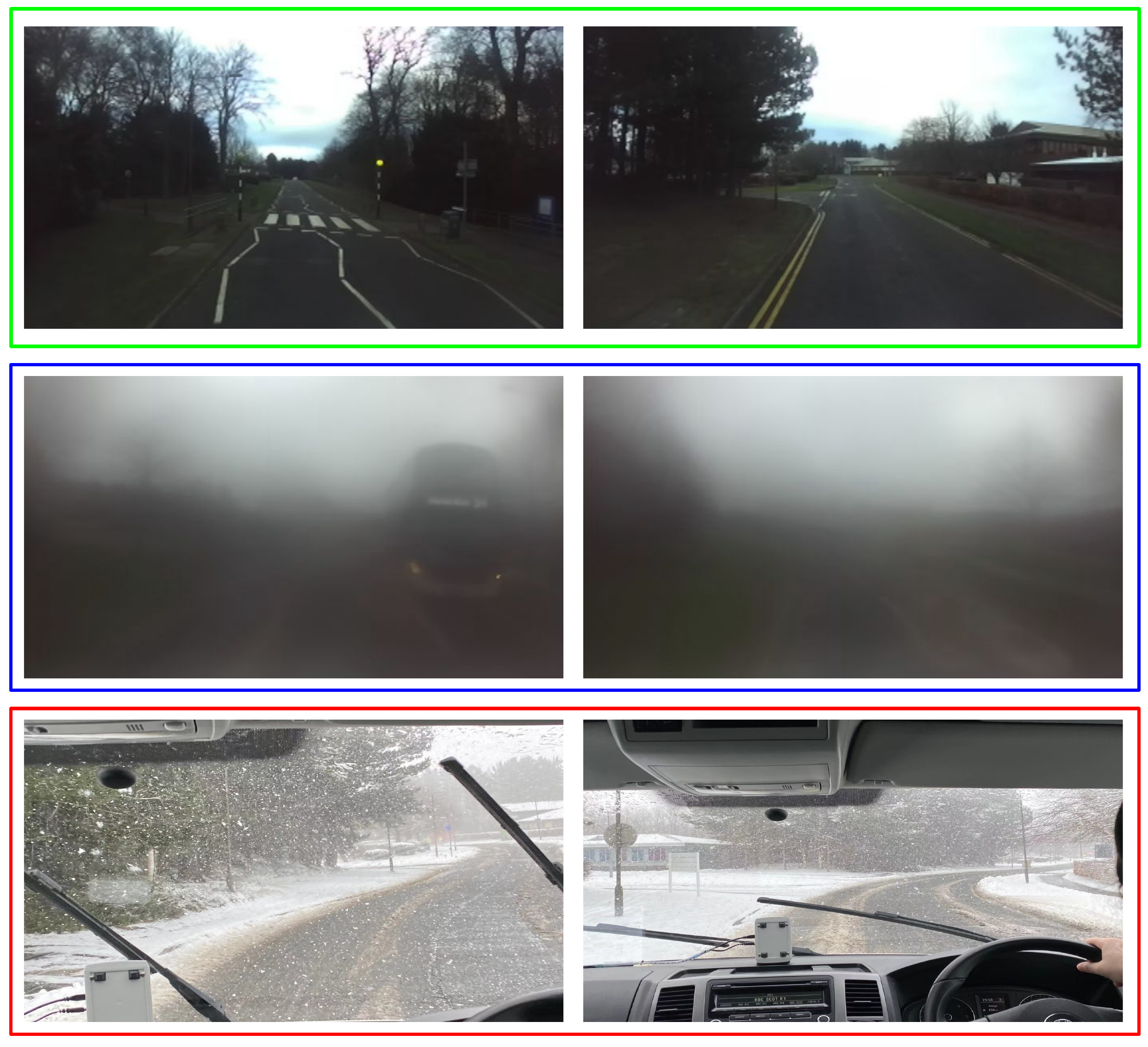}
    \caption{Images}
    \label{fig:campus_images}
    \end{subfigure}
    \caption{(a)-(c): Estimated trajectories and ground truth on multi-session traversals in normal, rain and snow conditions. (d)Top: image data in normal weather captured by our stereo camera, centre: image data in rain captured by our stereo camera, bottom: image data in snow captured by our phone for reference.}
    \label{fig:same_place_different_weather}
\end{figure*}


To further verify the superiority of radar against LiDAR and camera in adverse weathers and degraded visual environments, we perform qualitative evaluation by comparing the estimated trajectories with a high-precision Inertial Navigation System (inertial system fused with GPS) using our RADIATE dataset. Since ORB-SLAM2 fails to produce meaningful results due to the visual degradation caused by water drops, blurry effects in low-light conditions and occlusion from snow (see Fig. \ref{fig:radiate_camera_data} for example images), its results are not reported in this section.

\subsubsection{Experiments in Adverse Weather}

Estimated trajectories of SuMa, baseline and our odometry for Fog 1 and Fog 2 are shown in Figs. \ref{fig:fog1_google} and \ref{fig:fog2_google} respectively. We can see that our SLAM drifts less than SuMa and the baseline radar SLAM although they all suffer from drift without loop closure. SuMa also loses tracking for sequence Fog 2, which is likely due to the impact of fog on LiDAR sensing.

The impact of snowflakes on LiDAR reflection is more obvious. Fig. \ref{fig:same_place_snow_norlmal} shows the LiDAR point clouds of two of the same places in snowy and normal conditions. Depending on the snow density, we can see two types of degeneration of LiDAR in snow. It is clear that both the number of correct LiDAR reflections and point intensity dramatically drop in snow for place 1, while there are a lot of noisy detections around the origin for place 2. Both cases can be challenging for LiDAR based odometry/SLAM methods. This matches the results of the Snow sequence in Fig. \ref{fig:snow_google}. Specifically, when the snow was initially light, SuMa was operating well. However, when the snow gradually became heavier, the LiDAR data degraded and eventually SuMa lost track. The three examples of LiDAR scans at the point when SuMa fails are shown in Fig. \ref{fig:snow_google}. The very limited surrounding structures sensed by LiDAR makes it extremely challenging for LiDAR odometry/SLAM methods like SuMa. In contrast, our radar SLAM method is still able to operate accurately in heavy snow, estimating a more accurate trajectory than the baseline SLAM.

\subsubsection{Experiments on the Same Route in Different Weathers}
To compare different algorithms' performance on the same route but in different weather conditions, we also provide results here in normal weather, rain and snow conditions respectively. The estimated trajectories of SuMa, baseline SLAM and our SLAM result in normal weather are shown in Fig. \ref{fig:good_weather_google} while for the Rain sequence these are shown in Fig. \ref{fig:rain_google}. In the Rain sequence, there is moderate rain. LiDAR based SuMa is slightly affected, and as we can see at the beginning of the sequence, SuMa estimates a shorter length. Our radar SLAM also performs better than the baseline SLAM. In the Snow 2 sequence, there is moderate snow, and the results are shown in Fig. \ref{fig:snow2_google}. The Snow 2 sequence was taken while moving quickly. Therefore, without motion compensation, the baseline SLAM drifts heavily and cannot close the loop while our SLAM consistently performs well.
Hence, the results in Fig. \ref{fig:same_place_different_weather} once again confirm the our proposed SLAM system is robust in all weather conditions.

\subsubsection{Experiments at Night}
The estimated trajectories of SuMa, baseline SLAM and our SLAM on the Night sequence are shown in Fig. \ref{fig:night_google}. LiDAR based SuMa is almost unaffected by the dark night although it does not detect the loops. Both baseline and our SLAM perform well in the night sequence, producing more accurate trajectories after detecting loop closures.



\subsection{Average Completion Percentage}
We calculate the average completion percentage for each competing algorithm on each dataset, to evaluate the robustness of each algorithm representing a different sensor modality. The number of frames that a method completed before losing tracking is denoted as $K_{completed}$ while the total number of frames is denoted as $K_{total}$. The metric is computed as:

\begin{equation}
    Percentage = K_{completed} / K_{total} * 100 \%
\end{equation}
The MulRan dataset does not include camera data so it is shown as N/A for ORB-SLAM2. In the RADIATE dataset, the camera is either blocked by snow or blurred in the night so ORB-SLAM2 fails to initialize and it is also shown as N/A. In Table \ref{tab:completion} we can see that only the radar based methods are reliable and completed in all cases. Neither vision based nor LiDAR based methods manage to finish on all three datasets.

\begin{table}[h]
    \centering
    \caption{Completion Percentage \%}
    \footnotesize
    \begin{tabular}{ c|ccc}
    \hline
    & \multicolumn{3}{c}{\textbf{Dataset}} \\
    \textbf{Method}   & Oxford & MulRan & RADIATE  \\
    \hline
    SuMa   &20 &72 &63  \\
    Baseline SLAM &100 &100 &100     \\
    ORB-SLAM2 &100 &N/A & N/A      \\
    Our SLAM &100 &100 &100      \\
    \hline
    \end{tabular}\par
    \smallskip
 Completion Percentage on Different Datasets
   \label{tab:completion}
\end{table}






\subsection{Parameters Used}

The \textit{same} set of parameters provided in Table \ref{tab:completion} is employed in all the experiments, covering different cities, radar resolutions and ranges, weather conditions, road scenarios, etc.

\begin{table}[h]
    \centering
    \footnotesize
    \caption{Parameters for Radar SLAM}
    \begin{tabular}{ccp{4cm}}
    \hline
     Parameter & Value & Note  \\
    \hline
    Max polar distance & 87.5 & Maximum selected distance in radar reading in meters in our experiments \\
    Min Hessian & 700 & Minimum Hessian value a point to be considered as keypoint  \\
    $\delta_c$ & 3 & Pixel value for maximal clique in graph outlier rejection in Eq. \ref{eq:pairwise_constraint} \\
    Max tracked points & 60 & Maximum number of points in tracking\\
    Keyframe distance  & 2.0 &  Distance between keyframes in meters \\
    Keyframe rotation  & 0.2  &  Rotation between keyframes in radians     \\
    $r_{pca}$& 3 & PCA ratio to reject loop candidate in \ref{sec:PCA} \\
    \hline
    \end{tabular}\par
    \smallskip

   \label{tab:completion}
\end{table}

\subsection{Runtime}
\begin{figure}[h!]
    \centering
    \includegraphics[width=\linewidth]{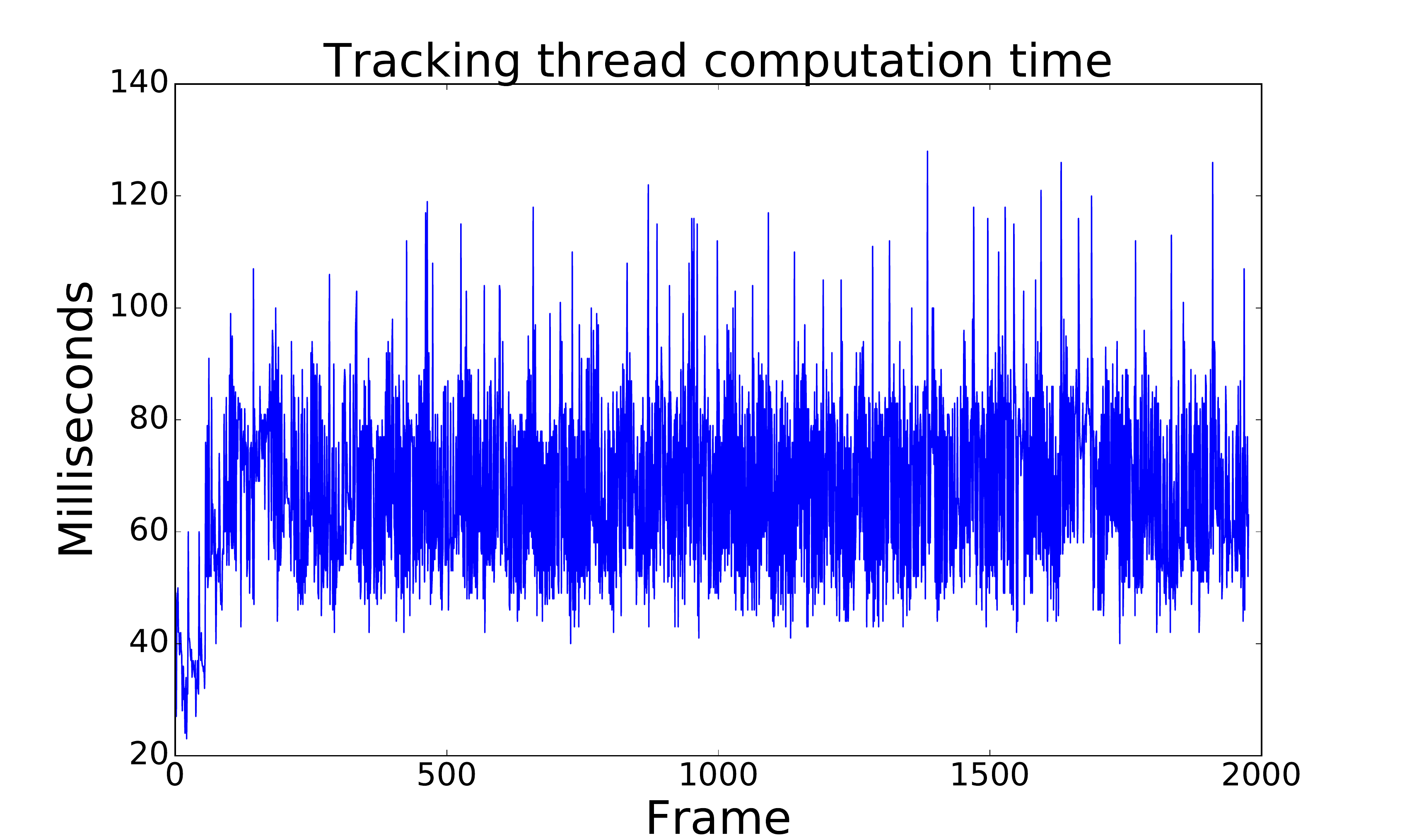}
    \caption{Computation time of tracking on the Rain sequence.}
    \label{fig:computation_time}
\end{figure}

The system is implemented in C++ without a GPU. The computation time of a tracking thread is shown in Fig. \ref{fig:computation_time} showing that our proposed system runs at ~8Hz, which is twice as fast as the 4 Hz radar frame rate, on a laptop with an Intel i7 2.60GHz CPU and 16 GB RAM. The loop closure and pose graph optimization are performed with an independent thread which does not affect our real-time performance.

\section{Conclusion} \label{sec:conclusion}
In this paper we have presented a FMCW radar based SLAM system that includes pose tracking, loop closure and pose graph optimization. To address the motion distortion problem in radar sensing, we formulate the pose tracking as an optimization problem that explicitly compensates for the motion without the aid of other sensors. A robust loop closure detection scheme is specifically designed for the FMCW radar. The proposed system is agnostic to the radar resolutions, radar range, environment and weather conditions. The same set of system parameters is used for the evaluation of three different datasets covering different cities and weather conditions.

Extensive experiments show that the proposed FMCW radar SLAM algorithm achieves comparable localization accuracy in normal weather compared to the state-of-the-art LiDAR and vision based SLAM algorithms. More importantly, it is the only one that is resilient to adverse weather conditions, e.g. snow and fog, demonstrating the superiority and promising potential of using FMCW radar as the primary sensor for long-term mobile robot localization and navigation tasks. 

For future work, we seek to use the map built by our SLAM system and perform long-term localization on it across all weather conditions.

\section*{Acknowledgements}
We thank Joshua Roe, Ted Ding, Saptarshi Mukherjee, Dr. Marcel Sheeny and Dr. Yun Wu for the help of our data collection.
This work was supported by EPSRC Robotics and Artificial Intelligence ORCA Hub (grant No. EP/R026173/1) and EU H2020 Programme under EUMarineRobots project (grant ID 731103).

\bibliographystyle{IEEEtran}
\bibliography{root}

\end{document}